\documentclass{article}

\PassOptionsToPackage{numbers, sort&compress}{natbib}

\usepackage[final]{neurips_2021}

\usepackage[utf8]{inputenc} %
\usepackage[T1]{fontenc}    %
\usepackage{hyperref}       %
\usepackage{url}            %
\usepackage{booktabs}       %
\usepackage{amsfonts}       %
\usepackage{nicefrac}       %
\usepackage{microtype}      %
\usepackage{xcolor}         %

\usepackage{algorithm}      %
\usepackage[noend]{algorithmic}    %
\usepackage{graphicx}
\usepackage{multirow}       %
\usepackage{makecell}       %
\usepackage{pgfplots}
\pgfplotsset{compat=1.16}
\usepackage{subfig}
\usepackage{enumitem}       %
\usepackage{wrapfig}        %
\usepackage{tabularx}       %
\usepackage{longtable}      %
\usepackage{pdflscape}      %
\usepackage{adjustbox}      %
\usepackage[graphicx]{realboxes} %
\usepackage{rotating}       %

\def\fillandplacepagenumber{%
 \par\pagestyle{empty}%
 \vbox to 0pt{\vss}\vfill
 \vbox to 0pt{\baselineskip0pt
   \hbox to\linewidth{\hss}%
   \baselineskip\footskip
   \hbox to\linewidth{%
     \hfil\thepage\hfil}\vss}}

\newcommand{\adj}{\mA}
\newcommand{\weight}{\mW}
\newcommand{\features}{\mX}
\newcommand{\featset}{\sX}
\newcommand{\softout}{\vs}
\newcommand{\neighbors}{\sN}

\newcommand{\pertm}{\tilde{\mX}_\epsilon}
\newcommand{\pertmset}{\tilde{\sX}_\epsilon}

\newtheorem{theorem}{Theorem}
\newtheorem{proposition}{Proposition}
\newtheorem{definition}{Definition}

\newenvironment{proof}{}{$\square$}

\graphicspath{ {./assets/} }

\usepackage{amsmath,amsfonts,bm}

\def\eqref#1{equation~\ref{#1}}

\def\1{\bm{1}}

\def\vc{{\bm{c}}}

\def\vi{{\bm{i}}}

\def\vp{{\bm{p}}}

\def\vs{{\bm{s}}}

\def\vu{{\bm{u}}}
\def\vv{{\bm{v}}}

\def\vx{{\bm{x}}}
\def\vy{{\bm{y}}}
\def\vz{{\bm{z}}}

\def\evc{{c}}

\def\evp{{p}}

\def\evu{{u}}

\def\evy{{y}}
\def\evz{{z}}

\def\mA{{\bm{A}}}
\def\mB{{\bm{B}}}

\def\mD{{\bm{D}}}

\def\mI{{\bm{I}}}

\def\mP{{\bm{P}}}

\def\mW{{\bm{W}}}
\def\mX{{\bm{X}}}

\DeclareMathAlphabet{\mathsfit}{\encodingdefault}{\sfdefault}{m}{sl}
\SetMathAlphabet{\mathsfit}{bold}{\encodingdefault}{\sfdefault}{bx}{n}

\def\gG{{\mathcal{G}}}

\def\sC{{\mathbb{C}}}

\def\sN{{\mathbb{N}}}

\def\sR{{\mathbb{R}}}

\def\sV{{\mathbb{V}}}

\def\sX{{\mathbb{X}}}

\newcommand{\E}{\mathbb{E}}

\newcommand{\R}{\mathbb{R}}

\DeclareMathOperator*{\argmin}{arg\,min}

\title{Robustness of Graph Neural Networks at Scale}

\author{%
  Simon~Geisler,
  Tobias Schmidt, Hakan Şirin, \\ \textbf{Daniel~Z\"ugner, Aleksandar Bojchevski, and Stephan~G\"unnemann} \\
  Technical University of Munich, Department of Informatics\\
  \texttt{\{geisler, schmidtt, sirin, zuegnerd, bojchevs, guennemann\}@in.tum.de} \\
}

\begin{document}

\maketitle

\begin{abstract}
  Graph Neural Networks (GNNs) are increasingly important given their popularity and the diversity of applications. Yet, existing studies of their vulnerability to adversarial attacks rely on relatively small graphs. We address this gap and study how to attack and defend GNNs at scale. We propose two sparsity-aware first-order optimization attacks that maintain an efficient representation despite optimizing over a number of parameters which is quadratic in the number of nodes. We show that common surrogate losses are not well-suited for global attacks on GNNs. Our alternatives can double the attack strength. Moreover, to improve GNNs' reliability we design a robust aggregation function, Soft Median, resulting in an effective defense at all scales. We evaluate our attacks and defense with standard GNNs on graphs more than 100 times larger compared to previous work. We even scale one order of magnitude further by extending our techniques to a scalable GNN.
\end{abstract}

\section{Introduction}\label{sec:intro} %

\begin{wrapfigure}[16]{r}{0.45\textwidth}
  \centering
  \vspace{-17pt}
  \resizebox{0.875\linewidth}{!}{\input{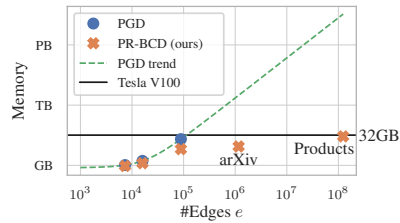}}
  \caption{GPU memory consumption for a global attack with Projected Gradient Descent (PGD)~\citep{Xu2019a}, its quadratic extrapolation, and our \emph{Projected Randomized Block Coordinate Descent (PR-BCD)} (\autoref{sec:attack}).
  Both yield similar adversarial accuracy.
  Beyond attacks, our defense %
  (\autoref{sec:defense}) also scales to these graphs. 
  \label{fig:memorycomparison}}
\end{wrapfigure}

The evidence that Graph Neural Networks (GNNs) are not robust to adversarial perturbations is compelling~\citep{Zugner2018, Dai2018, GNNBook-ch8-gunnemann}. However, the graphs in previous robustness studies are tiny. This is worrying, given that GNNs are already deployed in many real-world Internet-scale applications \citep{Bojchevski2020a, Ying2018a}.
For example, PubMed~\citep{Sen2008} (19,717 nodes) is often considered to be a large-scale graph and around 20~GB of memory is required for an attack based on its dense adjacency matrix.
Such memory requirements are impractical
and limit advancements of the field. In this work, we set the foundation for the holistic study of adversarial robustness of GNNs at scale.
We study graphs with up to 111 million nodes for local attacks (i.e.\ attacking a single node) and 2.5 million nodes for global attacks (i.e.\ attacking all nodes at once). As it turns out, GNNs at scale are also highly vulnerable to adversarial attacks. In \autoref{fig:memorycomparison}, we show the substantial improvement of memory efficiency of our attack over a popular prior work for attacking a GNN globally.\looseness=-1 

\textbf{Scope.} We focus on adversarial robustness w.r.t.\ structure attacks on GNNs for node classification %
\begin{equation}\label{eq:attack}
  \max_{\tilde{\adj} \text{ s.t.\ } \|\tilde{\adj} - \adj\|_0 < \Delta} \mathcal{L}(f_{\theta}(\tilde{\adj}, \features))
\end{equation}
with loss function \(\mathcal{L}\) (or its surrogate \(\mathcal{L}'\)), budget \(\Delta\), and \emph{fixed} model parameters \(\theta\). The GNN \(f_{\theta}(\adj, \features)\) is applied to a graph \(\gG=(\adj, \features)\) with node attributes \(\features \in \sR^{n \times d}\), the adjacency matrix \(\adj \in \left\{0, 1\right\}^{n \times n}\), and \(m\) edges.
We focus on \emph{evasion} (test time) attacks, but our methods can be used in
\emph{poisoning} (train time) attacks~\cite{Zugner2019a}. We distinguish between \emph{local} attacks on a single node and \emph{global} attacks that target a large fraction of nodes with a shared budget \(\Delta\).
We study white-box attacks since they have the most powerful threat model and can be used to understand the robustness w.r.t.\ ``worst-case noise'' of a model, as well as to assess the efficacy of defenses. For these reasons \emph{white-box attacks are important and practical from the perspective of a defender}.

\textbf{Broader impact.} Since we enable the study of robustness at scale, which previously was practically infeasible, an adversary can potentially abuse our attacks. 
The risk is minimized given that we assume perfect knowledge about the graph, model, and labels. Nonetheless, our findings suggest that one should be careful when deploying GNNs, and highlight that further research is needed. To mitigate this risk, we must be able to evaluate it. We also propose a scalable defense that shows strong performance empirically but we urge practitioners to consider potential trade-offs, e.g.\ improving robustness at the expense of accuracy for different groups of users.\looseness=-1

\textbf{Contributions.} We address three major challenges hindering the study of GNNs' adversarial robustness at scale and propose viable solutions including an extensive empirical evaluation: (1) Previous losses are not well-suited for global attacks on GNNs; (2) Attacks on GNNs scale quadratically in the number of nodes or worse; (3) Similarly, previous robust GNNs are typically not scalable.

\textbf{(1) Surrogate loss.} We study the limitation of state-of-the-art surrogate losses for attacking the accuracy of a GNN over all nodes~\citep{chen_fast_2018, Wu2019, Xu2019a, Zugner2019a, Chen2020, Ma2020, feng_scalable_2020} in \autoref{sec:ceisbad}. Especially in combination with small/realistic budgets \(\Delta\) and on large graphs, previous surrogate losses lead to weak attacks. In particular, Cross Entropy (CE) or the widely used Carlini-Wagner loss~\citep{Xu2019a, Carlini2017} are weak surrogates for such global attacks. Our novel losses that overcome these limitations easily improve the strength of the attack by 100\% on common datasets. For larger graphs, this gap even becomes more significant. %

\textbf{(2) Attacks.} Attacks solving a discrete optimization problem easily become computationally infeasible because of the vast amount of potential adjacency matrices (\smash{\(\mathcal{O}(2^{n^2})\)}).
An approximate solution can be found with first-order optimization but we then still optimize over a quadratic number of parameters (\(n^2\)). There is no trivial way to sparsify existing attacks as we need to represent each edge explicitly to obtain its gradient (i.e.\ space complexity \(\Theta(n^2)\)). Nevertheless, we overcome this limitation and propose two strategies to apply first-order optimization without the burden of a dense adjacency matrix. In \autoref{sec:attack}, we describe how to add/remove edges between existing nodes based on Randomized Block Coordinate Descent (R-BCD) at an additional memory requirement of \(\mathcal{O}(\Delta)\).
Due to the limited scalability of traditional GNNs, we also consider the case where we attack PPRGo \cite{Bojchevski2020a}, a scalable GNN. Here, we even obtain an algorithm with constant complexity in the nodes \(n\).

\textbf{(3) Defense.} We propose \emph{Soft Median} in \autoref{sec:defense} -- a computationally efficient, robust, differentiable aggregation function inspired by~\citet{Geisler2020}, by taking advantage of recent advancements in differentiable sorting~\cite{Prillo2020}. Using our Soft Median we observe similar robustness to~\citep{Geisler2020}, but with a significantly lower memory footprint, which enables us to defend GNNs at scale.

\section{Surrogate Losses for Global Attacks}\label{sec:ceisbad}

During training and in first-order attacks, we ideally wish to optimize a target metric which is often discontinuous (e.g.\ accuracy and 0/1 loss \(\mathcal{L}_{0/1}\)). However, for gradient-based optimization we commonly substitute the actual target loss by a \emph{surrogate} \(\mathcal{L}' \approx \mathcal{L}\) (e.g.\ cross entropy for \(\mathcal{L}_{0/1}\)). In the
context of i.i.d.\ samples (e.g.\ images), a single example is attacked in isolation with its own budget, which is similar to a \emph{local} attack for GNNs. When a single node's prediction is attacked, it is often sufficient to \textit{maximize}
the cross entropy for the attacked node/image (untargeted attack): \(
\text{CE}(y, \vp) 
= - \log(\evp_{c^*}) \).
where \(y\) is the label and \(\vp\) is the vector of confidence scores.

\begin{figure}[ht]
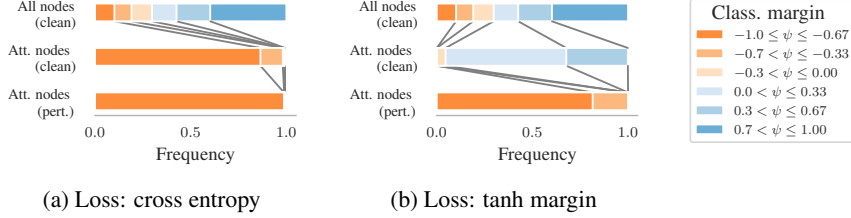

  \centering
  \vspace{-15pt}
  \makebox[\linewidth][c]{
    \(\begin{array}{ccc}
      \subfloat[Loss: cross entropy]{\resizebox{0.30\linewidth}{!}{\input{assets/global_prbcd_bar_pubmed_False_0.01_CE_relfeq_margin_attacked_no_legend.pgf}}} &
      \subfloat[Loss: tanh margin]{\resizebox{0.30\linewidth}{!}{\input{assets/global_prbcd_bar_pubmed_False_0.01_tanhMargin_relfeq_margin_attacked_no_legend.pgf}}} &
      \raisebox{2ex}[0pt][0pt]{\resizebox{0.18\linewidth}{!}{
\begingroup%
\makeatletter%
\begin{pgfpicture}%
\pgfpathrectangle{\pgfpointorigin}{\pgfqpoint{1.540343in}{1.365705in}}%
\pgfusepath{use as bounding box, clip}%
\begin{pgfscope}%
\pgfsetbuttcap%
\pgfsetmiterjoin%
\definecolor{currentfill}{rgb}{1.000000,1.000000,1.000000}%
\pgfsetfillcolor{currentfill}%
\pgfsetlinewidth{0.000000pt}%
\definecolor{currentstroke}{rgb}{1.000000,1.000000,1.000000}%
\pgfsetstrokecolor{currentstroke}%
\pgfsetstrokeopacity{0.000000}%
\pgfsetdash{}{0pt}%
\pgfpathmoveto{\pgfqpoint{0.000000in}{0.000000in}}%
\pgfpathlineto{\pgfqpoint{1.540343in}{0.000000in}}%
\pgfpathlineto{\pgfqpoint{1.540343in}{1.365705in}}%
\pgfpathlineto{\pgfqpoint{0.000000in}{1.365705in}}%
\pgfpathclose%
\pgfusepath{fill}%
\end{pgfscope}%
\begin{pgfscope}%
\pgfsetbuttcap%
\pgfsetmiterjoin%
\definecolor{currentfill}{rgb}{1.000000,1.000000,1.000000}%
\pgfsetfillcolor{currentfill}%
\pgfsetfillopacity{0.800000}%
\pgfsetlinewidth{1.003750pt}%
\definecolor{currentstroke}{rgb}{0.800000,0.800000,0.800000}%
\pgfsetstrokecolor{currentstroke}%
\pgfsetstrokeopacity{0.800000}%
\pgfsetdash{}{0pt}%
\pgfpathmoveto{\pgfqpoint{0.122222in}{0.100000in}}%
\pgfpathlineto{\pgfqpoint{1.418120in}{0.100000in}}%
\pgfpathquadraticcurveto{\pgfqpoint{1.440343in}{0.100000in}}{\pgfqpoint{1.440343in}{0.122222in}}%
\pgfpathlineto{\pgfqpoint{1.440343in}{1.243483in}}%
\pgfpathquadraticcurveto{\pgfqpoint{1.440343in}{1.265705in}}{\pgfqpoint{1.418120in}{1.265705in}}%
\pgfpathlineto{\pgfqpoint{0.122222in}{1.265705in}}%
\pgfpathquadraticcurveto{\pgfqpoint{0.100000in}{1.265705in}}{\pgfqpoint{0.100000in}{1.243483in}}%
\pgfpathlineto{\pgfqpoint{0.100000in}{0.122222in}}%
\pgfpathquadraticcurveto{\pgfqpoint{0.100000in}{0.100000in}}{\pgfqpoint{0.122222in}{0.100000in}}%
\pgfpathclose%
\pgfusepath{stroke,fill}%
\end{pgfscope}%
\begin{pgfscope}%
\definecolor{textcolor}{rgb}{0.150000,0.150000,0.150000}%
\pgfsetstrokecolor{textcolor}%
\pgfsetfillcolor{textcolor}%
\pgftext[x=0.275270in,y=1.106474in,left,base]{\color{textcolor}\rmfamily\fontsize{12.000000}{14.400000}\selectfont Class. margin}%
\end{pgfscope}%
\begin{pgfscope}%
\pgfsetbuttcap%
\pgfsetmiterjoin%
\definecolor{currentfill}{rgb}{0.991419,0.550727,0.232772}%
\pgfsetfillcolor{currentfill}%
\pgfsetlinewidth{1.003750pt}%
\definecolor{currentstroke}{rgb}{1.000000,1.000000,1.000000}%
\pgfsetstrokecolor{currentstroke}%
\pgfsetdash{}{0pt}%
\pgfpathmoveto{\pgfqpoint{0.144444in}{0.940741in}}%
\pgfpathlineto{\pgfqpoint{0.366667in}{0.940741in}}%
\pgfpathlineto{\pgfqpoint{0.366667in}{1.018519in}}%
\pgfpathlineto{\pgfqpoint{0.144444in}{1.018519in}}%
\pgfpathclose%
\pgfusepath{stroke,fill}%
\end{pgfscope}%
\begin{pgfscope}%
\definecolor{textcolor}{rgb}{0.150000,0.150000,0.150000}%
\pgfsetstrokecolor{textcolor}%
\pgfsetfillcolor{textcolor}%
\pgftext[x=0.455556in,y=0.940741in,left,base]{\color{textcolor}\rmfamily\fontsize{8.000000}{9.600000}\selectfont \(\displaystyle -1.0 \le \psi \le -0.67\)}%
\end{pgfscope}%
\begin{pgfscope}%
\pgfsetbuttcap%
\pgfsetmiterjoin%
\definecolor{currentfill}{rgb}{0.992157,0.726797,0.491503}%
\pgfsetfillcolor{currentfill}%
\pgfsetlinewidth{1.003750pt}%
\definecolor{currentstroke}{rgb}{1.000000,1.000000,1.000000}%
\pgfsetstrokecolor{currentstroke}%
\pgfsetdash{}{0pt}%
\pgfpathmoveto{\pgfqpoint{0.144444in}{0.785803in}}%
\pgfpathlineto{\pgfqpoint{0.366667in}{0.785803in}}%
\pgfpathlineto{\pgfqpoint{0.366667in}{0.863581in}}%
\pgfpathlineto{\pgfqpoint{0.144444in}{0.863581in}}%
\pgfpathclose%
\pgfusepath{stroke,fill}%
\end{pgfscope}%
\begin{pgfscope}%
\definecolor{textcolor}{rgb}{0.150000,0.150000,0.150000}%
\pgfsetstrokecolor{textcolor}%
\pgfsetfillcolor{textcolor}%
\pgftext[x=0.455556in,y=0.785803in,left,base]{\color{textcolor}\rmfamily\fontsize{8.000000}{9.600000}\selectfont \(\displaystyle -0.7 < \psi \le -0.33\)}%
\end{pgfscope}%
\begin{pgfscope}%
\pgfsetbuttcap%
\pgfsetmiterjoin%
\definecolor{currentfill}{rgb}{0.994833,0.874556,0.753033}%
\pgfsetfillcolor{currentfill}%
\pgfsetlinewidth{1.003750pt}%
\definecolor{currentstroke}{rgb}{1.000000,1.000000,1.000000}%
\pgfsetstrokecolor{currentstroke}%
\pgfsetdash{}{0pt}%
\pgfpathmoveto{\pgfqpoint{0.144444in}{0.630864in}}%
\pgfpathlineto{\pgfqpoint{0.366667in}{0.630864in}}%
\pgfpathlineto{\pgfqpoint{0.366667in}{0.708642in}}%
\pgfpathlineto{\pgfqpoint{0.144444in}{0.708642in}}%
\pgfpathclose%
\pgfusepath{stroke,fill}%
\end{pgfscope}%
\begin{pgfscope}%
\definecolor{textcolor}{rgb}{0.150000,0.150000,0.150000}%
\pgfsetstrokecolor{textcolor}%
\pgfsetfillcolor{textcolor}%
\pgftext[x=0.455556in,y=0.630864in,left,base]{\color{textcolor}\rmfamily\fontsize{8.000000}{9.600000}\selectfont \(\displaystyle -0.3 < \psi \le 0.00\)}%
\end{pgfscope}%
\begin{pgfscope}%
\pgfsetbuttcap%
\pgfsetmiterjoin%
\definecolor{currentfill}{rgb}{0.840692,0.901638,0.958662}%
\pgfsetfillcolor{currentfill}%
\pgfsetlinewidth{1.003750pt}%
\definecolor{currentstroke}{rgb}{1.000000,1.000000,1.000000}%
\pgfsetstrokecolor{currentstroke}%
\pgfsetdash{}{0pt}%
\pgfpathmoveto{\pgfqpoint{0.144444in}{0.475926in}}%
\pgfpathlineto{\pgfqpoint{0.366667in}{0.475926in}}%
\pgfpathlineto{\pgfqpoint{0.366667in}{0.553704in}}%
\pgfpathlineto{\pgfqpoint{0.144444in}{0.553704in}}%
\pgfpathclose%
\pgfusepath{stroke,fill}%
\end{pgfscope}%
\begin{pgfscope}%
\definecolor{textcolor}{rgb}{0.150000,0.150000,0.150000}%
\pgfsetstrokecolor{textcolor}%
\pgfsetfillcolor{textcolor}%
\pgftext[x=0.455556in,y=0.475926in,left,base]{\color{textcolor}\rmfamily\fontsize{8.000000}{9.600000}\selectfont \(\displaystyle 0.0 < \psi \le 0.33\)}%
\end{pgfscope}%
\begin{pgfscope}%
\pgfsetbuttcap%
\pgfsetmiterjoin%
\definecolor{currentfill}{rgb}{0.671895,0.814379,0.900654}%
\pgfsetfillcolor{currentfill}%
\pgfsetlinewidth{1.003750pt}%
\definecolor{currentstroke}{rgb}{1.000000,1.000000,1.000000}%
\pgfsetstrokecolor{currentstroke}%
\pgfsetdash{}{0pt}%
\pgfpathmoveto{\pgfqpoint{0.144444in}{0.320988in}}%
\pgfpathlineto{\pgfqpoint{0.366667in}{0.320988in}}%
\pgfpathlineto{\pgfqpoint{0.366667in}{0.398766in}}%
\pgfpathlineto{\pgfqpoint{0.144444in}{0.398766in}}%
\pgfpathclose%
\pgfusepath{stroke,fill}%
\end{pgfscope}%
\begin{pgfscope}%
\definecolor{textcolor}{rgb}{0.150000,0.150000,0.150000}%
\pgfsetstrokecolor{textcolor}%
\pgfsetfillcolor{textcolor}%
\pgftext[x=0.455556in,y=0.320988in,left,base]{\color{textcolor}\rmfamily\fontsize{8.000000}{9.600000}\selectfont \(\displaystyle 0.3 < \psi \le 0.67\)}%
\end{pgfscope}%
\begin{pgfscope}%
\pgfsetbuttcap%
\pgfsetmiterjoin%
\definecolor{currentfill}{rgb}{0.417086,0.680631,0.838231}%
\pgfsetfillcolor{currentfill}%
\pgfsetlinewidth{1.003750pt}%
\definecolor{currentstroke}{rgb}{1.000000,1.000000,1.000000}%
\pgfsetstrokecolor{currentstroke}%
\pgfsetdash{}{0pt}%
\pgfpathmoveto{\pgfqpoint{0.144444in}{0.166049in}}%
\pgfpathlineto{\pgfqpoint{0.366667in}{0.166049in}}%
\pgfpathlineto{\pgfqpoint{0.366667in}{0.243827in}}%
\pgfpathlineto{\pgfqpoint{0.144444in}{0.243827in}}%
\pgfpathclose%
\pgfusepath{stroke,fill}%
\end{pgfscope}%
\begin{pgfscope}%
\definecolor{textcolor}{rgb}{0.150000,0.150000,0.150000}%
\pgfsetstrokecolor{textcolor}%
\pgfsetfillcolor{textcolor}%
\pgftext[x=0.455556in,y=0.166049in,left,base]{\color{textcolor}\rmfamily\fontsize{8.000000}{9.600000}\selectfont \(\displaystyle 0.7 < \psi \le 1.00\)}%
\end{pgfscope}%
\end{pgfpicture}%
\makeatother%
\endgroup%
}} \\
    \end{array}\)
  }
    \caption{Margin \(\psi\) of test nodes vs.\ attacked nodes, before (``clean'') and after perturbation (``perturbed''). We attack the PubMed graph (\autoref{tab:datasets}) and a single-layer GCN with one percent of edges (\(\epsilon=0.01\)) as a budget. In stark contrast to the \text{tanh margin} (b), the CE loss (a) spends a lot of its budget on misclassified nodes (i.e.\ \(\psi < 0\)). See \autoref{sec:appendix_ceisbad} for more variants and details.}%
  \label{fig:negceprob}
\end{figure}

Many \emph{global} attacks for GNNs~\citep{chen_fast_2018, Wu2019, Xu2019a, Zugner2019a} maximize the average CE to attack all nodes with a combined budget \(\Delta\). However, a loss like CE can be ineffective, particularly when the number of nodes is large in comparison to the budget \(\nicefrac{\Delta}{n} \to 0\). While experimenting on large graphs, we often observed that the CE loss increases even though the accuracy does not decline (see \autoref{sec:appendix_ceisbad}). As we can see in \autoref{fig:negceprob}, this is due to CE's bias towards nodes that have a low confidence score. With CE and a sufficiently small budget \(\Delta \ll n\) we primarily attack nodes that are already misclassified, which means that the classification margin \(\psi = \min_{c \ne c^*} \evp_{c^*} - \evp_{c}\) is already negative.

\textbf{Global attack.} In contrast to attacking a single image/node, a global attack on a GNN has to (1) keep house with the budget \(\Delta\) and (2) find edges that degrade the overall accuracy maximally (i.e.\ target ``fragile'' nodes). Without additional information, intuitively, one would first attack low-confidence nodes close to the decision boundary. Hence, the surrogate loss should have a minimal (maximally negative) gradient at \(\psi \to 0^+\) (i.e.\ approaching \(\psi \to 0\) from \(\psi \ge 0\)). Moreover, if we solely want to lower the accuracy, then we can stop attacking a node once it is misclassified\footnote{We simply write \(\vp = f_{\theta}(\adj, \features)\) omitting that \(\vp\) belongs to specific node, i.e.\ \(\vp_i\) of node \(i\). We also overload this with the logits / pre-softmax activation as \(\vz = f_{\theta}(\adj, \features)\). See \autoref{sec:appendix_notation} for all notation.}:

\begin{definition}\label{definition:budgetaware}
    A surrogate loss \(\mathcal{L}'\) for global attacks \emph{\textbf{(I)}} should only incentivize perturbing nodes that are correctly classified: \(\nicefrac{\partial \mathcal{L}'}{\partial \evz_{c^*}} |_{\psi < 0} = 0\) and \emph{\textbf{(II)}} should favour nodes close to the decision boundary: \(\nicefrac{\partial \mathcal{L}'}{\partial \evz_{c^*}} |_{\psi_0} < \nicefrac{\partial \mathcal{L}'}{\partial \evz_{c^*}} |_{\psi_1}\) for any \(0 < \psi_0 < \psi_1\).
\end{definition}

Since \autoref{eq:attack} is in general a discrete and non-convex optimization problem that is often NP-complete~\citep{Bojchevski2019a, zhang_efficient_2018, andriushchenko_provably_2019, weng_towards_2018}, we propose to study the surrogate loss under the subsequent simplifying assumptions. Note that in an actual attack other influences (e.g.\ node degree) are still considered while solving the optimization problem. \emph{Assumption 1}: The set of attacked nodes is independent (their receptive fields do not overlap). Particularly on large graphs with small budgets, \(\nicefrac{\Delta}{n} \to 0\), deciding which node to attack becomes an increasingly local decision since the receptive field becomes insignificant in comparison to the rest of the graph. \emph{Assumption 2}: The budget required to change the prediction of node $i$ depends (only) on the margin: \(\Delta_i = g(|\psi_i|)\) for some increasing and non-negative function \(g(\cdot)\). That is, the larger the margin $\psi_i$, the harder it is to attack node $i$. As stated in \autoref{proposition:goodsurrogate}, with these assumptions, an optimizer with a surrogate loss compliant with \autoref{definition:budgetaware} is also optimizing the 0/1 loss \(\mathcal{L}_{0/1}\) (for proof see \autoref{sec:appendix_proofprop1}).
\begin{proposition}\label{proposition:goodsurrogate}
  Let \(\mathcal{L}'\) be the surrogate for the 0/1 loss \(\mathcal{L}_{0/1}\) used to attack a node classification algorithm \(f_{\theta}(\adj, \features)\) with a global budget \(\Delta\). %
  Suppose we greedily attack nodes in order of \(\nicefrac{\partial \mathcal{L}'}{\partial \evz_{c^*}}(\psi_0) \le \nicefrac{\partial \mathcal{L}'}{\partial \evz_{c^*}}(\psi_1) \le \dots \le \nicefrac{\partial \mathcal{L}'}{\partial \evz_{c^*}}(\psi_l)\) until the budget is exhausted \(\Delta < \sum_{i=0}^{l+1} \Delta_i\).
  Under Assumptions 1 \& 2, we then obtain the global optimum of
  \(
    \max_{\tilde{\adj}\text{ s.t.\ }\|\tilde{\adj} - \adj\|_0 < \Delta} \mathcal{L}_{0/1}(f_{\theta}(\tilde{\adj}, \features))\,
  \)
  if \(\mathcal{L}'\) has the properties \textbf{\emph{(I)}} \(\nicefrac{\partial \mathcal{L}'}{\partial \evz_{c^*}} |_{\psi < 0} = 0\) and \textbf{\emph{(II)}} \(\nicefrac{\partial \mathcal{L}'}{\partial \evz_{c^*}} |_{\psi_0} < \nicefrac{\partial \mathcal{L}'}{\partial \evz_{c^*}} |_{\psi_1}\) for any \(0 < \psi_0 < \psi_1\).\looseness=-1
\end{proposition}

Even under the simplifying Assumptions 1 \& 2, the Cross Entropy (CE) is not guaranteed to obtain the global optimum. The (CE) violates property (I) and in the worst case only perturbs nodes that are already misclassified (see \autoref{fig:negceprob}). The Carlini-Wagner (CW)~\cite{Carlini2017, Xu2019a} loss \(\text{CW} = \min(\max_{c \ne c^*} \evz_{c} - \evz_{c^*}, 0)\) violates property (II). It is also not guaranteed to obtain the global optimum, i.e.\ CW loss lacks focus on nodes close to the decision boundary. In the worst case, an attack with CW spends all budget on confident nodes---without flipping a single one.

We propose the Masked Cross Entropy \(\text{MCE} = \nicefrac{1}{|\sV^+|} \sum_{i \in \sV^+} -\log(\evp_{c^*}^{(i)})\) which fulfills both properties for binary classification by only considering correctly classified nodes \(\sV^+\) and, hence, may reach the global optimum. Empirically, for a greedy gradient-based attack the \(\text{MCE}\) comes with gains of more than 200\% in strength (see \autoref{fig:empsurrogate}). Surprisingly, if we apply \(\text{MCE}\) to a Projected Gradient Descent (PGD) attack, we observe hardly any improvement over CE. We identify two potential reasons for that. The first is due to the learning dynamics of PGD. Suppose a misclassified node does not receive any weight in the gradient update, now if the budget is exceeded after the update it is likely to be down-weighted. This can lead to nodes that oscillate around the decision boundary (for more details see \autoref{sec:appendix_ceisbad}). A similar behavior occurs for to the Carlini-Wagner loss in e.g.\ \autoref{fig:appendix_learningcurves_directed} (e).

In \autoref{definition:confidentandbudgetaware}, we relax properties (I)/(II) and propose to overcome these limitations via enforcing confidently misclassified nodes, i.e.\ we want the attacked nodes to be at a ``safe'' distance from the decision boundary. We propose the tanh of the margin in logit space, i.e. \(\text{tanh margin} = \tanh(\max_{c \ne c^*} \evz_{c} - \evz_{c^*})\). It obeys \autoref{definition:confidentandbudgetaware} and its effectiveness is apparent from \autoref{fig:negceprob}. For the empirical evaluation see \autoref{sec:empirical} and for more results as well as details on all selected losses see \autoref{sec:appendix_ceisbad}. In the appendix we also study further losses to deepen the understanding about the required properties. Additionally, in \autoref{sec:appendix_alternativeprop1}, we give an alternative \autoref{proposition:goodsurrogate} for a relaxed Assumption 2 s.t.\ \(\mathbb{E}[\Delta_i|\psi_i] = g(|\psi_i|)\)\looseness=-1
\begin{definition}\label{definition:confidentandbudgetaware}
    A surrogate loss \(\mathcal{L}'\) for global attacks that encourages confident misclassification \emph{\textbf{(A)}} should saturate \(\lim_{\psi \to -1^+} \mathcal{L}' < \infty\) and \emph{\textbf{(B)}} should favor points close to the decision boundary: \(\nicefrac{\partial \mathcal{L}'}{\partial \evz_{c^*}} |_{\psi_0} < \nicefrac{\partial \mathcal{L}'}{\partial \evz_{c^*}} |_{\psi_1} < 0\) for any \(0 < \psi_0 < \psi_1 < 1\) or \(-1 < \psi_1 < \psi_0 < 0\).
\end{definition}

\section{Scalable Attacks}\label{sec:attack}

Beginning with \citep{Dai2018, Zugner2018}, many adversarial attacks on the graph structure have been proposed.
As discussed, gradient-based attacks such as \citep{Zugner2018, Zugner2019a, Wu2019, Chen2020} aim for lower computational cost by approximating the corresponding discrete optimization problem. However, they optimize all possible entries in the \textit{dense} adjacency matrix \(\adj\) which comes with quadratic space complexity \(\Theta(n^2)\). Since previous attacks come with limited scalability (e.g.\ see \autoref{fig:memorycomparison}), GNNs robustness on larger graphs is largely unexplored. First, we propose a family of attacks that does not require a dense adjacency matrix and comes with linear complexity w.r.t.\ the budget \(\Delta\). Then, we further improve the complexity of our attack for a scalable GNN called PPRGo~\citep{Bojchevski2020a}.

\textbf{Related work.} \citet{li_adversarial_2021} evaluate their \emph{local} adversarial attack SGA only on a graph with around 200k nodes and SGA is specifically designed for Simplified Graph Convolution (SGC)~\citep{wu_simplifying_2019}. Note that a two-layer SGC is identical to Nettack's surrogate model. We consider arbitrary Graph Neural Networks and we even scale our \emph{global} attack to a graph ten times larger. With PPRGo we even outscale them by factor 500. \citet{feng_scalable_2020} partition the graph to lower the attack's memory footprint but still have a time complexity of \(\mathcal{O}(n^2)\). \citet{Dai2018} scale their reinforcement learning approach to a graph for financial transactions with 2.5 million nodes. In contrast to our work, they scale their \emph{local} attack only using a tiny budget \(\Delta\) of \emph{a single edge deletion} and only need to consider the receptive field of a single node. We scale our local attack to 111M nodes and allow large budgets \(\Delta\).\looseness=-1

\textbf{Large scale optimization.} In some big data use cases, the cost to calculate the gradient towards all variables can be prohibitively high. For this reason, coordinate descent has gained importance in machine learning and large-scale optimization~\citep{Wright2015}. \citet{Nesterov2012} proposed (and analyzed the convergence of) Randomized Block Coordinate Descent (R-BCD). In R-BCD only a subset (called a block) of variables is optimized at a time and, hence, only the gradients towards those variables are required. In many cases, this allows for a lower memory footprint and in some settings even converges faster than standard methods~\citep{Nesterov2017}.

For clarity, we model the perturbations \(\mP \in \{0, 1\}^{n \times n}\) explicitly ($\mP_{ij} = 1$ denotes an edge flip):\looseness=-1
\begin{equation}\label{eq:pgd}
  \max_{\mP\,\,\text{s.t.}\, \mP \in \{0, 1\}^{n \times n} \text{, } \sum \mP \le \Delta} \mathcal{L}(f_{\theta}(\adj \oplus \mP, \features))\,.
\end{equation}
Here, \(\oplus\) stands for an element-wise exclusive or and \(\Delta\) denotes the edge budget (i.e.\ the number of altered entries in the perturbed adjacency matrix). Naively, applying R-BCD to optimize towards the dense adjacency matrix would only save some computation on obtaining the respective gradient. It still has a space complexity of \(\mathcal{O}(n^2)\) on top of the complexity of the attacked model because we still have to store up to \(n^2\) parameters.
Note that the \(L_0\) perturbation constraint with limited budget \(\Delta\) implies that the solution will be sparse. We build upon this fact and in each epoch, in a survival-of-the-fittest manner, we keep that part of the search space which is ``promising'' and resample the rest. Despite the differences, we simply call our approach \textbf{Projected Randomized Block Coordinate Descent (PR-BCD)} and provide the pseudo code in Algo.~\ref{algo:prbcd} (a preliminary version appeared in \citep{geisler_attacking_2021}). On top of the GNN, PR-BCD comes with space complexity of \(\Theta(b)\) where \(b\) is the block size (number of coordinates) since \emph{everything can be implemented efficiently with sparse operations}. We typically choose \(\Delta\) to be a fraction of \(m\) and \(b > \Delta\), thus, in practice, we have a linear overhead.

\begin{figure}
    \begin{minipage}{0.49\textwidth}
      \centering
      \hbox{\resizebox{\linewidth}{!}{
\begingroup%
\makeatletter%
\begin{pgfpicture}%
\pgfpathrectangle{\pgfpointorigin}{\pgfqpoint{3.937196in}{0.388266in}}%
\pgfusepath{use as bounding box, clip}%
\begin{pgfscope}%
\pgfsetbuttcap%
\pgfsetmiterjoin%
\definecolor{currentfill}{rgb}{1.000000,1.000000,1.000000}%
\pgfsetfillcolor{currentfill}%
\pgfsetlinewidth{0.000000pt}%
\definecolor{currentstroke}{rgb}{1.000000,1.000000,1.000000}%
\pgfsetstrokecolor{currentstroke}%
\pgfsetstrokeopacity{0.000000}%
\pgfsetdash{}{0pt}%
\pgfpathmoveto{\pgfqpoint{0.000000in}{0.000000in}}%
\pgfpathlineto{\pgfqpoint{3.937196in}{0.000000in}}%
\pgfpathlineto{\pgfqpoint{3.937196in}{0.388266in}}%
\pgfpathlineto{\pgfqpoint{0.000000in}{0.388266in}}%
\pgfpathclose%
\pgfusepath{fill}%
\end{pgfscope}%
\begin{pgfscope}%
\pgfsetbuttcap%
\pgfsetmiterjoin%
\definecolor{currentfill}{rgb}{1.000000,1.000000,1.000000}%
\pgfsetfillcolor{currentfill}%
\pgfsetfillopacity{0.800000}%
\pgfsetlinewidth{1.003750pt}%
\definecolor{currentstroke}{rgb}{0.800000,0.800000,0.800000}%
\pgfsetstrokecolor{currentstroke}%
\pgfsetstrokeopacity{0.800000}%
\pgfsetdash{}{0pt}%
\pgfpathmoveto{\pgfqpoint{0.122222in}{0.100000in}}%
\pgfpathlineto{\pgfqpoint{3.814974in}{0.100000in}}%
\pgfpathquadraticcurveto{\pgfqpoint{3.837196in}{0.100000in}}{\pgfqpoint{3.837196in}{0.122222in}}%
\pgfpathlineto{\pgfqpoint{3.837196in}{0.266044in}}%
\pgfpathquadraticcurveto{\pgfqpoint{3.837196in}{0.288266in}}{\pgfqpoint{3.814974in}{0.288266in}}%
\pgfpathlineto{\pgfqpoint{0.122222in}{0.288266in}}%
\pgfpathquadraticcurveto{\pgfqpoint{0.100000in}{0.288266in}}{\pgfqpoint{0.100000in}{0.266044in}}%
\pgfpathlineto{\pgfqpoint{0.100000in}{0.122222in}}%
\pgfpathquadraticcurveto{\pgfqpoint{0.100000in}{0.100000in}}{\pgfqpoint{0.122222in}{0.100000in}}%
\pgfpathclose%
\pgfusepath{stroke,fill}%
\end{pgfscope}%
\begin{pgfscope}%
\pgfsetbuttcap%
\pgfsetroundjoin%
\pgfsetlinewidth{1.003750pt}%
\definecolor{currentstroke}{rgb}{0.298039,0.447059,0.690196}%
\pgfsetstrokecolor{currentstroke}%
\pgfsetdash{}{0pt}%
\pgfpathmoveto{\pgfqpoint{0.255556in}{0.149378in}}%
\pgfpathlineto{\pgfqpoint{0.255556in}{0.260489in}}%
\pgfusepath{stroke}%
\end{pgfscope}%
\begin{pgfscope}%
\pgfsetroundcap%
\pgfsetroundjoin%
\pgfsetlinewidth{1.003750pt}%
\definecolor{currentstroke}{rgb}{0.298039,0.447059,0.690196}%
\pgfsetstrokecolor{currentstroke}%
\pgfsetdash{}{0pt}%
\pgfpathmoveto{\pgfqpoint{0.144444in}{0.204933in}}%
\pgfpathlineto{\pgfqpoint{0.366667in}{0.204933in}}%
\pgfusepath{stroke}%
\end{pgfscope}%
\begin{pgfscope}%
\definecolor{textcolor}{rgb}{0.150000,0.150000,0.150000}%
\pgfsetstrokecolor{textcolor}%
\pgfsetfillcolor{textcolor}%
\pgftext[x=0.455556in,y=0.166044in,left,base]{\color{textcolor}\rmfamily\fontsize{8.000000}{9.600000}\selectfont Vanilla GCN}%
\end{pgfscope}%
\begin{pgfscope}%
\pgfsetbuttcap%
\pgfsetroundjoin%
\pgfsetlinewidth{1.003750pt}%
\definecolor{currentstroke}{rgb}{0.866667,0.517647,0.321569}%
\pgfsetstrokecolor{currentstroke}%
\pgfsetdash{}{0pt}%
\pgfpathmoveto{\pgfqpoint{1.454988in}{0.149378in}}%
\pgfpathlineto{\pgfqpoint{1.454988in}{0.260489in}}%
\pgfusepath{stroke}%
\end{pgfscope}%
\begin{pgfscope}%
\pgfsetroundcap%
\pgfsetroundjoin%
\pgfsetlinewidth{1.003750pt}%
\definecolor{currentstroke}{rgb}{0.866667,0.517647,0.321569}%
\pgfsetstrokecolor{currentstroke}%
\pgfsetdash{}{0pt}%
\pgfpathmoveto{\pgfqpoint{1.343877in}{0.204933in}}%
\pgfpathlineto{\pgfqpoint{1.566099in}{0.204933in}}%
\pgfusepath{stroke}%
\end{pgfscope}%
\begin{pgfscope}%
\definecolor{textcolor}{rgb}{0.150000,0.150000,0.150000}%
\pgfsetstrokecolor{textcolor}%
\pgfsetfillcolor{textcolor}%
\pgftext[x=1.654988in,y=0.166044in,left,base]{\color{textcolor}\rmfamily\fontsize{8.000000}{9.600000}\selectfont Vanilla GDC}%
\end{pgfscope}%
\begin{pgfscope}%
\pgfsetbuttcap%
\pgfsetroundjoin%
\pgfsetlinewidth{1.003750pt}%
\definecolor{currentstroke}{rgb}{0.333333,0.658824,0.407843}%
\pgfsetstrokecolor{currentstroke}%
\pgfsetdash{}{0pt}%
\pgfpathmoveto{\pgfqpoint{2.656117in}{0.149378in}}%
\pgfpathlineto{\pgfqpoint{2.656117in}{0.260489in}}%
\pgfusepath{stroke}%
\end{pgfscope}%
\begin{pgfscope}%
\pgfsetroundcap%
\pgfsetroundjoin%
\pgfsetlinewidth{1.003750pt}%
\definecolor{currentstroke}{rgb}{0.333333,0.658824,0.407843}%
\pgfsetstrokecolor{currentstroke}%
\pgfsetdash{}{0pt}%
\pgfpathmoveto{\pgfqpoint{2.545006in}{0.204933in}}%
\pgfpathlineto{\pgfqpoint{2.767228in}{0.204933in}}%
\pgfusepath{stroke}%
\end{pgfscope}%
\begin{pgfscope}%
\definecolor{textcolor}{rgb}{0.150000,0.150000,0.150000}%
\pgfsetstrokecolor{textcolor}%
\pgfsetfillcolor{textcolor}%
\pgftext[x=2.856117in,y=0.166044in,left,base]{\color{textcolor}\rmfamily\fontsize{8.000000}{9.600000}\selectfont Soft Median GDC}%
\end{pgfscope}%
\end{pgfpicture}%
\makeatother%
\endgroup%
}}
      \vspace{-10pt}
      \makebox[\linewidth][c]{
        \(\begin{array}{cc}
          \subfloat[Cora ML ($n=2.8k$)]{\resizebox{0.5\linewidth}{!}{\input{assets/global_PRBCD_novel_loss_cora_ml_0.1_block_size_no_legend.pgf}}} &
          \subfloat[arXiv ($n=170k$)]{\resizebox{0.48\linewidth}{!}{\input{assets/global_prbcd_arxiv_0.1_block_size_cmp_epochs_accuracy.pgf}}}  %
        \end{array}\)
      }
      \caption{
      Influence of block size \(b\) on PR-BCD (dashed \(L_0\) PGD~\citep{Xu2019a}) with tanh margin loss and \(\epsilon=0.1\).
      (a) shows adv.\ accuracy with three-sigma error over five seeds. We resample \(E_{\text{res.}}=50\) epochs and then fine-tune 250. (b) shows adv.\ accuracy over epochs \(t\) with \(E \cdot b=\text{const.}\) \label{fig:randomblocksizeinfluence}}
    \end{minipage}
    \hfill
    \begin{minipage}{0.47\textwidth}
    \begin{algorithm}[H]
      \small 
      \caption{Projected Randomized Block\\Coordinate Descent (PR-BCD)}
      \label{algo:prbcd}
      \begin{algorithmic}[1]
        \STATE {\bfseries Input:} Gr.\ \((\adj, \features)\), lab.\ \(\vy\), GNN \(f_{\theta}(\cdot)\), loss \(\mathcal{L}\)
        \STATE {\bfseries Parameter:} budget\ \(\Delta\), block size \(b\), epochs \(E\) \& \(E_{\text{res.}}\), heuristic \(h(\dots)\), learning rate \(\alpha_t\)
        \STATE Draw w/o replacement \(\vi_0 {\in} \{0, 1, \dots, n^2 - 1\}^b\)\hspace{-0.5em}
        \STATE Initialize zeros for \(\vp_0 \in \R^b\)
        \FOR{\(t \in \{1,2, \dots, E\}\)}
        \STATE \(\hat{\vy} \leftarrow f_{\theta}(\adj \oplus \vp_{t-1}, \features)\)
        \STATE \(\vp_{t} \leftarrow \vp_{t-1} + \alpha_{t} \nabla_{\vp_{t-1}[\vi_{t-1}]} \mathcal{L}(\hat{\vy}, \vy)\)
        \STATE Projection \(\vp_{t} \leftarrow \Pi_{\E[\text{Bernoulli}(\vp_t)] \le \Delta} (\vp_{t})\)
        \STATE \(\vi_{t} \leftarrow \vi_{t-1}\)
        \IF{\(t \le E_{\text{res.}}\)}
        \STATE \(\text{mask}_{\text{res.}} \leftarrow h(\vp_{t})\)
        \STATE \(\vp_t[\text{mask}_{\text{res.}}] \leftarrow \mathbf{0}\)
        \STATE Resample \(\vi_{t}[\text{mask}_{\text{res.}}]\)
        \ENDIF
        \ENDFOR
        \STATE \(\mP \sim \text{Bernoulli}(\vp_{E})\) s.t.\ \(\sum \mP \le \Delta\)
        \STATE Return \(\adj \oplus \mP\)
      \end{algorithmic}
    \end{algorithm}
    \end{minipage}
\end{figure}

\textbf{PR-BCD}. For \(L_0\)-norm PGD we relax the discrete edge perturbations \(\mP\) from \(\{0, 1\}^{(n \times n)}\) to \([0, 1]^{(n \times n)}\) as proposed by~\citet{Xu2019a}. Each entry of \(\mP\) denotes the probability for flipping it. In each epoch we only look at a randomly sampled, non-contiguous block of \(\mP\) of size \(b\) (line 3, line 10-13) and additionally ignore the diagonal elements (i.e.\ self-loops). If using an undirected graph, the potential edges are restricted to the upper/lower triangular \(n \times n\) matrix. In each epoch \(t \in \{1,2, \dots\}\), \(\vp\) is added to / subtracted from the discrete edge weight (line 6).
Note, we overload \(\oplus\) s.t.\ \(\adj_{ij} \oplus p_{ij} = \adj_{ij} + p_{ij}\) if \(\adj_{ij} = 0\) and \(\adj_{ij} - p_{ij}\) otherwise. We use \(\vp\) and \(\mP\) interchangeably while \(\vp\) only corresponds to the current subset/block of \(\mP_{\vi_t}\). After each gradient update (line 7), the projection \(\Pi_{\E[\text{Bernoulli}(\vp)] \le \Delta} (\vp)\) adjusts the probability mass such that \(\E[\text{Bernoulli}(\vp)] = \sum_{i \in b} \evp_i \le \Delta\) and that \(\vp \in [0, 1]\) (line 8). 
In the end we draw \(b\) sample s.t.\ \(\mP \in \{0, 1\}^{(n \times n)}\) via \(\mP \sim \text{Bernoulli}(\vp)\) (line 14).\looseness=-1

The projection \(\Pi_{\E[\text{Bernoulli}(\vp)] \le \Delta} (\vp)\) likely results in many zero elements, but is not guaranteed to be sparse (for details see \autoref{sec:appendix_l0pgd}). If \(\vp\) has more than 50\% non-zero entries, we remove the entries with the lowest probability mass such that 50\% of the search space is resampled. Otherwise, we resample all zero entries in \(\vp\). However, one also might apply a more sophisticated heuristic \(h(\vp)\) which we leave for future work (see line 11). After \(E_{\text{res.}}\) epochs we fine-tune \(\vp\), i.e.\ we stop resampling and decay the learning rate as in~\cite{Xu2019a}. We also employ early stopping for both stages (\(t \le E_{\text{res.}} \text{ and } t > E_{\text{res.}}\) with the epoch \(t\)) such that we take the result of the epoch with highest loss \(\mathcal{L}\).\looseness=-1

\textbf{Block size \(b\).} With growing \(n\) it is unrealistic that each possible entry of the adjacency matrix was part of at least one random search space of (P)R-BCD. As is apparent, with a constant search space size, the number of mutually exclusive chunks of the perturbation matrix grows with \(\Theta(n^2)\) and this would imply a quadratic runtime. However, as evident in randomized black-box attacks~\citep{Waniek2018}, it is not necessary to test every possible edge to obtain an effective attack. In \autoref{fig:randomblocksizeinfluence} (a), we analyze the influence of the block size \(b\) on the adversarial accuracy. On small datasets and over a wide range of block sizes \(b\), our method performs comparably (or sometimes even better) to its dense equivalent. For larger graphs, we observe that the block size \(b\) has a stronger influence on the adversarial accuracy. However, as shown in \autoref{fig:randomblocksizeinfluence} (b), one might increase the number of epochs for an improved attack strength. This indicates that PR-BCD successfully keeps the harmful edges.

\textbf{GR-BCD.} As an alternative to PR-BCD, we propose Greedy R-BCD (GR-BCD) which greedily flips the entries with the largest gradient in the block so that after \(E\) iterations the budget is met. It is even a little bit more scalable as it does \emph{not} require \(b > \Delta\) (see \autoref{sec:appendix_pgrbcd} for details and pseudo code).

\textbf{Limitations.} We solely propose approximate attacks that do not provide any guarantee on how well they approximate the actual optimization problem and, hence, only provide an \emph{upper} bound on e.g.\ the adversarial accuracy. We also recommend monitoring the relaxation error. One could use certificates to get the respective lower bound, provided they were scalable enough. Even sparse smoothing~\cite{Bojchevski2020} might be too slow since we need many forward passes. As our attacks rely on the gradient they also require that the victim model is (approximately) differentiable. Otherwise, the approximation can become inappropriate. Moreover, we are limited by the scalability of the attacked GNN as we discuss next. For the theoretical complexities of all studied attacks, we refer to \autoref{sec:appendix_theoretical_complexities}. \looseness=-1

\textbf{Scalable GNNs.}
Up to now, we implicitly assumed that we have enough memory to obtain the predictions and gradient towards the edges. 
GNNs that typically process the whole graph ``at once'', are inherently limited in their scalability. Our PR-BCD attack is even applicable when operating at those limits (see experiments on Products in \autoref{sec:empirical}). To push the limits further, we now consider more scalable GNNs.
Some notable scalable GNNs either sample subgraphs~\citep{Chen2018a, Chiang2019} or, such as PPRGo~\citep{Bojchevski2020a}, simplify the message passing operation. Next, we extend our PR-BCD to a local attack on PPRGo with \emph{constant complexity} \emph{including the (Soft Median) PPRGo} (we introduce the Soft Median in \autoref{sec:defense}).\looseness=-1

\textbf{PPRGo.}To scale to massive graphs effectively, we need to obtain sublinear/constant complexity w.r.t.\ the number of nodes. This severely restricts the possibilities of how one might approach a global attack and is the reason why we now focus on local attacks (i.e.\ attacking single node \(i\)). For an \(L\)-layer message passing GNN we need to recursively compute the \(L\)-hop neighborhood to obtain the prediction of a single node. This makes it difficult to obtain a sublinear space complexity (here including the GNN)---especially if one considers
arbitrary edge insertions. In contrast, PPRGo~\citep{Bojchevski2020a} leverages the Personalized Page Rank (PPR) matrix
\(\boldsymbol{\Pi} = \alpha(\mI - (1-\alpha)\mD^{-1}\mA)^{-1}\) (row normalization) to reduce the number of explicit message passing steps to one (\autoref{eq:pprgo} with feat.\ encoder \(f_{\text{enc}}\)).\looseness=-1

\vspace{-7pt}
\begin{minipage}{.61\linewidth}
\begin{equation}\label{eq:pprgo}
    \vp = \text{softmax} \left[ \text{AGG} \left \{ \left( \boldsymbol{\Pi}_{uv}, f_{\text{enc}}(\vx_u) \right), \forall \, u \in \neighbors'(v) \right \} \right]
\end{equation}
\end{minipage}
\begin{minipage}{.375\linewidth}
\begin{equation}\label{eq:pprrowupdate}
    \tilde{\boldsymbol{\Pi}}_i
    = \alpha \left( \boldsymbol{\Pi}'_i - \frac{\boldsymbol{\Pi}'_{ii} \vv \boldsymbol{\Pi}'}{1 + \vv \boldsymbol{\Pi}'_{:i}} \right)
\end{equation}
\end{minipage}%

\textbf{Differentiable PPR Update.} For a local attack on PPRGo, we require a differentiable update of the respective PPR Scores for an edge perturbation on a weighted graph. We achieve this using the Sherman-Morrison formula through a closed-form
rank-one update of row \(i\) of the PPR matrix in \autoref{eq:pprrowupdate}
where \(\boldsymbol{\Pi}' = \alpha^{-1}\boldsymbol{\Pi}\) and, with degree matrix \(\mD\), \(\vv = (\mD_{ii} + \sum \vp)^{-1} (\mA_i + \vp) - \mD^{-1}_{ii} \mA_i\). This suffices to attack incoming edges of a node and since everything is differentiable (\(\nicefrac{\partial\mathcal{L}'}{\partial \vp}\)) we do not need a surrogate model (common practice~\cite{Zugner2018, li_adversarial_2021, Wang2020}). In \autoref{sec:appendix_localpprupdate}, we give details on the derivation and show how we can leverage the fact that PPRGo uses a \emph{top-\(k\)-sparsified} PPR matrix to obtain constant complexity \(\mathcal{O}(b k)\) (assuming \(b \ll n\) and \(k \ll n\)). With \(\Delta < b\), our approach comes with no restriction on how we can insert or remove incoming edges of a specific node. Other approaches such as~\citep{Dai2018, li_adversarial_2021} gain scalability via restricting the set of admissible nodes for edge perturbations.\looseness=-1

\section{Scalable Defense}\label{sec:defense}

To complete the robustness picture we now shift focus to defenses. Unfortunately, we are not aware of any defense that scales to graphs significantly larger than PubMed.
Thus, we propose a novel, scalable defense based on a robust message-passing aggregation, relying on recent advancements in differentiable sorting~\cite{Prillo2020}. Our \emph{Soft Median} not only comes with the best possible breakdown point of 0.5 but also can have a lower error than its hard equivalent for finite perturbations (see \autoref{sec:appendix_empirical_error}). Moreover, our Soft Median performs similarly to the recent Soft Medoid~\citep{Geisler2020}, but comes with better computational complexity w.r.t.\ the neighborhood size, lower memory footprint, and enables us to scale to bigger graphs. We can also use this aggregation neatly in the PPRGo architecture resulting in the \emph{first defense that scales to massive graphs with over 111M nodes} (see \autoref{eq:pprgo}).

\textbf{Background.} We typically have the message passing framework of a GNN:
\begin{equation}\label{eq:mean-how-powerfull}
  \mathbf{h}^{(l)}_v = \sigma^{(l)} \left[ \text{AGG}^{(l)} \left \{ \left( \adj_{vu}, \mathbf{h}^{(l-1)}_u \weight^{(l)} \right), \forall \, u\in \neighbors'(v) \right \} \right]
\end{equation}
with neighborhood \(\neighbors'(v) = \neighbors(v) \cup v\) including the node itself, the \(l\)-th layer message passing aggregation \(\text{AGG}^{(l)}\), embedding \(\mathbf{h}^{(l)}_v\), normalized adjacency matrix \(\adj\), weights \(\weight^{(l)}\), and activation \(\sigma^{(l)}\).\looseness=-1

\textbf{Related Work.} Following~\citet{GNNBook-ch8-gunnemann}, we classify defenses into three categories: (1) preprocessing~\citep{Entezari2020, Wu2019, Jin2019}, (2) training procedure~\citep{Xu2019a, Zugner2019a, Chen2020a}, and (3) modifications of the architecture~\citep{Zhu2019, Zhang2019a, Geisler2020, Jin2021a, Zhang2020a, Wu2020a, Tang2020, chen_understanding_2021}. All these previous defenses were not evaluated on graphs substantially larger than PubMed. Note GNNGuard~\citep{Zhang2020a} was only evaluated on a subset of arXiv, covering 20\% of the nodes and 6\% of the edges. Even though our attacks lend themselves well for adversarial training but we leave it for future work due to the overhead during training. Instead, we build the observation of \citet{Geisler2020} that common aggregations (e.g.\ sum or mean) in \autoref{eq:mean-how-powerfull} are known to be non-robust. They propose a differentiable robust aggregation for \(\text{AGG}^{(l)}\) and call it Soft Medoid. It is a continuous relaxation of the Medoid and requires the row/column sum over the distance matrix of the embedding of the nodes in the neighborhood. Hence this operation has a quadratic complexity w.r.t.\ the neighborhood size and comes with a sizable memory overhead during training and inference.

\textbf{Soft Median.} Intuitively, the Soft Median is a weighted mean where the weight for each instance is determined based on the distance to the dimension-wise median \(\bar{\vx}\). This way, instances far from the dimension-wise median are filtered out. We define the Soft Median as
\begin{equation}\label{eq:softmedian}
  \mu_{\text{SoftMedian}}(\features)
  = \text{softmax}\left(-\nicefrac{\vc}{T\sqrt{d}} \right)^\top \features
  = \softout^\top\features \approx \argmin\nolimits_{\vx' \in \featset} \| \bar{\vx} - \vx' \|,
\end{equation}
with the distances \(\evc_{v} = \|\bar{\vx} - \features_{v,:}\|\) and number of dimensions \(d\). We use \(\features\) as well as \(\featset\) interchangeably. For a single dimension, this closely resembles the soft sorting operator as proposed in~\citep{Prillo2020} for the central element and can be understood as a soft version of the median. To apply it to multivariate inputs, we rely on the dimension-wise median which can be computed efficiently for practical choices of \(d\). In contrast to the Soft Medoid, we do not require the distances between all input instances %
which makes the Soft Median much more efficient. Assuming \(d\) is sufficiently small, the Soft Median scales linearly with the number of inputs \(|\neighbors'(v)|\).

\textbf{The temperature}. The temperature parameter \(T\) controls the steepness of the weight distribution \(\softout\) between the neighbors.
In the extreme case as \(T \to 0\) we recover the instance which is closest to the dimension-wise Median (i.e. \(\argmin_{\vx' \in \featset} \| \bar{\vx} - \vx' \|\)). In the other extreme case \(T\to\infty\), the Soft Median is equivalent to the sample mean. We observe a similar empirical behavior as~\citet{Geisler2020} and we decide on a temperature value in our experiments by grid search. 

\textbf{Breakdown point.} 
For any finite \(T\), our proposed Soft Median has the best possible breakdown point of 0.5 as we state formally in \autoref{theorem:softmedianbreakdown} (for proof see \autoref{sec:appendix_prooftheo1}). Note that despite the lower complexity compared to Soft Medoid, we maintain the same breakdown point:

\begin{theorem}\label{theorem:softmedianbreakdown}
  Let \(\featset = \{ \mathbf{\mathbf{x}}_1, \dots, \mathbf{\mathbf{x}}_n\} \) be a collection of points in \(\mathbb{R}^d\) with finite coordinates and temperature \(T \in [0, \infty) \). Then the Soft Median location estimator (\autoref{eq:softmedian}) has the finite sample breakdown point of \(\epsilon^*(\mu_{\text{Soft Median}}, \features) = \nicefrac{1}{n} \lfloor \nicefrac{(n+1)}{2}\rfloor \) (asympt.\ \( \lim_{n \to \infty} \epsilon^*(\mu_{\text{SoftMedian}}, \features) = 0.5 \)).
\end{theorem}

\begin{wrapfigure}[16]{R}{0.5\textwidth}
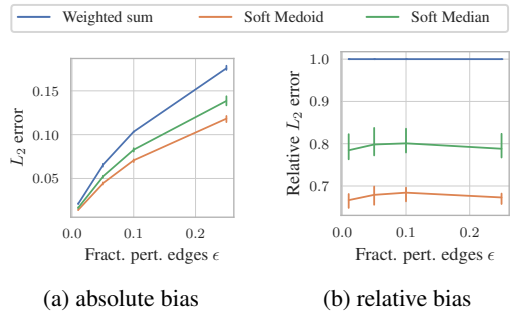

  \vspace{-18pt}
  \centering
  \hbox{\resizebox{\linewidth}{!}{
\begingroup%
\makeatletter%
\begin{pgfpicture}%
\pgfpathrectangle{\pgfpointorigin}{\pgfqpoint{3.657037in}{0.388266in}}%
\pgfusepath{use as bounding box, clip}%
\begin{pgfscope}%
\pgfsetbuttcap%
\pgfsetmiterjoin%
\definecolor{currentfill}{rgb}{1.000000,1.000000,1.000000}%
\pgfsetfillcolor{currentfill}%
\pgfsetlinewidth{0.000000pt}%
\definecolor{currentstroke}{rgb}{1.000000,1.000000,1.000000}%
\pgfsetstrokecolor{currentstroke}%
\pgfsetstrokeopacity{0.000000}%
\pgfsetdash{}{0pt}%
\pgfpathmoveto{\pgfqpoint{0.000000in}{0.000000in}}%
\pgfpathlineto{\pgfqpoint{3.657037in}{0.000000in}}%
\pgfpathlineto{\pgfqpoint{3.657037in}{0.388266in}}%
\pgfpathlineto{\pgfqpoint{0.000000in}{0.388266in}}%
\pgfpathclose%
\pgfusepath{fill}%
\end{pgfscope}%
\begin{pgfscope}%
\pgfsetbuttcap%
\pgfsetmiterjoin%
\definecolor{currentfill}{rgb}{1.000000,1.000000,1.000000}%
\pgfsetfillcolor{currentfill}%
\pgfsetfillopacity{0.800000}%
\pgfsetlinewidth{1.003750pt}%
\definecolor{currentstroke}{rgb}{0.800000,0.800000,0.800000}%
\pgfsetstrokecolor{currentstroke}%
\pgfsetstrokeopacity{0.800000}%
\pgfsetdash{}{0pt}%
\pgfpathmoveto{\pgfqpoint{0.122222in}{0.100000in}}%
\pgfpathlineto{\pgfqpoint{3.534815in}{0.100000in}}%
\pgfpathquadraticcurveto{\pgfqpoint{3.557037in}{0.100000in}}{\pgfqpoint{3.557037in}{0.122222in}}%
\pgfpathlineto{\pgfqpoint{3.557037in}{0.266044in}}%
\pgfpathquadraticcurveto{\pgfqpoint{3.557037in}{0.288266in}}{\pgfqpoint{3.534815in}{0.288266in}}%
\pgfpathlineto{\pgfqpoint{0.122222in}{0.288266in}}%
\pgfpathquadraticcurveto{\pgfqpoint{0.100000in}{0.288266in}}{\pgfqpoint{0.100000in}{0.266044in}}%
\pgfpathlineto{\pgfqpoint{0.100000in}{0.122222in}}%
\pgfpathquadraticcurveto{\pgfqpoint{0.100000in}{0.100000in}}{\pgfqpoint{0.122222in}{0.100000in}}%
\pgfpathclose%
\pgfusepath{stroke,fill}%
\end{pgfscope}%
\begin{pgfscope}%
\pgfsetroundcap%
\pgfsetroundjoin%
\pgfsetlinewidth{1.003750pt}%
\definecolor{currentstroke}{rgb}{0.298039,0.447059,0.690196}%
\pgfsetstrokecolor{currentstroke}%
\pgfsetdash{}{0pt}%
\pgfpathmoveto{\pgfqpoint{0.144444in}{0.204933in}}%
\pgfpathlineto{\pgfqpoint{0.366667in}{0.204933in}}%
\pgfusepath{stroke}%
\end{pgfscope}%
\begin{pgfscope}%
\definecolor{textcolor}{rgb}{0.150000,0.150000,0.150000}%
\pgfsetstrokecolor{textcolor}%
\pgfsetfillcolor{textcolor}%
\pgftext[x=0.455556in,y=0.166044in,left,base]{\color{textcolor}\rmfamily\fontsize{8.000000}{9.600000}\selectfont Weighted sum}%
\end{pgfscope}%
\begin{pgfscope}%
\pgfsetroundcap%
\pgfsetroundjoin%
\pgfsetlinewidth{1.003750pt}%
\definecolor{currentstroke}{rgb}{0.866667,0.517647,0.321569}%
\pgfsetstrokecolor{currentstroke}%
\pgfsetdash{}{0pt}%
\pgfpathmoveto{\pgfqpoint{1.409428in}{0.204933in}}%
\pgfpathlineto{\pgfqpoint{1.631650in}{0.204933in}}%
\pgfusepath{stroke}%
\end{pgfscope}%
\begin{pgfscope}%
\definecolor{textcolor}{rgb}{0.150000,0.150000,0.150000}%
\pgfsetstrokecolor{textcolor}%
\pgfsetfillcolor{textcolor}%
\pgftext[x=1.720539in,y=0.166044in,left,base]{\color{textcolor}\rmfamily\fontsize{8.000000}{9.600000}\selectfont Soft Medoid}%
\end{pgfscope}%
\begin{pgfscope}%
\pgfsetroundcap%
\pgfsetroundjoin%
\pgfsetlinewidth{1.003750pt}%
\definecolor{currentstroke}{rgb}{0.333333,0.658824,0.407843}%
\pgfsetstrokecolor{currentstroke}%
\pgfsetdash{}{0pt}%
\pgfpathmoveto{\pgfqpoint{2.572121in}{0.204933in}}%
\pgfpathlineto{\pgfqpoint{2.794343in}{0.204933in}}%
\pgfusepath{stroke}%
\end{pgfscope}%
\begin{pgfscope}%
\definecolor{textcolor}{rgb}{0.150000,0.150000,0.150000}%
\pgfsetstrokecolor{textcolor}%
\pgfsetfillcolor{textcolor}%
\pgftext[x=2.883232in,y=0.166044in,left,base]{\color{textcolor}\rmfamily\fontsize{8.000000}{9.600000}\selectfont Soft Median}%
\end{pgfscope}%
\end{pgfpicture}%
\makeatother%
\endgroup%
}}
  \vspace{-14pt}
  \makebox[\linewidth][c]{
  \(\begin{array}{cc}
    \subfloat[absolute bias]{\resizebox{0.475\linewidth}{!}{\input{assets/global_pgd_error_0_no_legend.pgf}}} & 
    \subfloat[relative bias]{\resizebox{0.475\linewidth}{!}{\input{assets/global_pgd_relerror_0_no_legend.pgf}}} \\
  \end{array}\)
  }
  \caption{Empirical bias \(B(\epsilon)\) for the second layer of a GCN with GDC preproc.~\citep{Klicpera2019a} network under PGD attack with \(L_2\) distance. We use \(T=0.2\) and a relative budget of \(\epsilon=0.25\). \label{fig:empbiascurve}}
\end{wrapfigure}
We define the \emph{weighted} Soft Median as
\begin{equation}\label{eq:resulting-wsm}
  \mu_{\text{WSM}}(\features, \mathbf{a}) = C\,(\softout \circ \mathbf{a})^\top\features
\end{equation}
where \(\softout\) is the softmax weight of \autoref{eq:softmedian} obtained using the weighted dimension-wise Median, \(C\) normalizes s.t.\ \(\sum \softout \circ \mathbf{a} = \sum \mathbf{a}\), \(\circ\) is the element-wise multiplication, and \(\mathbf{a}\) the edges weights. Similarly to~\citep{Geisler2020}, we recover the message passing operation of a GCN~\cite{Kipf2017} for \(T \to \infty\).

\textbf{Empirical robustness.} The optimal breakdown point only assesses worst-case perturbations. Therefore, in \autoref{fig:empbiascurve}, we analyze the \(L_2\) distance in the latent space after the first message passing operation for a clean vs.\ perturbed graph. Empirically the Soft Median has a 20\% lower error than the weighted sum of a GCN (we call it sum since the weights do not sum up to 1). While here the Soft Medoid seems to be more robust, this is not consistent with the adversarial accuracy values in \autoref{sec:empirical}. Interestingly, the Soft Median can outperform its hard equivalent in terms of the finite error as we show in \autoref{sec:appendix_empirical_error}.\looseness=-1

\textbf{Limitations.} Our Soft Median has the best possible breakdown point which is a (well-established) indicator for robustness but does not prove adversarial robustness. As for most defenses, ours can provide a false sense of robustness. If possible, use attacks and certification techniques to verify the application-specific efficacy. Similar to the Soft Medoid in~\citep{Schuchardt2021, Geisler2020}, we show that the Soft Median can improve the certified robustness in \autoref{sec:appendix_proovable_robustness}. Naturally, our Soft Median also comes with higher cost than e.g.\ a na\"ive summation despite having the same asymptotic complexity. Nevertheless, the overhead seems to be reasonable as we show in our experiments and in combination with PPRGo one can mitigate the slightly higher memory requirements with smaller batch size.\looseness=-1

\section{Empirical Evaluation}\label{sec:empirical}

\begin{wraptable}[13]{r}{0.6\textwidth}
  \vspace{-11pt}
  \centering
  \caption{Dataset summary. For the dense adjacency matrix we assume 4 bytes per entry. We represent the sparse (COO) matrix via two 8 byte integer pointers and a 4 bytes float value per edge. We highlight configurations above 30 GB.}
  \label{tab:datasets}
  \resizebox{\linewidth}{!}{
    \begin{tabular}{lrrrrr}
    \toprule
    \textbf{Dataset} & \textbf{\#Nodes $n$} & \textbf{Size (dense)} & \textbf{Size (sparse)} \\
    \midrule
    Cora ML~\citep{Bojchevski2018} &            2.8 k &              35.88 MB &              168.32 kB \\
    Citeseer~\citep{McCallum2000}  &            3.3 k &              43.88 MB &               94.30 kB \\
    PubMed~\citep{Sen2008}         &           19.7 k &               1.56 GB &                1.77 MB \\
    arXiv~\citep{Hu2020}           &          169.3 k &             \textbf{114.71 GB} &               23.32 MB \\
    Products~\citep{Hu2020}        &            2.4 M &              \textbf{23.99 TB} &                2.47 GB \\
    Papers 100M~\citep{Hu2020}     &          111.1 M &              \textbf{49.34 PB} &               \textbf{32.31 GB} \\
    \bottomrule
    \end{tabular}
  }
\end{wraptable}
In the following, we present our experiments (consisting of approx.\ 2,500 runs) to show the efficacy and scalability of our methods on the six graphs detailed in \autoref{tab:datasets}. \textbf{Attacks:} We benchmark \textbf{our} \underline{GR-BCD} and \underline{PR-BCD} against \emph{PGD}~\citep{Xu2019a}, \emph{greedy FGSM} (similar to~\citet{Dai2018}) as well as \emph{DICE}~\citep{Waniek2018}. \textbf{Defenses:} Besides regular/vanilla GNNs we compare \textbf{our} \underline{Soft Median GDC/PPRGo} with \emph{Soft Medoid GDC}~\cite{Geisler2020}, \emph{SVD GCN}~\citep{Entezari2020}, \emph{RGCN}~\citep{Zhu2019}, and \emph{Jaccard GCN}~\citep{Wu2019}. For the Soft Median, we follow Soft Medoid GDC~\cite{Geisler2020} and diffuse the adjacency matrix with PPR/GDC~\citep{Klicpera2019a} and use PPRGo's efficient implementation to calculate the PPR scores. For the OGB datasets we use the public splits and otherwise sample 20 nodes per class for training/validation. We typically report the average over three random seeds/splits and the 3-sigma error of the mean. The full setup and details about baselines are given in \autoref{sec:appendix_empiricalsetup}.For supplementary material including the code and configuration see \url{https://www.in.tum.de/daml/robustness-of-gnns-at-scale}.

\begin{figure}[b!]
\centering
\hbox{\resizebox{\linewidth}{!}{
\begingroup%
\makeatletter%
\begin{pgfpicture}%
\pgfpathrectangle{\pgfpointorigin}{\pgfqpoint{6.247184in}{0.388266in}}%
\pgfusepath{use as bounding box, clip}%
\begin{pgfscope}%
\pgfsetbuttcap%
\pgfsetmiterjoin%
\definecolor{currentfill}{rgb}{1.000000,1.000000,1.000000}%
\pgfsetfillcolor{currentfill}%
\pgfsetlinewidth{0.000000pt}%
\definecolor{currentstroke}{rgb}{1.000000,1.000000,1.000000}%
\pgfsetstrokecolor{currentstroke}%
\pgfsetstrokeopacity{0.000000}%
\pgfsetdash{}{0pt}%
\pgfpathmoveto{\pgfqpoint{0.000000in}{0.000000in}}%
\pgfpathlineto{\pgfqpoint{6.247184in}{0.000000in}}%
\pgfpathlineto{\pgfqpoint{6.247184in}{0.388266in}}%
\pgfpathlineto{\pgfqpoint{0.000000in}{0.388266in}}%
\pgfpathclose%
\pgfusepath{fill}%
\end{pgfscope}%
\begin{pgfscope}%
\pgfsetbuttcap%
\pgfsetmiterjoin%
\definecolor{currentfill}{rgb}{1.000000,1.000000,1.000000}%
\pgfsetfillcolor{currentfill}%
\pgfsetfillopacity{0.800000}%
\pgfsetlinewidth{1.003750pt}%
\definecolor{currentstroke}{rgb}{0.800000,0.800000,0.800000}%
\pgfsetstrokecolor{currentstroke}%
\pgfsetstrokeopacity{0.800000}%
\pgfsetdash{}{0pt}%
\pgfpathmoveto{\pgfqpoint{0.122222in}{0.100000in}}%
\pgfpathlineto{\pgfqpoint{6.124962in}{0.100000in}}%
\pgfpathquadraticcurveto{\pgfqpoint{6.147184in}{0.100000in}}{\pgfqpoint{6.147184in}{0.122222in}}%
\pgfpathlineto{\pgfqpoint{6.147184in}{0.266044in}}%
\pgfpathquadraticcurveto{\pgfqpoint{6.147184in}{0.288266in}}{\pgfqpoint{6.124962in}{0.288266in}}%
\pgfpathlineto{\pgfqpoint{0.122222in}{0.288266in}}%
\pgfpathquadraticcurveto{\pgfqpoint{0.100000in}{0.288266in}}{\pgfqpoint{0.100000in}{0.266044in}}%
\pgfpathlineto{\pgfqpoint{0.100000in}{0.122222in}}%
\pgfpathquadraticcurveto{\pgfqpoint{0.100000in}{0.100000in}}{\pgfqpoint{0.122222in}{0.100000in}}%
\pgfpathclose%
\pgfusepath{stroke,fill}%
\end{pgfscope}%
\begin{pgfscope}%
\pgfsetbuttcap%
\pgfsetroundjoin%
\pgfsetlinewidth{1.003750pt}%
\definecolor{currentstroke}{rgb}{0.298039,0.447059,0.690196}%
\pgfsetstrokecolor{currentstroke}%
\pgfsetdash{}{0pt}%
\pgfpathmoveto{\pgfqpoint{0.255556in}{0.149378in}}%
\pgfpathlineto{\pgfqpoint{0.255556in}{0.260489in}}%
\pgfusepath{stroke}%
\end{pgfscope}%
\begin{pgfscope}%
\pgfsetroundcap%
\pgfsetroundjoin%
\pgfsetlinewidth{1.003750pt}%
\definecolor{currentstroke}{rgb}{0.298039,0.447059,0.690196}%
\pgfsetstrokecolor{currentstroke}%
\pgfsetdash{}{0pt}%
\pgfpathmoveto{\pgfqpoint{0.144444in}{0.204933in}}%
\pgfpathlineto{\pgfqpoint{0.366667in}{0.204933in}}%
\pgfusepath{stroke}%
\end{pgfscope}%
\begin{pgfscope}%
\definecolor{textcolor}{rgb}{0.150000,0.150000,0.150000}%
\pgfsetstrokecolor{textcolor}%
\pgfsetfillcolor{textcolor}%
\pgftext[x=0.455556in,y=0.166044in,left,base]{\color{textcolor}\rmfamily\fontsize{8.000000}{9.600000}\selectfont CE}%
\end{pgfscope}%
\begin{pgfscope}%
\pgfsetbuttcap%
\pgfsetroundjoin%
\pgfsetlinewidth{1.003750pt}%
\definecolor{currentstroke}{rgb}{0.866667,0.517647,0.321569}%
\pgfsetstrokecolor{currentstroke}%
\pgfsetdash{}{0pt}%
\pgfpathmoveto{\pgfqpoint{0.954396in}{0.149378in}}%
\pgfpathlineto{\pgfqpoint{0.954396in}{0.260489in}}%
\pgfusepath{stroke}%
\end{pgfscope}%
\begin{pgfscope}%
\pgfsetroundcap%
\pgfsetroundjoin%
\pgfsetlinewidth{1.003750pt}%
\definecolor{currentstroke}{rgb}{0.866667,0.517647,0.321569}%
\pgfsetstrokecolor{currentstroke}%
\pgfsetdash{}{0pt}%
\pgfpathmoveto{\pgfqpoint{0.843285in}{0.204933in}}%
\pgfpathlineto{\pgfqpoint{1.065507in}{0.204933in}}%
\pgfusepath{stroke}%
\end{pgfscope}%
\begin{pgfscope}%
\definecolor{textcolor}{rgb}{0.150000,0.150000,0.150000}%
\pgfsetstrokecolor{textcolor}%
\pgfsetfillcolor{textcolor}%
\pgftext[x=1.154396in,y=0.166044in,left,base]{\color{textcolor}\rmfamily\fontsize{8.000000}{9.600000}\selectfont margin}%
\end{pgfscope}%
\begin{pgfscope}%
\pgfsetbuttcap%
\pgfsetroundjoin%
\pgfsetlinewidth{1.003750pt}%
\definecolor{currentstroke}{rgb}{0.333333,0.658824,0.407843}%
\pgfsetstrokecolor{currentstroke}%
\pgfsetdash{}{0pt}%
\pgfpathmoveto{\pgfqpoint{1.848463in}{0.149378in}}%
\pgfpathlineto{\pgfqpoint{1.848463in}{0.260489in}}%
\pgfusepath{stroke}%
\end{pgfscope}%
\begin{pgfscope}%
\pgfsetroundcap%
\pgfsetroundjoin%
\pgfsetlinewidth{1.003750pt}%
\definecolor{currentstroke}{rgb}{0.333333,0.658824,0.407843}%
\pgfsetstrokecolor{currentstroke}%
\pgfsetdash{}{0pt}%
\pgfpathmoveto{\pgfqpoint{1.737352in}{0.204933in}}%
\pgfpathlineto{\pgfqpoint{1.959574in}{0.204933in}}%
\pgfusepath{stroke}%
\end{pgfscope}%
\begin{pgfscope}%
\definecolor{textcolor}{rgb}{0.150000,0.150000,0.150000}%
\pgfsetstrokecolor{textcolor}%
\pgfsetfillcolor{textcolor}%
\pgftext[x=2.048463in,y=0.166044in,left,base]{\color{textcolor}\rmfamily\fontsize{8.000000}{9.600000}\selectfont CW}%
\end{pgfscope}%
\begin{pgfscope}%
\pgfsetbuttcap%
\pgfsetroundjoin%
\pgfsetlinewidth{1.003750pt}%
\definecolor{currentstroke}{rgb}{0.768627,0.305882,0.321569}%
\pgfsetstrokecolor{currentstroke}%
\pgfsetdash{}{0pt}%
\pgfpathmoveto{\pgfqpoint{2.588228in}{0.149378in}}%
\pgfpathlineto{\pgfqpoint{2.588228in}{0.260489in}}%
\pgfusepath{stroke}%
\end{pgfscope}%
\begin{pgfscope}%
\pgfsetroundcap%
\pgfsetroundjoin%
\pgfsetlinewidth{1.003750pt}%
\definecolor{currentstroke}{rgb}{0.768627,0.305882,0.321569}%
\pgfsetstrokecolor{currentstroke}%
\pgfsetdash{}{0pt}%
\pgfpathmoveto{\pgfqpoint{2.477117in}{0.204933in}}%
\pgfpathlineto{\pgfqpoint{2.699339in}{0.204933in}}%
\pgfusepath{stroke}%
\end{pgfscope}%
\begin{pgfscope}%
\definecolor{textcolor}{rgb}{0.150000,0.150000,0.150000}%
\pgfsetstrokecolor{textcolor}%
\pgfsetfillcolor{textcolor}%
\pgftext[x=2.788228in,y=0.166044in,left,base]{\color{textcolor}\rmfamily\fontsize{8.000000}{9.600000}\selectfont NCE}%
\end{pgfscope}%
\begin{pgfscope}%
\pgfsetbuttcap%
\pgfsetroundjoin%
\pgfsetlinewidth{1.003750pt}%
\definecolor{currentstroke}{rgb}{0.505882,0.447059,0.701961}%
\pgfsetstrokecolor{currentstroke}%
\pgfsetdash{}{0pt}%
\pgfpathmoveto{\pgfqpoint{3.375473in}{0.149378in}}%
\pgfpathlineto{\pgfqpoint{3.375473in}{0.260489in}}%
\pgfusepath{stroke}%
\end{pgfscope}%
\begin{pgfscope}%
\pgfsetroundcap%
\pgfsetroundjoin%
\pgfsetlinewidth{1.003750pt}%
\definecolor{currentstroke}{rgb}{0.505882,0.447059,0.701961}%
\pgfsetstrokecolor{currentstroke}%
\pgfsetdash{}{0pt}%
\pgfpathmoveto{\pgfqpoint{3.264362in}{0.204933in}}%
\pgfpathlineto{\pgfqpoint{3.486584in}{0.204933in}}%
\pgfusepath{stroke}%
\end{pgfscope}%
\begin{pgfscope}%
\definecolor{textcolor}{rgb}{0.150000,0.150000,0.150000}%
\pgfsetstrokecolor{textcolor}%
\pgfsetfillcolor{textcolor}%
\pgftext[x=3.575473in,y=0.166044in,left,base]{\color{textcolor}\rmfamily\fontsize{8.000000}{9.600000}\selectfont elu margin}%
\end{pgfscope}%
\begin{pgfscope}%
\pgfsetbuttcap%
\pgfsetroundjoin%
\pgfsetlinewidth{1.003750pt}%
\definecolor{currentstroke}{rgb}{0.576471,0.470588,0.376471}%
\pgfsetstrokecolor{currentstroke}%
\pgfsetdash{}{0pt}%
\pgfpathmoveto{\pgfqpoint{4.459694in}{0.149378in}}%
\pgfpathlineto{\pgfqpoint{4.459694in}{0.260489in}}%
\pgfusepath{stroke}%
\end{pgfscope}%
\begin{pgfscope}%
\pgfsetroundcap%
\pgfsetroundjoin%
\pgfsetlinewidth{1.003750pt}%
\definecolor{currentstroke}{rgb}{0.576471,0.470588,0.376471}%
\pgfsetstrokecolor{currentstroke}%
\pgfsetdash{}{0pt}%
\pgfpathmoveto{\pgfqpoint{4.348583in}{0.204933in}}%
\pgfpathlineto{\pgfqpoint{4.570805in}{0.204933in}}%
\pgfusepath{stroke}%
\end{pgfscope}%
\begin{pgfscope}%
\definecolor{textcolor}{rgb}{0.150000,0.150000,0.150000}%
\pgfsetstrokecolor{textcolor}%
\pgfsetfillcolor{textcolor}%
\pgftext[x=4.659694in,y=0.166044in,left,base]{\color{textcolor}\rmfamily\fontsize{8.000000}{9.600000}\selectfont MCE}%
\end{pgfscope}%
\begin{pgfscope}%
\pgfsetbuttcap%
\pgfsetroundjoin%
\pgfsetlinewidth{1.003750pt}%
\definecolor{currentstroke}{rgb}{0.854902,0.545098,0.764706}%
\pgfsetstrokecolor{currentstroke}%
\pgfsetdash{}{0pt}%
\pgfpathmoveto{\pgfqpoint{5.266610in}{0.149378in}}%
\pgfpathlineto{\pgfqpoint{5.266610in}{0.260489in}}%
\pgfusepath{stroke}%
\end{pgfscope}%
\begin{pgfscope}%
\pgfsetroundcap%
\pgfsetroundjoin%
\pgfsetlinewidth{1.003750pt}%
\definecolor{currentstroke}{rgb}{0.854902,0.545098,0.764706}%
\pgfsetstrokecolor{currentstroke}%
\pgfsetdash{}{0pt}%
\pgfpathmoveto{\pgfqpoint{5.155499in}{0.204933in}}%
\pgfpathlineto{\pgfqpoint{5.377721in}{0.204933in}}%
\pgfusepath{stroke}%
\end{pgfscope}%
\begin{pgfscope}%
\definecolor{textcolor}{rgb}{0.150000,0.150000,0.150000}%
\pgfsetstrokecolor{textcolor}%
\pgfsetfillcolor{textcolor}%
\pgftext[x=5.466610in,y=0.166044in,left,base]{\color{textcolor}\rmfamily\fontsize{8.000000}{9.600000}\selectfont tanh margin}%
\end{pgfscope}%
\end{pgfpicture}%
\makeatother%
\endgroup%
}}
\vspace{-13pt}
\begin{minipage}{.52\textwidth}
  \vspace{12pt}
  \resizebox{\linewidth}{!}{\input{assets/compare_losses_binary_nl.pgf}}
  \caption{Losses for the binary case. The losses are grouped via their basic properties (see text).\label{fig:losses}}
\end{minipage}
\hspace{0.05cm}
\begin{minipage}{.46\textwidth}
  \centering
  \makebox[\linewidth][c]{
    \(\arraycolsep=1pt\def\arraystretch{2}\begin{array}{ccc}
      \subfloat[GR-BCD]{\resizebox{0.475\linewidth}{!}{\input{assets/global_transfer_GreedyRBCD_pubmed_VanillaGCN_surrloss_no_legend.pgf}}} &
      \subfloat[PR-BCD]{\resizebox{0.475\linewidth}{!}{\input{assets/global_transfer_PRBCD_pubmed_VanillaGCN_surrloss_no_legend.pgf}}} \\
    \end{array}\)
  }
  \caption{Attacking GCN on Pubmed. The lower the adv.\ accuracy the better the loss.\label{fig:empsurrogate}}
\end{minipage}
\end{figure}

\textbf{Time and memory cost.} We want to stress again that most of the baselines barely scale to PubMed using a common 11GB GeForce GTX 1080 Ti (as we do). We only use a 32GB Tesla V100 for the experiments on Products with a full-batch GNN, since a three-layer GCN requires roughly 30 GB already during training. Extrapolating the overhead on PubMed to the largest dataset, Papers 100M, \emph{traditional attacks and defenses would require roughly 1 exabyte ($10^{18}$ bytes) while for ours 11 GB suffice}. Our attacks and defenses are also fast. On arXiv (170 k nodes), we train for 500 epochs and run the global PR-BCD attack for 500 epochs. The whole training and attacking procedure requires less than 2 minutes. Moreover, one epoch on Papers 100M with the local PR-BCD attack takes less than 10 seconds. See \autoref{sec:appendix_timememorycost} for further details and \autoref{sec:appendix_theoretical_complexities} for theoretical complexities. %

\textbf{Surrogate Loss.} We illustrate the losses in \autoref{fig:losses}, where we clustered the losses in three groups. (1) incentivizing low margins: Cross Entropy CE and \(\text{margin}\). (2) focusing on high-confidence nodes: Carlini-Wagner CW, the (neg.) CE of the most-likely, non-target class \(\text{NCE}\), and \(\text{ELU Margin}\). (3) focusing on nodes close 
to the decision boundary: \underline{\(\text{MCE}\)} \textbf{(ours)} and \underline{\(\text{tanh margin}\)} \textbf{(ours)}. In \autoref{fig:empsurrogate}, we see that the losses of category (3), or equivalently obeying the properties of \autoref{definition:budgetaware} or \autoref{definition:confidentandbudgetaware}, are superior to the other losses. For example, with \(\text{MCE}\) and FGSM the accuracy drops twice as much as with CE. A detailed discussion and mathematical
formulation of all losses can be found in \autoref{sec:appendix_ceisbad}. Additionally, we report further experiments backing our claims and discuss the losses' properties in more detail. Subsequently, we use \(\text{MCE}\) for greedy attacks and \(\text{tanh margin}\) otherwise.

\begin{table}[t!]
\centering
\caption{Comparing attacks (transfer from Vanilla GCN) and defenses. We show the adversarial accuracy for \(\epsilon=0.1\) on Cora ML, and the clean test accuracy (last column).
We only highlight the \textbf{strongest defense} as the attacks perform similarly. \underline{Our approaches} are underlined. See \autoref{sec:appendix_global} for more datasets, budgets, and adaptive/direct attacks.}
\label{tab:global_small}
\vskip 0.11in
\resizebox{0.925\linewidth}{!}{
\begin{tabular}{llccccc}
\toprule
    \textbf{Attack} & \textbf{FGSM} & \underline{\textbf{GR-BCD}} &    \textbf{PGD} & \underline{\textbf{PR-BCD}} &   \textbf{Acc.} \\
    \midrule
    \underline{Soft Median GDC} &           0.769 $\pm$ 0.002 &           0.765 $\pm$ 0.001 &           0.758 $\pm$ 0.002 &           0.752 $\pm$ 0.002 &           0.824 $\pm$ 0.002 \\
    \underline{Soft Median PPRGo} &  \textbf{0.778 $\pm$ 0.001} &  \textbf{0.781 $\pm$ 0.002} &  \textbf{0.769 $\pm$ 0.001} &  \textbf{0.770 $\pm$ 0.001} &           0.821 $\pm$ 0.001 \\
    Vanilla GCN &           0.641 $\pm$ 0.003 &           0.622 $\pm$ 0.003 &           0.662 $\pm$ 0.003 &           0.645 $\pm$ 0.002 &           0.827 $\pm$ 0.003 \\
    Vanilla GDC &           0.672 $\pm$ 0.005 &           0.677 $\pm$ 0.005 &           0.679 $\pm$ 0.002 &           0.674 $\pm$ 0.004 &  \textbf{0.842 $\pm$ 0.003} \\
    Vanilla PPRGo &           0.724 $\pm$ 0.003 &           0.726 $\pm$ 0.002 &           0.704 $\pm$ 0.001 &           0.700 $\pm$ 0.002 &           0.826 $\pm$ 0.002 \\
    Soft Medoid GDC &           0.773 $\pm$ 0.005 &           0.775 $\pm$ 0.003 &           0.759 $\pm$ 0.003 &           0.761 $\pm$ 0.003 &           0.819 $\pm$ 0.002 \\
    SVD GCN &           0.751 $\pm$ 0.007 &           0.755 $\pm$ 0.006 &           0.719 $\pm$ 0.005 &           0.724 $\pm$ 0.006 &           0.781 $\pm$ 0.005 \\
    Jaccard GCN &           0.661 $\pm$ 0.002 &           0.664 $\pm$ 0.001 &           0.673 $\pm$ 0.002 &           0.667 $\pm$ 0.003 &           0.818 $\pm$ 0.003 \\
    RGCN &           0.654 $\pm$ 0.007 &           0.665 $\pm$ 0.005 &           0.671 $\pm$ 0.007 &           0.664 $\pm$ 0.004 &           0.819 $\pm$ 0.002 \\
    \bottomrule
    \end{tabular}
}
\end{table}

\begin{figure}[t!]
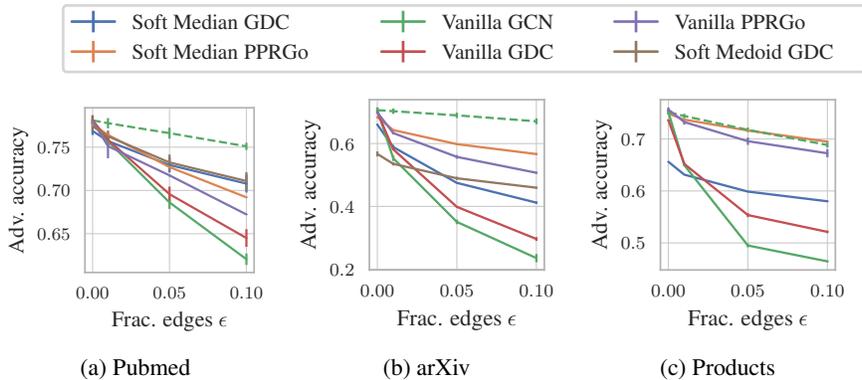

  \centering
  \hbox{\hspace{50pt}\resizebox{0.8\linewidth}{!}{
\begingroup%
\makeatletter%
\begin{pgfpicture}%
\pgfpathrectangle{\pgfpointorigin}{\pgfqpoint{4.335479in}{0.543199in}}%
\pgfusepath{use as bounding box, clip}%
\begin{pgfscope}%
\pgfsetbuttcap%
\pgfsetmiterjoin%
\definecolor{currentfill}{rgb}{1.000000,1.000000,1.000000}%
\pgfsetfillcolor{currentfill}%
\pgfsetlinewidth{0.000000pt}%
\definecolor{currentstroke}{rgb}{1.000000,1.000000,1.000000}%
\pgfsetstrokecolor{currentstroke}%
\pgfsetstrokeopacity{0.000000}%
\pgfsetdash{}{0pt}%
\pgfpathmoveto{\pgfqpoint{0.000000in}{0.000000in}}%
\pgfpathlineto{\pgfqpoint{4.335479in}{0.000000in}}%
\pgfpathlineto{\pgfqpoint{4.335479in}{0.543199in}}%
\pgfpathlineto{\pgfqpoint{0.000000in}{0.543199in}}%
\pgfpathclose%
\pgfusepath{fill}%
\end{pgfscope}%
\begin{pgfscope}%
\pgfsetbuttcap%
\pgfsetmiterjoin%
\definecolor{currentfill}{rgb}{1.000000,1.000000,1.000000}%
\pgfsetfillcolor{currentfill}%
\pgfsetfillopacity{0.800000}%
\pgfsetlinewidth{1.003750pt}%
\definecolor{currentstroke}{rgb}{0.800000,0.800000,0.800000}%
\pgfsetstrokecolor{currentstroke}%
\pgfsetstrokeopacity{0.800000}%
\pgfsetdash{}{0pt}%
\pgfpathmoveto{\pgfqpoint{0.122222in}{0.100000in}}%
\pgfpathlineto{\pgfqpoint{4.213256in}{0.100000in}}%
\pgfpathquadraticcurveto{\pgfqpoint{4.235479in}{0.100000in}}{\pgfqpoint{4.235479in}{0.122222in}}%
\pgfpathlineto{\pgfqpoint{4.235479in}{0.420977in}}%
\pgfpathquadraticcurveto{\pgfqpoint{4.235479in}{0.443199in}}{\pgfqpoint{4.213256in}{0.443199in}}%
\pgfpathlineto{\pgfqpoint{0.122222in}{0.443199in}}%
\pgfpathquadraticcurveto{\pgfqpoint{0.100000in}{0.443199in}}{\pgfqpoint{0.100000in}{0.420977in}}%
\pgfpathlineto{\pgfqpoint{0.100000in}{0.122222in}}%
\pgfpathquadraticcurveto{\pgfqpoint{0.100000in}{0.100000in}}{\pgfqpoint{0.122222in}{0.100000in}}%
\pgfpathclose%
\pgfusepath{stroke,fill}%
\end{pgfscope}%
\begin{pgfscope}%
\pgfsetbuttcap%
\pgfsetroundjoin%
\pgfsetlinewidth{1.003750pt}%
\definecolor{currentstroke}{rgb}{0.298039,0.447059,0.690196}%
\pgfsetstrokecolor{currentstroke}%
\pgfsetdash{}{0pt}%
\pgfpathmoveto{\pgfqpoint{0.255556in}{0.304311in}}%
\pgfpathlineto{\pgfqpoint{0.255556in}{0.415422in}}%
\pgfusepath{stroke}%
\end{pgfscope}%
\begin{pgfscope}%
\pgfsetroundcap%
\pgfsetroundjoin%
\pgfsetlinewidth{1.003750pt}%
\definecolor{currentstroke}{rgb}{0.298039,0.447059,0.690196}%
\pgfsetstrokecolor{currentstroke}%
\pgfsetdash{}{0pt}%
\pgfpathmoveto{\pgfqpoint{0.144444in}{0.359866in}}%
\pgfpathlineto{\pgfqpoint{0.366667in}{0.359866in}}%
\pgfusepath{stroke}%
\end{pgfscope}%
\begin{pgfscope}%
\definecolor{textcolor}{rgb}{0.150000,0.150000,0.150000}%
\pgfsetstrokecolor{textcolor}%
\pgfsetfillcolor{textcolor}%
\pgftext[x=0.455556in,y=0.320977in,left,base]{\color{textcolor}\rmfamily\fontsize{8.000000}{9.600000}\selectfont Soft Median GDC}%
\end{pgfscope}%
\begin{pgfscope}%
\pgfsetbuttcap%
\pgfsetroundjoin%
\pgfsetlinewidth{1.003750pt}%
\definecolor{currentstroke}{rgb}{0.866667,0.517647,0.321569}%
\pgfsetstrokecolor{currentstroke}%
\pgfsetdash{}{0pt}%
\pgfpathmoveto{\pgfqpoint{0.255556in}{0.149378in}}%
\pgfpathlineto{\pgfqpoint{0.255556in}{0.260489in}}%
\pgfusepath{stroke}%
\end{pgfscope}%
\begin{pgfscope}%
\pgfsetroundcap%
\pgfsetroundjoin%
\pgfsetlinewidth{1.003750pt}%
\definecolor{currentstroke}{rgb}{0.866667,0.517647,0.321569}%
\pgfsetstrokecolor{currentstroke}%
\pgfsetdash{}{0pt}%
\pgfpathmoveto{\pgfqpoint{0.144444in}{0.204933in}}%
\pgfpathlineto{\pgfqpoint{0.366667in}{0.204933in}}%
\pgfusepath{stroke}%
\end{pgfscope}%
\begin{pgfscope}%
\definecolor{textcolor}{rgb}{0.150000,0.150000,0.150000}%
\pgfsetstrokecolor{textcolor}%
\pgfsetfillcolor{textcolor}%
\pgftext[x=0.455556in,y=0.166044in,left,base]{\color{textcolor}\rmfamily\fontsize{8.000000}{9.600000}\selectfont Soft Median PPRGo}%
\end{pgfscope}%
\begin{pgfscope}%
\pgfsetbuttcap%
\pgfsetroundjoin%
\pgfsetlinewidth{1.003750pt}%
\definecolor{currentstroke}{rgb}{0.333333,0.658824,0.407843}%
\pgfsetstrokecolor{currentstroke}%
\pgfsetdash{}{0pt}%
\pgfpathmoveto{\pgfqpoint{1.853270in}{0.304311in}}%
\pgfpathlineto{\pgfqpoint{1.853270in}{0.415422in}}%
\pgfusepath{stroke}%
\end{pgfscope}%
\begin{pgfscope}%
\pgfsetroundcap%
\pgfsetroundjoin%
\pgfsetlinewidth{1.003750pt}%
\definecolor{currentstroke}{rgb}{0.333333,0.658824,0.407843}%
\pgfsetstrokecolor{currentstroke}%
\pgfsetdash{}{0pt}%
\pgfpathmoveto{\pgfqpoint{1.742159in}{0.359866in}}%
\pgfpathlineto{\pgfqpoint{1.964381in}{0.359866in}}%
\pgfusepath{stroke}%
\end{pgfscope}%
\begin{pgfscope}%
\definecolor{textcolor}{rgb}{0.150000,0.150000,0.150000}%
\pgfsetstrokecolor{textcolor}%
\pgfsetfillcolor{textcolor}%
\pgftext[x=2.053270in,y=0.320977in,left,base]{\color{textcolor}\rmfamily\fontsize{8.000000}{9.600000}\selectfont Vanilla GCN}%
\end{pgfscope}%
\begin{pgfscope}%
\pgfsetbuttcap%
\pgfsetroundjoin%
\pgfsetlinewidth{1.003750pt}%
\definecolor{currentstroke}{rgb}{0.768627,0.305882,0.321569}%
\pgfsetstrokecolor{currentstroke}%
\pgfsetdash{}{0pt}%
\pgfpathmoveto{\pgfqpoint{1.853270in}{0.149378in}}%
\pgfpathlineto{\pgfqpoint{1.853270in}{0.260489in}}%
\pgfusepath{stroke}%
\end{pgfscope}%
\begin{pgfscope}%
\pgfsetroundcap%
\pgfsetroundjoin%
\pgfsetlinewidth{1.003750pt}%
\definecolor{currentstroke}{rgb}{0.768627,0.305882,0.321569}%
\pgfsetstrokecolor{currentstroke}%
\pgfsetdash{}{0pt}%
\pgfpathmoveto{\pgfqpoint{1.742159in}{0.204933in}}%
\pgfpathlineto{\pgfqpoint{1.964381in}{0.204933in}}%
\pgfusepath{stroke}%
\end{pgfscope}%
\begin{pgfscope}%
\definecolor{textcolor}{rgb}{0.150000,0.150000,0.150000}%
\pgfsetstrokecolor{textcolor}%
\pgfsetfillcolor{textcolor}%
\pgftext[x=2.053270in,y=0.166044in,left,base]{\color{textcolor}\rmfamily\fontsize{8.000000}{9.600000}\selectfont Vanilla GDC}%
\end{pgfscope}%
\begin{pgfscope}%
\pgfsetbuttcap%
\pgfsetroundjoin%
\pgfsetlinewidth{1.003750pt}%
\definecolor{currentstroke}{rgb}{0.505882,0.447059,0.701961}%
\pgfsetstrokecolor{currentstroke}%
\pgfsetdash{}{0pt}%
\pgfpathmoveto{\pgfqpoint{3.054399in}{0.304311in}}%
\pgfpathlineto{\pgfqpoint{3.054399in}{0.415422in}}%
\pgfusepath{stroke}%
\end{pgfscope}%
\begin{pgfscope}%
\pgfsetroundcap%
\pgfsetroundjoin%
\pgfsetlinewidth{1.003750pt}%
\definecolor{currentstroke}{rgb}{0.505882,0.447059,0.701961}%
\pgfsetstrokecolor{currentstroke}%
\pgfsetdash{}{0pt}%
\pgfpathmoveto{\pgfqpoint{2.943288in}{0.359866in}}%
\pgfpathlineto{\pgfqpoint{3.165511in}{0.359866in}}%
\pgfusepath{stroke}%
\end{pgfscope}%
\begin{pgfscope}%
\definecolor{textcolor}{rgb}{0.150000,0.150000,0.150000}%
\pgfsetstrokecolor{textcolor}%
\pgfsetfillcolor{textcolor}%
\pgftext[x=3.254399in,y=0.320977in,left,base]{\color{textcolor}\rmfamily\fontsize{8.000000}{9.600000}\selectfont Vanilla PPRGo}%
\end{pgfscope}%
\begin{pgfscope}%
\pgfsetbuttcap%
\pgfsetroundjoin%
\pgfsetlinewidth{1.003750pt}%
\definecolor{currentstroke}{rgb}{0.576471,0.470588,0.376471}%
\pgfsetstrokecolor{currentstroke}%
\pgfsetdash{}{0pt}%
\pgfpathmoveto{\pgfqpoint{3.054399in}{0.149378in}}%
\pgfpathlineto{\pgfqpoint{3.054399in}{0.260489in}}%
\pgfusepath{stroke}%
\end{pgfscope}%
\begin{pgfscope}%
\pgfsetroundcap%
\pgfsetroundjoin%
\pgfsetlinewidth{1.003750pt}%
\definecolor{currentstroke}{rgb}{0.576471,0.470588,0.376471}%
\pgfsetstrokecolor{currentstroke}%
\pgfsetdash{}{0pt}%
\pgfpathmoveto{\pgfqpoint{2.943288in}{0.204933in}}%
\pgfpathlineto{\pgfqpoint{3.165511in}{0.204933in}}%
\pgfusepath{stroke}%
\end{pgfscope}%
\begin{pgfscope}%
\definecolor{textcolor}{rgb}{0.150000,0.150000,0.150000}%
\pgfsetstrokecolor{textcolor}%
\pgfsetfillcolor{textcolor}%
\pgftext[x=3.254399in,y=0.166044in,left,base]{\color{textcolor}\rmfamily\fontsize{8.000000}{9.600000}\selectfont Soft Medoid GDC}%
\end{pgfscope}%
\end{pgfpicture}%
\makeatother%
\endgroup%
}}
  \vspace{-14pt}
  \begin{minipage}[t]{0.85\linewidth}
    \(\arraycolsep=1pt\def\arraystretch{2}\begin{array}{ccc}
      \subfloat[Pubmed]{\resizebox{0.32\linewidth}{!}{\input{assets/global_transfer_PRBCD_pubmed_pertaccuracy_no_legend.pgf}}} &
      \subfloat[arXiv]{\resizebox{0.32\linewidth}{!}{\input{assets/global_transfer_PRBCD_ogbn-arxiv_pertaccuracy_no_legend.pgf}}} &
      \subfloat[Products]{\resizebox{0.32\linewidth}{!}{\input{assets/global_transfer_PRBCD_ogbn-products_pertaccuracy_no_legend.pgf}}} \\
    \end{array}\)
  \end{minipage}
  \caption{PR-BCD (DICE dashed) on the large datasets (transfer) where the adversarial accuracy denotes the accuracy after attacking with budget \(\Delta = \epsilon m\).\label{fig:prbcdlarge}}
\end{figure}

\textbf{Robustness w.r.t.\ global attacks.} In \autoref{tab:global_small}, we present the experimental results for our proposed global attacks on the small dataset Cora ML since most baselines do not scale much further. Our attacks are as strong as their dense equivalents despite being much more scalable. In \autoref{fig:prbcdlarge}, we compare our PR-BCD attack on all baselines that fit into memory or can be trained within 24 hours on the bigger datasets. On the large products dataset, it suffices to perturb roughly 2\% of the edges to push the accuracy below 60\%, i.e.\ reaching the performance of an MLP~\citep{Hu2020}. We conclude that GNNs on large graphs are indeed not robust (see also \autoref{sec:appendix_graphsize}). Our defense Soft Median GDC and Soft Median PPRGo are consistently among the best models tested over all scales. For example with \(\epsilon=0.1\), the accuracy of a Vanilla GCN drops by an absolute 20\% while for the Soft Median PPRGo we only lose 5\%. To fit our Soft Median GDC on Products into memory, we had to reduce the number
of hidden dimensions in comparison to its baselines. However, note that even a Vanilla GCN requires almost the entire memory of the 32~GB GPU. Despite the small sacrifice in clean accuracy, we already outperform most baselines for a budget of \(\epsilon>0.01\). We also faced similar scaling limitations for the Soft Medoid GDC baseline on arXiv. This highlights the lower memory requirements for our Soft Median. In \autoref{sec:appendix_global}, we present more exhaustive results and adaptive/direct attacks supporting the robustness of our defense but highlighting the importance of adaptiveness.\looseness=-1 

\begin{wrapfigure}[34]{r}{0.45\textwidth}
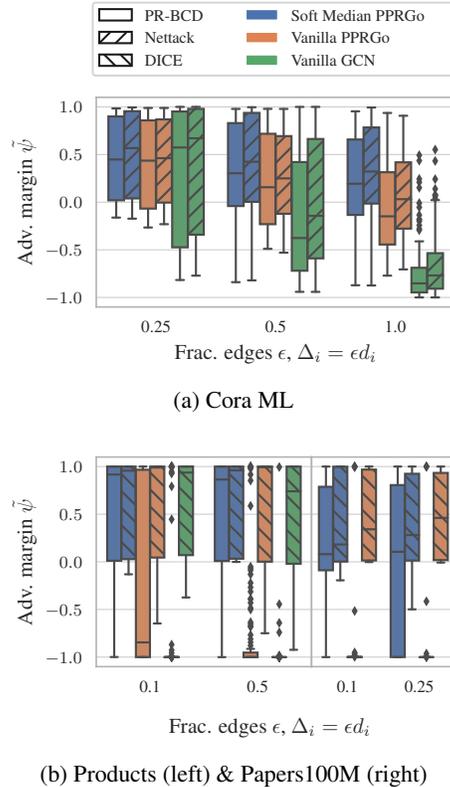

  \vspace{-7pt}
  \centering
  \hbox{\hspace{28pt} \resizebox{0.825\linewidth}{!}{\input{assets/local_prbcd_vs_nettack_cora_ml_boxplmargin_legend.pgf}}}
  \vspace{-14pt}
  \makebox[\linewidth][c]{
    \(\arraycolsep=1pt\def\arraystretch{2}\begin{array}{cc}
      \subfloat[Cora ML]{\resizebox{1\linewidth}{!}{\input{assets/local_prbcd_vs_nettack_cora_ml_boxplmargin_nl.pgf}}} \\
      \subfloat[Products (left) \& Papers100M (right)]{\resizebox{1\linewidth}{!}{\input{assets/local_papers_and_products_boxplmargin.pgf}}} \\
    \end{array}\)
  }
  \caption{Adversarial classification margins \(\tilde{\psi}_i\) of the attacked nodes. In (a), we compare our local PR-BCD attack with Nettack~\cite{Zugner2018} on (undirected) Cora ML. In (b), we show the results on the (directed) large-scale datasets Products (2.5 million nodes) and Papers 100M (111 million nodes), respectively. Our Soft Medoid PPRGo resists the attacks much better than the baselines. \label{fig:emplocal}}
\end{wrapfigure}

\textbf{Robustness w.r.t.\ local attacks.} In \autoref{fig:emplocal}, we compare the results of our local PR-BCD with Nettack on Cora ML (undirected). We define the budget \(\Delta_i = \epsilon d_i\) and select the nodes for each budget s.t.\ \(\Delta_i \ge 1\). Similarly to~\citet{Zugner2018}, we apply the attack only to the 10 nodes with highest confidence, 10 with lowest, and 20 random nodes (all correctly classified). For more datasets and budgets see \autoref{sec:appendix_local}.
Our attack seems to be slightly stronger than Nettack on all architectures and budgets. Nettack and PR-BCD use a different strategy to make the original combinatorial optimization problem feasible (see \autoref{sec:intro}). Nettack uses a linearized surrogate model to select the adversarial edges. Evidently, this leads to a weaker attack compared to relaxing the optimization problem as we proposed with PR-BCD. On the large datasets Products and Papers 100M (directed), we outperform the simple DICE baseline substantially. We compare to DICE since Nettack is not scalable enough. In comparison to the small datasets, the Vanilla GCN/PPRGo are extremely fragile
and much lower budgets \(\Delta_i\) suffice to flip almost every 
node's prediction. Our proposed defense Soft Median PPRGo on the other hand remains similarly robust as on the small datasets.
On Papers 100M with \(\Delta_i=0.25\), the Soft Median PPRGo reduces the attacker's success rate from around 90\% to just 30\% (90\% vs. 1\% on Products with \(\Delta_i=0.5\)).

\begin{table}[t]
\centering
\caption{Attack success rate of \underline{PR-BCD} (ours) and SGA~\citep{li_adversarial_2021} on a Vanilla GCN and Vanilla SGC~\citep{wu_simplifying_2019} (stronger is bold). For poisoning we retrain on the perturbed graph of an evasion attack.\label{tab:compare_sga}}
\resizebox{0.9125\linewidth}{!}{
\begin{tabular}{lcccccccccc}
\toprule
               &             & \textbf{Attack} & \multicolumn{3}{c}{\underline{PR-BCD}} & \multicolumn{3}{c}{SGA} \\
\multicolumn{3}{c}{\textbf{Frac. edges $\epsilon$, $\Delta_i = \epsilon d_i$}} &             0.25 &             0.50 &             1.00 &             0.25 &             0.50 &             1.00 \\
\midrule
\multirow{4}{*}{\rotatebox{90}{\textbf{Cora ML}}} & \multirow{2}{*}{GCN} & evasion &  \textbf{0.38 $\pm$ 0.04} &  \textbf{0.65 $\pm$ 0.04} &  \textbf{0.96 $\pm$ 0.02} &  0.36 $\pm$ 0.04 &  0.51 $\pm$ 0.05 &  0.82 $\pm$ 0.03 \\
               &             & poisoning &  0.46 $\pm$ 0.05 &  \textbf{0.79 $\pm$ 0.04} &  \textbf{0.97 $\pm$ 0.02} &  \textbf{0.47 $\pm$ 0.05} &  0.64 $\pm$ 0.04 &  0.95 $\pm$ 0.02 \\
\cline{2-9}
               & \multirow{2}{*}{SGC} & evasion &  \textbf{0.45 $\pm$ 0.05} &  \textbf{0.57 $\pm$ 0.05} &  \textbf{0.97 $\pm$ 0.02} &  0.37 $\pm$ 0.04 &  0.43 $\pm$ 0.05 &  0.95 $\pm$ 0.02 \\
               &             & poisoning &  \textbf{0.50 $\pm$ 0.05} &  \textbf{0.66 $\pm$ 0.04} &  0.96 $\pm$ 0.02 &  0.47 $\pm$ 0.05 &  0.62 $\pm$ 0.04 &  \textbf{0.97 $\pm$ 0.01} \\
\cline{1-9}
\cline{2-9}
\multirow{4}{*}{\rotatebox{90}{\textbf{Citeseer}}} & \multirow{2}{*}{GCN} & evasion &  \textbf{0.42 $\pm$ 0.05} &  \textbf{0.66 $\pm$ 0.04} &  \textbf{0.85 $\pm$ 0.03} &  0.34 $\pm$ 0.04 &  0.50 $\pm$ 0.05 &  0.72 $\pm$ 0.04 \\
               &             & poisoning &  0.53 $\pm$ 0.05 &  \textbf{0.78 $\pm$ 0.04} &  \textbf{0.94 $\pm$ 0.02} &  \textbf{0.54 $\pm$ 0.05} &  0.76 $\pm$ 0.04 &  0.93 $\pm$ 0.02 \\
\cline{2-9}
               & \multirow{2}{*}{SGC} & evasion &  \textbf{0.39 $\pm$ 0.04} &  \textbf{0.62 $\pm$ 0.04} &  \textbf{0.91 $\pm$ 0.03} &  0.35 $\pm$ 0.04 &  0.55 $\pm$ 0.05 &  0.85 $\pm$ 0.03 \\
               &             & poisoning &  \textbf{0.49 $\pm$ 0.05} &  \textbf{0.77 $\pm$ 0.04} &  0.92 $\pm$ 0.03 &  \textbf{0.49 $\pm$ 0.05} &  0.74 $\pm$ 0.04 &  \textbf{0.97 $\pm$ 0.01} \\
\cline{1-9}
\cline{2-9}
\multirow{4}{*}{\rotatebox{90}{\textbf{arXiv}}} & \multirow{2}{*}{GCN} & evasion &  \textbf{0.92 $\pm$ 0.03} &  \textbf{1.00 $\pm$ 0.00} &  \textbf{1.00 $\pm$ 0.00} &  0.58 $\pm$ 0.05 &  0.90 $\pm$ 0.03 &  0.98 $\pm$ 0.01 \\
               &             & poisoning &  \textbf{0.82 $\pm$ 0.03} &  \textbf{0.99 $\pm$ 0.01} &  \textbf{1.00 $\pm$ 0.00} &  0.52 $\pm$ 0.05 &  0.82 $\pm$ 0.04 &  0.98 $\pm$ 0.01 \\
\cline{2-9}
               & \multirow{2}{*}{SGC} & evasion &  \textbf{0.91 $\pm$ 0.03} &  \textbf{0.97 $\pm$ 0.01} &  \textbf{1.00 $\pm$ 0.00} &  0.83 $\pm$ 0.04 &  0.94 $\pm$ 0.02 &  0.94 $\pm$ 0.02 \\
               &             & poisoning &  \textbf{0.91 $\pm$ 0.03} &  \textbf{0.97 $\pm$ 0.01} &  \textbf{1.00 $\pm$ 0.00} &  0.83 $\pm$ 0.04 &  0.94 $\pm$ 0.02 &  0.94 $\pm$ 0.02 \\
\bottomrule
\end{tabular}
}
\end{table}

In \autoref{tab:compare_sga}, we compare to the SGA attack of \citet{li_adversarial_2021} that transfers the attacks from a SGC~\citep{wu_simplifying_2019} surrogate. With our PR-BCD we attack the respective model directly. We follow SGA and obtain a poisoning attack by applying the perturbations of an evasion attack to the graph before training. Our PR-BCD clearly dominates SGA--even on SGC. This demonstrates how generally applicable our PR-BCD is without any modifications. We hypothesize that PR-BCD is stronger since, in contrast to SGA, it does not constrain the edge perturbations to be within a subgraph. Moreover, the large gap for a GCN highlights the importance of adaptive attacks (i.e.\ no surrogate). Also in terms of scalability, we find PR-BCD to be superior, even though SGA is efficient on graphs up to the size of arXiv. However, on products, we observe that for \(s=3\) SGC message passing steps we sometimes require more than 11 Gb and \(s=4\) we typically require more than 32 Gb. However, our PR-BCD with PPRGo scales to graphs 2 magnitudes larger (Papers100M) and requires less than 11 GB (see \autoref{sec:appendix_timememorycost}).\looseness=-1

\section{Conclusion}\label{sec:conclusion} %

We study the adversarial robustness of GNNs at scale. We tackle all three of the identified challenges: (1) we introduce surrogate losses for global attacks that can double the attack strength, (2) we principally scale first-order attacks that optimize over the quadratic number of possible edges, and (3) we propose a scalable defense using our novel Soft Median which is differentiable as well as provably robust. We show that our attacks and defenses are practical by scaling to graphs of up to 111 million nodes. In some settings our defense reduces the attack's success rate from around 90 \% to 1 \%. Most importantly, our work enables the assessment of robustness for massive-scale applications with GNNs.\looseness=-1

\begin{ack}
This research was supported by the Helmholtz Association under the joint research school “Munich School for Data Science - MUDS“.
\end{ack}

\bibliography{references}

\clearpage
\appendix

\counterwithin{figure}{section}
\counterwithin{table}{section}
\counterwithin{equation}{section}

\section{Notation}\label{sec:appendix_notation}

In this section, we explicitly summarize the notation. We intended to introduce all terms when needed. However, due to the symbiosis of Graph Neural Networks (GNNs), adversarial robustness, adversarial attacks, adversarial defenses, large-scale optimization, and robust statistics we inevitably need a large number of different symbols. We give the important symbols in \autoref{tab:appendix_notation}. Recall that we formulate GNNs as a recursive transformation and aggregation over the features/embedding of the neighboring nodes (with a potentially weighted/normalized adjacency matrix \(\adj\)), i.e.\:
\begin{equation}\label{eq:appendix_messagepassing}
    \mathbf{h}^{(l)}_v = \sigma^{(l)} \left[ \text{AGG}^{(l)} \left \{ \left( \adj_{vu}, \mathbf{h}^{(l-1)}_u \weight^{(l)} \right), \forall \, u\in \neighbors'(v) \right \} \right]
\end{equation}

\textbf{Asymptotics.} To describe the growth of a function \(f(n)\) and similarly the complexity of algorithms, we use \(f(n) = \mathcal{O}(g(n))\) and \(f(n) = \Theta(g(n))\) (here \(n\) does not denote the number of nodes). Roughly speaking, \(f(n) = \mathcal{O}(g(n))\) means that the growth is upper bounded (up to constant factors) and \(f(n) = \Theta(g(n))\) means that \(f(n)\) grows as fast as \(g(n)\). While we do not give a formal definition, we quickly want to recall the well-known facts (that hold under certain conditions which are naturally fulfilled in our applications/analysis):
\begin{equation}
    \lim_{n \to \infty} \frac{f(n)}{g(n)} < \infty \Rightarrow f(n) = \mathcal{O}(g(n))
\end{equation}
\begin{equation}
    0 < \lim_{n \to \infty} \frac{f(n)}{g(n)} < \infty \Rightarrow f(n) = \Theta(g(n))
\end{equation}

\begin{longtable}{p{.20\textwidth}  p{.7\textwidth}}
\caption{Here we list the most important symbols used in this work.} \label{tab:appendix_notation} \\
\toprule
   \(k\)                             & typically used as the threshold in some top \(k\) operation/sparsification \\
   \(d\)                             & number of dimension / features (e.g.\ \(\features\) is of shape \(n \times d\)) \\
   \text{AGG}                        & some aggregation (e.g.\ sum, max, Soft Median, ...) \\
   \(\vu\vv^\top\)                   & two (column) vectors constructing a rank-1 matrix \\

   \(\gG=(\adj, \features)\)         & the (attributed) graph \\
   \(n\)                             & number of nodes \\
   \(m\)                             & number of edges \\
   \(\adj\)                          & (clean) adjacency matrix (shape \(n  \times n\))\\
   \(\mathbf{a}\)                    & weights of a row/neighborhood given by the (weighted) adjacency matrix \\
   \(\features\)                     & features / node attributes as a matrix  (shape \(n \times d\)) \\x
   \(\featset\)                      & features / node attributes as a set \\
   \(\neighbors(i)\)                 & the (direct) neighbors of node \(i\) \\
   \(\mD\)                           & the degree matrix of a graph \\ 
   \(\boldsymbol{\Pi}\)                           & Stationary distribution of a random walk with restarts / Personalized Page Rank (PPR) matrix, here \(\boldsymbol{\Pi} = \alpha(\mI - (1-\alpha)\mD^{-1}\mA)^{-1}\) \\
   \(f_{\theta}(\adj, \features)\)   & Graph Neural Network (GNN) for node classification \\
   \(\theta\)                        & (all) model parameters \\
   \(\weight\)                       & weight matrix (contained in \(\theta\)) \\
   \(\mathbf{h}\)                    & embeddings / hidden state \\
   \(\sigma(\mathbf{h})\)            & some (nonlinear) activation function \\
   \(\text{softmax}(\mathbf{h})\)         & the softmax operation/activation \\
   \(T\)                             & temperature parameter to control the steepness of the softmax \\
   \(\softout\)                      & the weight vector of a softmax operation \\
   \(\vp, \vp_i\)                    & probability/confidence scores predicted for an arbitrary node \(i\): \(\vp = f_{\theta}(\adj, \features)\) \\
   \(\vz, \vz_i\)                    & logits / pre-softmax activation predicted for an arbitrary node \(i\): \(\vz = f_{\theta}(\adj, \features)\) (overloaded notation) \\
   \(y, \evy_i\)                     & label (ground truth) \\
   \(\mathcal{L}\)                   & (target) loss \\
   \(\mathcal{L}_{0/1}\)             & 0/1 loss corresponding to the accuracy \\
   \(\mathcal{L}'\)                  & surrogate loss \\
   \(l\)                             & typically the layer index, e.g.\ \(\sigma^{(l)}(\mathbf{h}^{(l-1)})\) is the \(l\)-th layer activation\\
   \(L\)                             & number of layers \\
   \(\sC\)                           & set of classes \\
   \(c\)                             & we typically use \(c\) while iterating the classes\\
   \(c^*\)                           & denotes the target class / ground truth \\
   \(\sV^+\)                             & set of correctly classified nodes \\
  \(\psi\)                           & classification margin in the confidence space \(\psi = \min_{c \ne c^*} \evp_{c^*} - \evp_{c}\) \\

   \(\tilde{\adj}\)                  & (perturbed) adjacency matrix, e.g.\ during/after an attack \\
   \(\Delta\)                        & budget of an attack \\
   \(\epsilon\)                      & relative budget usually w.r.t.\ the number of edges \(m\), i.e.\ \(\Delta = \epsilon \cdot m\) \\
   \(\alpha\)                        & learning rate \\

   \(t\)                             & index of epoch, i.e.\ \(t \in \{0, \dots, E\}\) \\
   \(E\)                             & number of epochs \\
   \(E_{\text{res.}}\)               & number of epochs with resampling of the random block (see Algo.~\ref{algo:prbcd}) \\
   \(\Pi(\dots)\)                    & A projection in projected gradient descent \\
   \(\mP\)                           & perturbations \(\tilde{\adj} = \adj \oplus \mP\) \\
   \(\vp\)                           & in context of PR-BCD, \(\vp\) corresponds to the current subset/block \(\mP_{\vi_t}\) \\
   \(\vi\)                           & indices representing the current block in GR-BCD/PR-BCD\\
   \(\oplus\)                        & exclusive or operation if inputs are binary. If inputs are floats: \(\adj_{ij} \oplus p_{ij} = \adj_{ij} + p_{ij}\) if \(\adj_{ij} = 0\) and \(\adj_{ij} - p_{ij}\) otherwise \\
   \(\Pi_{\text{criterion}}(\features)\) & project operation w.r.t.\ some ``criterion'' \\

   \(\features\) and \(\featset\)    & inputs in the context of the analysis of the Soft Median (matrix and set notation) \\
   \(\mu(\features)\)                & some location estimate based on the inputs \(\features\) \\
   \(\pertm\) and \(\pertmset\)      & perturbed feature matrix used in breakdown point analysis (matrix and set notation)\\
   \(\vc\)                           & the cost/distances of the input instances to the dimension-wise median \\
   \(\circ\)                         & the element-wise multiplication \\
   \(C\)                             & the normalization of the weighted Soft Median \\
\bottomrule
\end{longtable}

\section{Surrogate Losses}\label{sec:appendix_ceisbad}

Hereinafter, we supplement the elaborations and experiments of the main part focusing on the surrogate losses. For the proof of \autoref{proposition:goodsurrogate} we refer to \autoref{sec:appendix_proofprop1}. We give a full definition of the losses in \autoref{tab:appendix_global_losses}. To simplify notation, we define the losses for a single node (except for \(\text{MCE}\)) and denote the correct class with \(c^*\). Note that \(\sV^+\) is the set of correctly classified nodes, \(\vp\) is the vector of confidence scores, and \(\vz\) is the vector with logits.

\begin{table}[ht]
\centering
\caption{Correspondence of global losses and their fulfilled properties corresponding to \autoref{sec:ceisbad}. ``+'' means that the property is obeyed and ``-'' that it is violated. Nevertheless,  we add the subjective category ``o'' to denote if the loss is partially / approximately consistent with the property. ``o/+'' denotes that the property is only exactly fulfilled for binary classification. \(\text{NCE}\) is the acronym for non-target class CE, and \emph{elu} stands for Exponential Linear Unit (smooth ReLU relaxation \(\text{elu}(z) = \min[\alpha \cdot (\exp(z) - 1),\, \text{ReLU}(z)] \) with \(\alpha=1\) in all our experiments}).\label{tab:appendix_global_losses}
\vskip 0.11in
\resizebox{1\linewidth}{!}{
    \begin{tabular}{lllcccc}
    \toprule
    \multicolumn{2}{l}{Category/Group} & \(\downarrow\) Loss \textbackslash Properties \(\rightarrow\) & (I) & (II) & (A) & (B) \\
    \midrule
    (1) & \multirow{2}{3cm}{focus on negative margins} & \(\text{CE} = -\evz_{c^*} + \log(\sum_{c \in \sC} \exp(\evz_{c}))\) & -  & +   & -  & o   \\
    & & \(\text{margin} = \max_{c \ne c^*} \evz_{c} - \evz_{c^*}\) & -   & -   & -  & -  \\
    (2) & \multirow{3}{3cm}{focus on high-confidence nodes} & \(\text{CW} = \min(\max_{c \ne c^*} \evz_{c} - \evz_{c^*}, 0)\) & +  & -   & +  & -  \\
    & & \(\text{NCE} = \max_{c \ne c^*} \evz_{c} - \log(\sum_{c' \in \sC} \exp(\evz_{c'}))\) & o   & +   & +  & o   \\
    & & \(\text{elu margin} = -\text{elu}(\max_{c \ne c^*} \evz_{c^*} - \evz_{c})\) & o   & -   & +  & o   \\
    (3) & \multirow{2}{3cm}{focus on nodes close to decision boundary} & \(\text{MCE} = \frac{-1}{|\sV^+|} \sum_{i \in \sV^+} \evz_{i, c^*} - \log(\sum_{c \in \sC} \exp(\evz_{i, c}))\) & +  & o/+   & +  & o   \\
    & & \(\text{tanh margin} = -\tanh(\max_{c \ne c^*} \evz_{c^*} - \evz_{c})\) & o   & +   & +  & +  \\
    \bottomrule
    \end{tabular}
}
\end{table}

\citet{Ma2020} propose a black-box attack based on a random-walk importance score. They select 1\% of nodes based on some centrality score and then gradually increase the feature perturbation of the selected nodes. They report that their (initial) black-box attack based on this importance score is only effective for low budgets. Since they propose their attack based on an analysis of a white-box attack, they conclude that this is due to a mismatch between accuracy and CW. First, note that their definition of the CW loss is what we call margin loss: $\text{margin} = \max_{c \ne c^\ast} \mathbf{z}_{c} - \mathbf{z}_{c^\ast}$. Instead, we follow \citet{Xu2019a} $\text{CW} = \min(\max_{c \ne c^\ast} \mathbf{z}_{c} - \mathbf{z}_{c^\ast}, 0)$ (see Eq.\ 6 in \citep{Xu2019a} with $\kappa=0$).

In summary, their finding is largely unrelated to our observations for the following reasons: (1) They select a fixed number of nodes (1\%) and observe that for severe feature perturbations the loss changes but the accuracy does not. So it seems like that if the perturbation budget of the attacked nodes is large enough then the predictions of the whole receptive field are successfully flipped. Instead, our study is exclusively for structure perturbations. (2) They study how to spread the perturbed nodes over the graph. Instead, we discuss that e.g.\ with CE most of the budget is spent on nodes that are wrongly classified in the clean graph (Properties I and A). (3) In contrast to \citet{Ma2020}, we also consider the fact that e.g. the CW loss comes with the risk of unsuccessfully spending all/too much budget on high-confidence nodes (Properties II and B).

\subsection{Learning Dynamics}

\begin{figure}[ht]
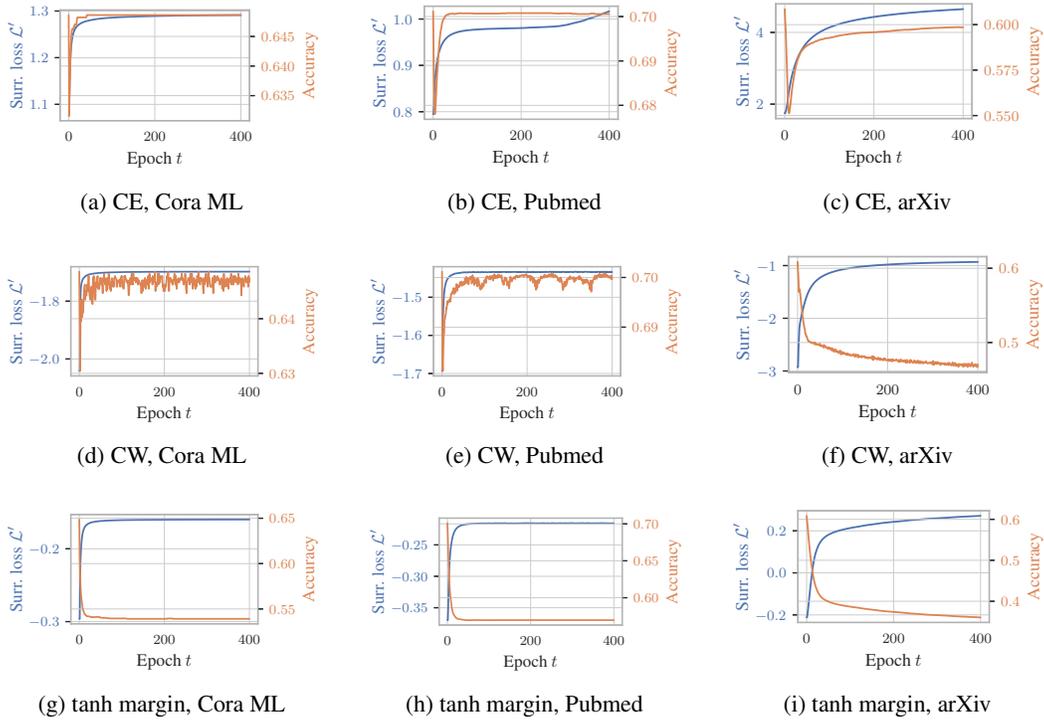

  \centering
  \makebox[\linewidth][c]{
    \(\begin{array}{ccc}
      \subfloat[CE, Cora ML]{\resizebox{0.32\linewidth}{!}{\input{assets/global_prbcd_attack_loss_cora_ml_False_0.01_CE.pgf}}} &
      \subfloat[CE, Pubmed]{\resizebox{0.32\linewidth}{!}{\input{assets/global_prbcd_attack_loss_pubmed_False_0.01_CE.pgf}}} &
      \subfloat[CE, arXiv]{\resizebox{0.32\linewidth}{!}{\input{assets/global_prbcd_attack_loss_ogbn-arxiv_False_0.01_CE.pgf}}}\\
      \subfloat[CW, Cora ML]{\resizebox{0.32\linewidth}{!}{\input{assets/global_prbcd_attack_loss_cora_ml_False_0.01_CW.pgf}}} &
      \subfloat[CW, Pubmed]{\resizebox{0.32\linewidth}{!}{\input{assets/global_prbcd_attack_loss_pubmed_False_0.01_CW.pgf}}} &
      \subfloat[CW, arXiv]{\resizebox{0.32\linewidth}{!}{\input{assets/global_prbcd_attack_loss_ogbn-arxiv_False_0.01_CW.pgf}}}\\
      \subfloat[tanh margin, Cora ML]{\resizebox{0.32\linewidth}{!}{\input{assets/global_prbcd_attack_loss_cora_ml_False_0.01_tanhMargin.pgf}}} &
      \subfloat[tanh margin, Pubmed]{\resizebox{0.32\linewidth}{!}{\input{assets/global_prbcd_attack_loss_pubmed_False_0.01_tanhMargin.pgf}}} &
      \subfloat[tanh margin, arXiv]{\resizebox{0.32\linewidth}{!}{\input{assets/global_prbcd_attack_loss_ogbn-arxiv_False_0.01_tanhMargin.pgf}}}\\
    \end{array}\)
  }
  \caption{PGD attack on a single layer GCN with \emph{directed} graph and \(\epsilon=0.01\).}%
  \label{fig:appendix_learningcurves_directed}
\end{figure}

The necessity of studying the surrogate losses originates from an unexpected behavior of the Cross Entropy (CE) loss and accuracy during an attack. In some cases, the loss increases significantly while the accuracy stays constant or even increases. Similarly, the so-called Carlini-Wagner (CW) loss~\cite{Xu2019a, Carlini2017} is very noisy over the epochs \(t\) during a global attack (e.g.\ see \autoref{fig:appendix_learningcurves_directed} (e)). Besides the violation of \autoref{definition:budgetaware} and \ref{definition:confidentandbudgetaware}, we hypothesize that the CW is inappropriate due to the effect on the optimization dynamics as we quickly explained in \autoref{sec:ceisbad}. We now first discuss the learning curves and then come back to the phenomenon.

\begin{figure}[ht]
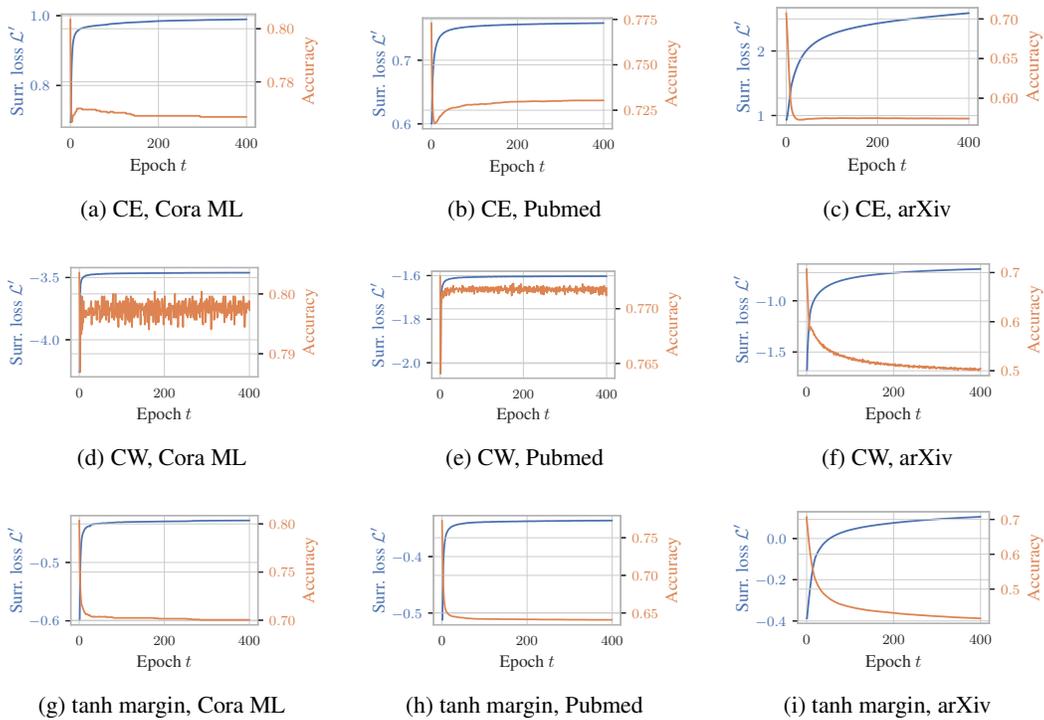

  \centering
  \makebox[\linewidth][c]{
    \(\begin{array}{ccc}
      \subfloat[CE, Cora ML]{\resizebox{0.32\linewidth}{!}{\input{assets/global_prbcd_attack_loss_cora_ml_True_0.01_CE.pgf}}} &
      \subfloat[CE, Pubmed]{\resizebox{0.32\linewidth}{!}{\input{assets/global_prbcd_attack_loss_pubmed_True_0.01_CE.pgf}}} &
      \subfloat[CE, arXiv]{\resizebox{0.32\linewidth}{!}{\input{assets/global_prbcd_attack_loss_ogbn-arxiv_True_0.01_CE.pgf}}}\\
      \subfloat[CW, Cora ML]{\resizebox{0.32\linewidth}{!}{\input{assets/global_prbcd_attack_loss_cora_ml_True_0.01_CW.pgf}}} &
      \subfloat[CW, Pubmed]{\resizebox{0.32\linewidth}{!}{\input{assets/global_prbcd_attack_loss_pubmed_True_0.01_CW.pgf}}} &
      \subfloat[CW, arXiv]{\resizebox{0.32\linewidth}{!}{\input{assets/global_prbcd_attack_loss_ogbn-arxiv_True_0.01_CW.pgf}}}\\
      \subfloat[tanh margin, Cora ML]{\resizebox{0.32\linewidth}{!}{\input{assets/global_prbcd_attack_loss_cora_ml_True_0.01_tanhMargin.pgf}}} &
      \subfloat[tanh margin, Pubmed]{\resizebox{0.32\linewidth}{!}{\input{assets/global_prbcd_attack_loss_pubmed_True_0.01_tanhMargin.pgf}}} &
      \subfloat[tanh margin, arXiv]{\resizebox{0.32\linewidth}{!}{\input{assets/global_prbcd_attack_loss_ogbn-arxiv_True_0.01_tanhMargin.pgf}}}\\
    \end{array}\)
  }
  \caption{PGD attack (no fine tuning) on a single layer GCN with \emph{undirected} graph  graph and \(\epsilon=0.01\).}%
  \label{fig:appendix_learningcurves_undirected}
\end{figure}

In \autoref{fig:appendix_learningcurves_directed}, we show the loss and accuracy during an attack for a large subset of datasets (see \autoref{tab:datasets}). Here, we study a single-layer GCN on a directed graph since this comes with a 1-to-1 correspondence between modified edges and attacked nodes. Especially for small budgets, the (CE) can increase while the accuracy does not decline. This shows the mismatch between (CE) and accuracy. In \autoref{fig:appendix_learningcurves_directed} (a-c), one can see that in the first epochs the accuracy reduces but then recovers almost to the clean accuracy during the attack. This happens despite the monotonic increase of the CE loss. In \autoref{fig:appendix_learningcurves_undirected}, we see that a similar, slightly-weaker behavior also holds for the common case of a two-layer / three-layer GCN on an undirected graph. That the observed effects appear to be weaker can be attributed to (1) the fact that an \emph{undirected} edge always influences both nodes and (2) the diffusion through multiple message-passing steps.

Now we come back to the phenomenon that the node's prediction can oscillate around the decision boundary (as pointed out in \autoref{sec:ceisbad}). The main reason is the zero gradient if $\psi < 0$ in the MCE (and CW) loss: $\text{CW} = \min(\max_{c \ne c^\ast} \mathbf{z}_{c} - \mathbf{z}_{c^\ast}, 0)$. To fully explain the reasons we need to dive into the PGD update and project step in epoch $t$:
5
\begin{equation}
    \mathbf{p}_{t} = \Pi_{\mathbb{E}[\text{Bernoulli}(\mathbf{p}_t)] \le \Delta} \left[ \mathbf{p}_{t-1} + \alpha_{t-1} \nabla \mathcal{L}(\mathbf{p}_{t-1}, \dots) \right]\,.
\end{equation}
We can rewrite this expression to
\begin{equation}
\mathbf{p}_{t} = \Pi_{[0, 1]}  ( \mathbf{p}_{t-1} \,\, \underbrace{+ \alpha_{t} \nabla \mathcal{L}(\mathbf{p}_{t-1}, \dots)}_{\text{gradient update}} \underbrace{- \eta_t \textbf{1}}_{\text{correction}} )
\end{equation}
where $\Pi_{[0, 1]}$ clamps the values to the range $[0, 1]$ ($\Rightarrow \mathbf{p}_{t} \in [0,1]^b$) and $\eta_t$ is chosen s.t. $\mathbb{E}[\text{Bernoulli}(\mathbf{p}_t)] \le \Delta$ (i.e. $\sum \Pi_{[0, 1]}(\mathbf{p}_t) \le \Delta$). There are two competing terms: 1) the gradient update $\alpha_{t-1} \nabla \mathcal{L}(\mathbf{p}_{t-1}, \dots)$ and 2) the correction $\eta_t \textbf{1}$ (typically lowers all weights in $\mathbf{p}_{t}$). For reasonable parameter choices, the potential perturbations in $\mathbf{p}_{t}$ are competing since our budget is limited (i.e. to maximize the loss we would like to flip more edges than budget we have). Then, after some epochs ($t > t_0$), we will have $\eta_t > 0$ and subtract $\eta_t$ from each element in $\mathbf{p}_{t-1}$.

Now if we choose a loss $\mathcal{L}$ (e.g. CW or MCE loss) that has zero gradient, as soon as a node $v$ is misclassified ($\psi_v < 0$), the responsible edge(s) will not benefit from a "gradient update" anymore but $\eta_t > 0$ is still subtracted. So after some iterations node $v$ will be again correctly classified since the required edge flips in $\mathbf{p}_t$ lost weight/strength. This leads to instability.

The symptoms are particularly visible in Figure 9e for the CW loss (after $t_0 = 25$ epochs the accuracy oscillates around 0.7). Moreover, the accuracy for the CW loss (Figure 9 d-f and Figure 10 d-f) are noisier than for the CE or tanh margin losses (other subfigures).

\clearpage
\subsection{What Nodes Are Being Attacked?}

\begin{figure}[H]
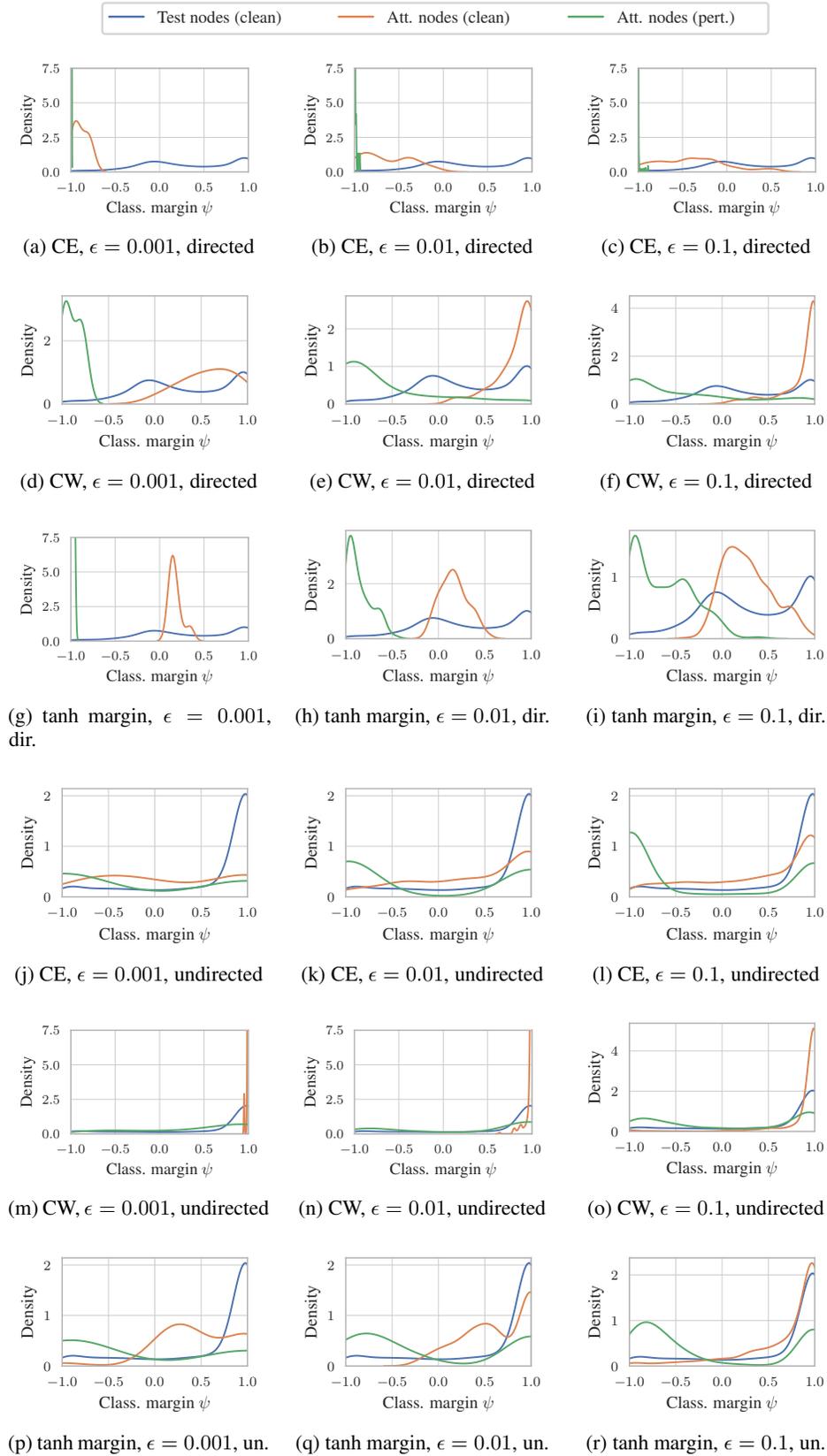

  \centering
  \hbox{\hspace{55pt}\resizebox{0.75\linewidth}{!}{
\begingroup%
\makeatletter%
\begin{pgfpicture}%
\pgfpathrectangle{\pgfpointorigin}{\pgfqpoint{4.491960in}{0.399973in}}%
\pgfusepath{use as bounding box, clip}%
\begin{pgfscope}%
\pgfsetbuttcap%
\pgfsetmiterjoin%
\definecolor{currentfill}{rgb}{1.000000,1.000000,1.000000}%
\pgfsetfillcolor{currentfill}%
\pgfsetlinewidth{0.000000pt}%
\definecolor{currentstroke}{rgb}{1.000000,1.000000,1.000000}%
\pgfsetstrokecolor{currentstroke}%
\pgfsetstrokeopacity{0.000000}%
\pgfsetdash{}{0pt}%
\pgfpathmoveto{\pgfqpoint{0.000000in}{0.000000in}}%
\pgfpathlineto{\pgfqpoint{4.491960in}{0.000000in}}%
\pgfpathlineto{\pgfqpoint{4.491960in}{0.399973in}}%
\pgfpathlineto{\pgfqpoint{0.000000in}{0.399973in}}%
\pgfpathclose%
\pgfusepath{fill}%
\end{pgfscope}%
\begin{pgfscope}%
\pgfsetbuttcap%
\pgfsetmiterjoin%
\definecolor{currentfill}{rgb}{1.000000,1.000000,1.000000}%
\pgfsetfillcolor{currentfill}%
\pgfsetfillopacity{0.800000}%
\pgfsetlinewidth{1.003750pt}%
\definecolor{currentstroke}{rgb}{0.800000,0.800000,0.800000}%
\pgfsetstrokecolor{currentstroke}%
\pgfsetstrokeopacity{0.800000}%
\pgfsetdash{}{0pt}%
\pgfpathmoveto{\pgfqpoint{0.122222in}{0.100000in}}%
\pgfpathlineto{\pgfqpoint{4.369738in}{0.100000in}}%
\pgfpathquadraticcurveto{\pgfqpoint{4.391960in}{0.100000in}}{\pgfqpoint{4.391960in}{0.122222in}}%
\pgfpathlineto{\pgfqpoint{4.391960in}{0.277751in}}%
\pgfpathquadraticcurveto{\pgfqpoint{4.391960in}{0.299973in}}{\pgfqpoint{4.369738in}{0.299973in}}%
\pgfpathlineto{\pgfqpoint{0.122222in}{0.299973in}}%
\pgfpathquadraticcurveto{\pgfqpoint{0.100000in}{0.299973in}}{\pgfqpoint{0.100000in}{0.277751in}}%
\pgfpathlineto{\pgfqpoint{0.100000in}{0.122222in}}%
\pgfpathquadraticcurveto{\pgfqpoint{0.100000in}{0.100000in}}{\pgfqpoint{0.122222in}{0.100000in}}%
\pgfpathclose%
\pgfusepath{stroke,fill}%
\end{pgfscope}%
\begin{pgfscope}%
\pgfsetroundcap%
\pgfsetroundjoin%
\pgfsetlinewidth{1.003750pt}%
\definecolor{currentstroke}{rgb}{0.298039,0.447059,0.690196}%
\pgfsetstrokecolor{currentstroke}%
\pgfsetdash{}{0pt}%
\pgfpathmoveto{\pgfqpoint{0.144444in}{0.211104in}}%
\pgfpathlineto{\pgfqpoint{0.366667in}{0.211104in}}%
\pgfusepath{stroke}%
\end{pgfscope}%
\begin{pgfscope}%
\definecolor{textcolor}{rgb}{0.150000,0.150000,0.150000}%
\pgfsetstrokecolor{textcolor}%
\pgfsetfillcolor{textcolor}%
\pgftext[x=0.455556in,y=0.172215in,left,base]{\color{textcolor}\rmfamily\fontsize{8.000000}{9.600000}\selectfont Test nodes (clean)}%
\end{pgfscope}%
\begin{pgfscope}%
\pgfsetroundcap%
\pgfsetroundjoin%
\pgfsetlinewidth{1.003750pt}%
\definecolor{currentstroke}{rgb}{0.866667,0.517647,0.321569}%
\pgfsetstrokecolor{currentstroke}%
\pgfsetdash{}{0pt}%
\pgfpathmoveto{\pgfqpoint{1.623303in}{0.211104in}}%
\pgfpathlineto{\pgfqpoint{1.845525in}{0.211104in}}%
\pgfusepath{stroke}%
\end{pgfscope}%
\begin{pgfscope}%
\definecolor{textcolor}{rgb}{0.150000,0.150000,0.150000}%
\pgfsetstrokecolor{textcolor}%
\pgfsetfillcolor{textcolor}%
\pgftext[x=1.934414in,y=0.172215in,left,base]{\color{textcolor}\rmfamily\fontsize{8.000000}{9.600000}\selectfont Att. nodes (clean)}%
\end{pgfscope}%
\begin{pgfscope}%
\pgfsetroundcap%
\pgfsetroundjoin%
\pgfsetlinewidth{1.003750pt}%
\definecolor{currentstroke}{rgb}{0.333333,0.658824,0.407843}%
\pgfsetstrokecolor{currentstroke}%
\pgfsetdash{}{0pt}%
\pgfpathmoveto{\pgfqpoint{3.104669in}{0.211104in}}%
\pgfpathlineto{\pgfqpoint{3.326891in}{0.211104in}}%
\pgfusepath{stroke}%
\end{pgfscope}%
\begin{pgfscope}%
\definecolor{textcolor}{rgb}{0.150000,0.150000,0.150000}%
\pgfsetstrokecolor{textcolor}%
\pgfsetfillcolor{textcolor}%
\pgftext[x=3.415780in,y=0.172215in,left,base]{\color{textcolor}\rmfamily\fontsize{8.000000}{9.600000}\selectfont Att. nodes (pert.)}%
\end{pgfscope}%
\end{pgfpicture}%
\makeatother%
\endgroup%
}}
  \vspace{-10pt}
  \makebox[\linewidth][c]{
    \(\begin{array}{ccc}
      \subfloat[CE, $\epsilon=0.001$, directed]{\resizebox{0.28\linewidth}{!}{\input{assets/global_prbcd_kde_nl_cora_ml_False_0.001_CE_margin_attacked.pgf}}} &
      \subfloat[CE, $\epsilon=0.01$, directed]{\resizebox{0.28\linewidth}{!}{\input{assets/global_prbcd_kde_nl_cora_ml_False_0.01_CE_margin_attacked.pgf}}} &
      \subfloat[CE, $\epsilon=0.1$, directed]{\resizebox{0.28\linewidth}{!}{\input{assets/global_prbcd_kde_nl_cora_ml_False_0.1_CE_margin_attacked.pgf}}} \\
      \subfloat[CW, $\epsilon=0.001$, directed]{\resizebox{0.28\linewidth}{!}{\input{assets/global_prbcd_kde_nl_cora_ml_False_0.001_CW_margin_attacked.pgf}}} &
      \subfloat[CW, $\epsilon=0.01$, directed]{\resizebox{0.28\linewidth}{!}{\input{assets/global_prbcd_kde_nl_cora_ml_False_0.01_CW_margin_attacked.pgf}}} &
      \subfloat[CW, $\epsilon=0.1$, directed]{\resizebox{0.28\linewidth}{!}{\input{assets/global_prbcd_kde_nl_cora_ml_False_0.1_CW_margin_attacked.pgf}}} \\
      \subfloat[tanh margin, $\epsilon=0.001$, dir.]{\resizebox{0.28\linewidth}{!}{\input{assets/global_prbcd_kde_nl_cora_ml_False_0.001_tanhMargin_margin_attacked.pgf}}} &
      \subfloat[tanh margin, $\epsilon=0.01$, dir.]{\resizebox{0.28\linewidth}{!}{\input{assets/global_prbcd_kde_nl_cora_ml_False_0.01_tanhMargin_margin_attacked.pgf}}} &
      \subfloat[tanh margin, $\epsilon=0.1$, dir.]{\resizebox{0.28\linewidth}{!}{\input{assets/global_prbcd_kde_nl_cora_ml_False_0.1_tanhMargin_margin_attacked.pgf}}} \\
      \subfloat[CE, $\epsilon=0.001$, undirected]{\resizebox{0.28\linewidth}{!}{\input{assets/global_prbcd_kde_nl_cora_ml_True_0.001_CE_margin_attacked.pgf}}} &
      \subfloat[CE, $\epsilon=0.01$, undirected]{\resizebox{0.28\linewidth}{!}{\input{assets/global_prbcd_kde_nl_cora_ml_True_0.01_CE_margin_attacked.pgf}}} &
      \subfloat[CE, $\epsilon=0.1$, undirected]{\resizebox{0.28\linewidth}{!}{\input{assets/global_prbcd_kde_nl_cora_ml_True_0.1_CE_margin_attacked.pgf}}} \\
      \subfloat[CW, $\epsilon=0.001$, undirected]{\resizebox{0.28\linewidth}{!}{\input{assets/global_prbcd_kde_nl_cora_ml_True_0.001_CW_margin_attacked.pgf}}} &
      \subfloat[CW, $\epsilon=0.01$, undirected]{\resizebox{0.28\linewidth}{!}{\input{assets/global_prbcd_kde_nl_cora_ml_True_0.01_CW_margin_attacked.pgf}}} &
      \subfloat[CW, $\epsilon=0.1$, undirected]{\resizebox{0.28\linewidth}{!}{\input{assets/global_prbcd_kde_nl_cora_ml_True_0.1_CW_margin_attacked.pgf}}} \\
      \subfloat[tanh margin, $\epsilon=0.001$, un.]{\resizebox{0.28\linewidth}{!}{\input{assets/global_prbcd_kde_nl_cora_ml_True_0.001_tanhMargin_margin_attacked.pgf}}} &
      \subfloat[tanh margin, $\epsilon=0.01$, un.]{\resizebox{0.28\linewidth}{!}{\input{assets/global_prbcd_kde_nl_cora_ml_True_0.01_tanhMargin_margin_attacked.pgf}}} &
      \subfloat[tanh margin, $\epsilon=0.1$, un.]{\resizebox{0.28\linewidth}{!}{\input{assets/global_prbcd_kde_nl_cora_ml_True_0.1_tanhMargin_margin_attacked.pgf}}} \\
    \end{array}\)
  }
  \caption{Distribution of nodes attacked before/after PGD attack on Vanilla GCN on Cora ML.}%
  \label{fig:appendix_margin_densities_cora_ml}
\end{figure}

\begin{figure}[H]
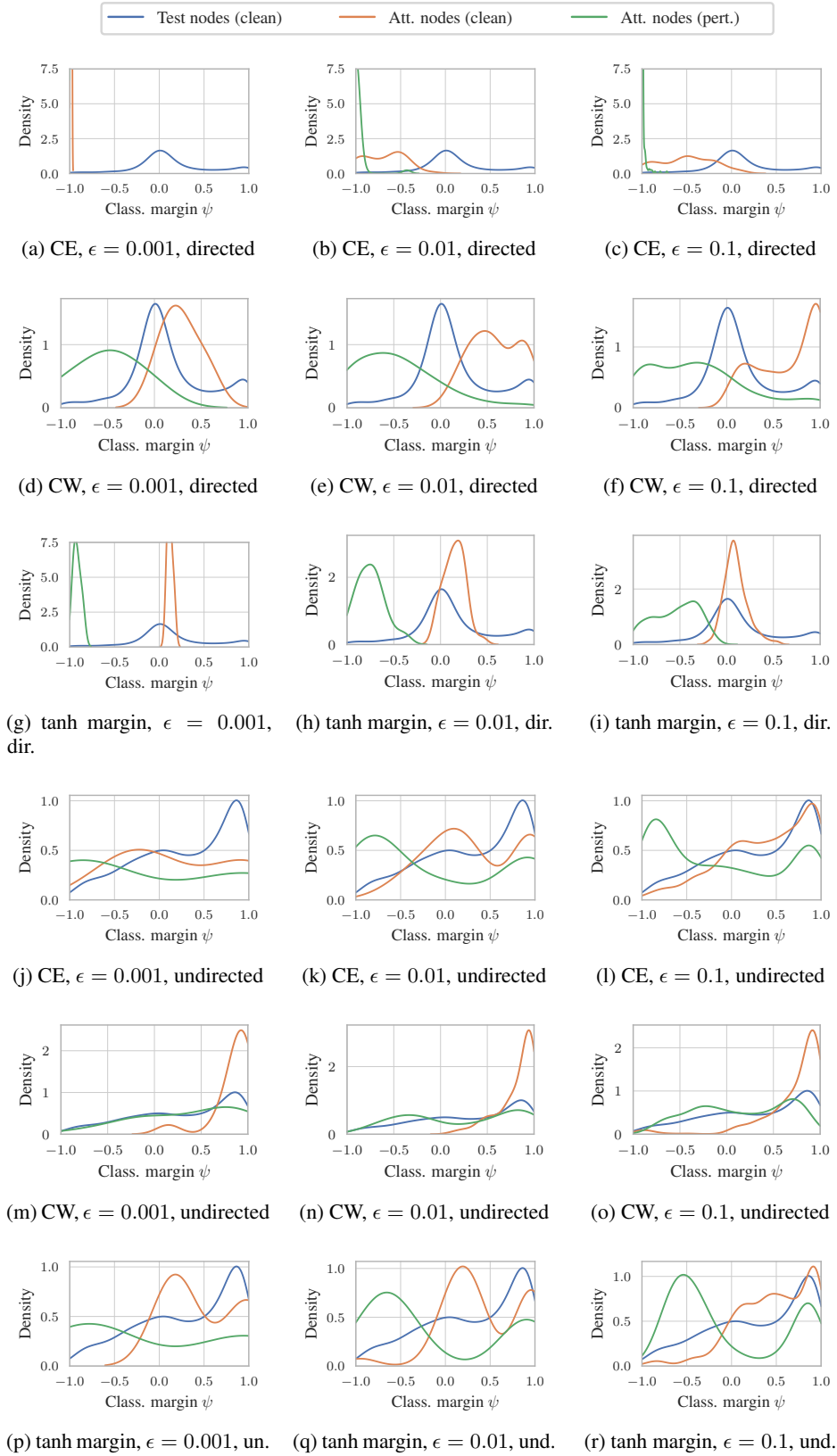

  \centering
  \hbox{\hspace{55pt}\resizebox{0.75\linewidth}{!}{
\begingroup%
\makeatletter%
\begin{pgfpicture}%
\pgfpathrectangle{\pgfpointorigin}{\pgfqpoint{4.491960in}{0.399973in}}%
\pgfusepath{use as bounding box, clip}%
\begin{pgfscope}%
\pgfsetbuttcap%
\pgfsetmiterjoin%
\definecolor{currentfill}{rgb}{1.000000,1.000000,1.000000}%
\pgfsetfillcolor{currentfill}%
\pgfsetlinewidth{0.000000pt}%
\definecolor{currentstroke}{rgb}{1.000000,1.000000,1.000000}%
\pgfsetstrokecolor{currentstroke}%
\pgfsetstrokeopacity{0.000000}%
\pgfsetdash{}{0pt}%
\pgfpathmoveto{\pgfqpoint{0.000000in}{0.000000in}}%
\pgfpathlineto{\pgfqpoint{4.491960in}{0.000000in}}%
\pgfpathlineto{\pgfqpoint{4.491960in}{0.399973in}}%
\pgfpathlineto{\pgfqpoint{0.000000in}{0.399973in}}%
\pgfpathclose%
\pgfusepath{fill}%
\end{pgfscope}%
\begin{pgfscope}%
\pgfsetbuttcap%
\pgfsetmiterjoin%
\definecolor{currentfill}{rgb}{1.000000,1.000000,1.000000}%
\pgfsetfillcolor{currentfill}%
\pgfsetfillopacity{0.800000}%
\pgfsetlinewidth{1.003750pt}%
\definecolor{currentstroke}{rgb}{0.800000,0.800000,0.800000}%
\pgfsetstrokecolor{currentstroke}%
\pgfsetstrokeopacity{0.800000}%
\pgfsetdash{}{0pt}%
\pgfpathmoveto{\pgfqpoint{0.122222in}{0.100000in}}%
\pgfpathlineto{\pgfqpoint{4.369738in}{0.100000in}}%
\pgfpathquadraticcurveto{\pgfqpoint{4.391960in}{0.100000in}}{\pgfqpoint{4.391960in}{0.122222in}}%
\pgfpathlineto{\pgfqpoint{4.391960in}{0.277751in}}%
\pgfpathquadraticcurveto{\pgfqpoint{4.391960in}{0.299973in}}{\pgfqpoint{4.369738in}{0.299973in}}%
\pgfpathlineto{\pgfqpoint{0.122222in}{0.299973in}}%
\pgfpathquadraticcurveto{\pgfqpoint{0.100000in}{0.299973in}}{\pgfqpoint{0.100000in}{0.277751in}}%
\pgfpathlineto{\pgfqpoint{0.100000in}{0.122222in}}%
\pgfpathquadraticcurveto{\pgfqpoint{0.100000in}{0.100000in}}{\pgfqpoint{0.122222in}{0.100000in}}%
\pgfpathclose%
\pgfusepath{stroke,fill}%
\end{pgfscope}%
\begin{pgfscope}%
\pgfsetroundcap%
\pgfsetroundjoin%
\pgfsetlinewidth{1.003750pt}%
\definecolor{currentstroke}{rgb}{0.298039,0.447059,0.690196}%
\pgfsetstrokecolor{currentstroke}%
\pgfsetdash{}{0pt}%
\pgfpathmoveto{\pgfqpoint{0.144444in}{0.211104in}}%
\pgfpathlineto{\pgfqpoint{0.366667in}{0.211104in}}%
\pgfusepath{stroke}%
\end{pgfscope}%
\begin{pgfscope}%
\definecolor{textcolor}{rgb}{0.150000,0.150000,0.150000}%
\pgfsetstrokecolor{textcolor}%
\pgfsetfillcolor{textcolor}%
\pgftext[x=0.455556in,y=0.172215in,left,base]{\color{textcolor}\rmfamily\fontsize{8.000000}{9.600000}\selectfont Test nodes (clean)}%
\end{pgfscope}%
\begin{pgfscope}%
\pgfsetroundcap%
\pgfsetroundjoin%
\pgfsetlinewidth{1.003750pt}%
\definecolor{currentstroke}{rgb}{0.866667,0.517647,0.321569}%
\pgfsetstrokecolor{currentstroke}%
\pgfsetdash{}{0pt}%
\pgfpathmoveto{\pgfqpoint{1.623303in}{0.211104in}}%
\pgfpathlineto{\pgfqpoint{1.845525in}{0.211104in}}%
\pgfusepath{stroke}%
\end{pgfscope}%
\begin{pgfscope}%
\definecolor{textcolor}{rgb}{0.150000,0.150000,0.150000}%
\pgfsetstrokecolor{textcolor}%
\pgfsetfillcolor{textcolor}%
\pgftext[x=1.934414in,y=0.172215in,left,base]{\color{textcolor}\rmfamily\fontsize{8.000000}{9.600000}\selectfont Att. nodes (clean)}%
\end{pgfscope}%
\begin{pgfscope}%
\pgfsetroundcap%
\pgfsetroundjoin%
\pgfsetlinewidth{1.003750pt}%
\definecolor{currentstroke}{rgb}{0.333333,0.658824,0.407843}%
\pgfsetstrokecolor{currentstroke}%
\pgfsetdash{}{0pt}%
\pgfpathmoveto{\pgfqpoint{3.104669in}{0.211104in}}%
\pgfpathlineto{\pgfqpoint{3.326891in}{0.211104in}}%
\pgfusepath{stroke}%
\end{pgfscope}%
\begin{pgfscope}%
\definecolor{textcolor}{rgb}{0.150000,0.150000,0.150000}%
\pgfsetstrokecolor{textcolor}%
\pgfsetfillcolor{textcolor}%
\pgftext[x=3.415780in,y=0.172215in,left,base]{\color{textcolor}\rmfamily\fontsize{8.000000}{9.600000}\selectfont Att. nodes (pert.)}%
\end{pgfscope}%
\end{pgfpicture}%
\makeatother%
\endgroup%
}}
  \vspace{-10pt}
  \makebox[\linewidth][c]{
    \(\begin{array}{ccc}
      \subfloat[CE, $\epsilon=0.001$, directed]{\resizebox{0.28\linewidth}{!}{\input{assets/global_prbcd_kde_nl_citeseer_False_0.001_CE_margin_attacked.pgf}}} &
      \subfloat[CE, $\epsilon=0.01$, directed]{\resizebox{0.28\linewidth}{!}{\input{assets/global_prbcd_kde_nl_citeseer_False_0.01_CE_margin_attacked.pgf}}} &
      \subfloat[CE, $\epsilon=0.1$, directed]{\resizebox{0.28\linewidth}{!}{\input{assets/global_prbcd_kde_nl_citeseer_False_0.1_CE_margin_attacked.pgf}}} \\
      \subfloat[CW, $\epsilon=0.001$, directed]{\resizebox{0.28\linewidth}{!}{\input{assets/global_prbcd_kde_nl_citeseer_False_0.001_CW_margin_attacked.pgf}}} &
      \subfloat[CW, $\epsilon=0.01$, directed]{\resizebox{0.28\linewidth}{!}{\input{assets/global_prbcd_kde_nl_citeseer_False_0.01_CW_margin_attacked.pgf}}} &
      \subfloat[CW, $\epsilon=0.1$, directed]{\resizebox{0.28\linewidth}{!}{\input{assets/global_prbcd_kde_nl_citeseer_False_0.1_CW_margin_attacked.pgf}}} \\
      \subfloat[tanh margin, $\epsilon=0.001$, dir.]{\resizebox{0.28\linewidth}{!}{\input{assets/global_prbcd_kde_nl_citeseer_False_0.001_tanhMargin_margin_attacked.pgf}}} &
      \subfloat[tanh margin, $\epsilon=0.01$, dir.]{\resizebox{0.28\linewidth}{!}{\input{assets/global_prbcd_kde_nl_citeseer_False_0.01_tanhMargin_margin_attacked.pgf}}} &
      \subfloat[tanh margin, $\epsilon=0.1$, dir.]{\resizebox{0.28\linewidth}{!}{\input{assets/global_prbcd_kde_nl_citeseer_False_0.1_tanhMargin_margin_attacked.pgf}}} \\
      \subfloat[CE, $\epsilon=0.001$, undirected]{\resizebox{0.28\linewidth}{!}{\input{assets/global_prbcd_kde_nl_citeseer_True_0.001_CE_margin_attacked.pgf}}} &
      \subfloat[CE, $\epsilon=0.01$, undirected]{\resizebox{0.28\linewidth}{!}{\input{assets/global_prbcd_kde_nl_citeseer_True_0.01_CE_margin_attacked.pgf}}} &
      \subfloat[CE, $\epsilon=0.1$, undirected]{\resizebox{0.28\linewidth}{!}{\input{assets/global_prbcd_kde_nl_citeseer_True_0.1_CE_margin_attacked.pgf}}} \\
      \subfloat[CW, $\epsilon=0.001$, undirected]{\resizebox{0.28\linewidth}{!}{\input{assets/global_prbcd_kde_nl_citeseer_True_0.001_CW_margin_attacked.pgf}}} &
      \subfloat[CW, $\epsilon=0.01$, undirected]{\resizebox{0.28\linewidth}{!}{\input{assets/global_prbcd_kde_nl_citeseer_True_0.01_CW_margin_attacked.pgf}}} &
      \subfloat[CW, $\epsilon=0.1$, undirected]{\resizebox{0.28\linewidth}{!}{\input{assets/global_prbcd_kde_nl_citeseer_True_0.1_CW_margin_attacked.pgf}}} \\
      \subfloat[tanh margin, $\epsilon=0.001$, un.]{\resizebox{0.28\linewidth}{!}{\input{assets/global_prbcd_kde_nl_citeseer_True_0.001_tanhMargin_margin_attacked.pgf}}} &
      \subfloat[tanh margin, $\epsilon=0.01$, und.]{\resizebox{0.28\linewidth}{!}{\input{assets/global_prbcd_kde_nl_citeseer_True_0.01_tanhMargin_margin_attacked.pgf}}} &
      \subfloat[tanh margin, $\epsilon=0.1$, und.]{\resizebox{0.28\linewidth}{!}{\input{assets/global_prbcd_kde_nl_citeseer_True_0.1_tanhMargin_margin_attacked.pgf}}} \\
    \end{array}\)
  }
  \caption{Distribution of nodes attacked before/after PGD attack on Vanilla GCN on Citeseer.}%
  \label{fig:appendix_margin_densities_citeseer}
\end{figure}

In \autoref{fig:appendix_margin_densities_cora_ml}, \ref{fig:appendix_margin_densities_citeseer}, \ref{fig:appendix_margin_densities_pubmed} and \ref{fig:appendix_margin_densities_arxiv}, we show the distribution for different datasets and budgets. Here we underline that our findings are consistent for these variations in the experiment setup. These plots essentially show the same thing as similarly to \autoref{fig:negceprob} for more configurations. However, instead of bar plots we show density plots since they allow more nuanced conclusions. We distinguish again between a directed and undirected graph. With CE we mostly attack nodes with a negative margin. For example in plot \autoref{fig:appendix_margin_densities_citeseer} (a), the distribution for the attacked nodes is extremely spiky (at (\(\psi=-1\))) leading to a failing kernel density estimate. On the contrary, CW attacks correctly classified nodes proportionally to the clean distribution, and \emph{our} tanh margin focuses on nodes close to the decision boundary. We see that our observation holds over a wide range of datasets and budgets. All the stated observations are particularly clear if we consider tiny budgets and a directed graph.
In the undirected case, using the CE and tanh margin also target confident nodes. Similarly to before, we attribute this effect to (1) the fact that on an undirected graph an edge always influences both nodes and (2) the diffusion via the recursive message passing. For large budgets and an undirected graph the differences between the losses become less significant. Simultaneously, an increasing budget becomes less realistic for many applications.
     
In \autoref{fig:appendix_empsurrogate}, which is similar to \autoref{fig:empsurrogate}, we provide larger budgets and more datasets. As long as the budget is sufficiently small, our observations hold: (1) for the greedy FGSM attack the MCE is the strongest loss and (2) for the projected gradient descent attacks (PGD and our PR-BCD) the tanh margin outperforms other losses. With larger budgets, the elu margin seems to be a particularly strong choice. Note that the elu margin diverges for \(\psi \to 1\) but, in contrast to the CW, it is smooth and encourages confident misclassifications (see \autoref{fig:losses}). We hypothesize that for sufficiently large budgets, it makes sense to incentivize first attacking high-confidence nodes instead of nodes close to the decision boundary since those are presumably the most difficult to convert. However, such large budgets (e.g.\ \(\epsilon > 0.25\)) are not realistic for most applications. On arXiv, the elu margin is already slightly stronger than the tanh margin for \(\epsilon > 0.075\). This is probably largely driven by the fact that we only attack 29\% of nodes (i.e.\ the test set) while the budget is calculated relatively to all edges. Hence, in comparison to Cora ML, Citeseer, or Pubmed, we effectively have a four times larger budget per attacked node. %
\clearpage

\begin{figure}[H]
  \centering
  \hbox{\hspace{55pt}\resizebox{0.75\linewidth}{!}{
\begingroup%
\makeatletter%
\begin{pgfpicture}%
\pgfpathrectangle{\pgfpointorigin}{\pgfqpoint{4.491960in}{0.399973in}}%
\pgfusepath{use as bounding box, clip}%
\begin{pgfscope}%
\pgfsetbuttcap%
\pgfsetmiterjoin%
\definecolor{currentfill}{rgb}{1.000000,1.000000,1.000000}%
\pgfsetfillcolor{currentfill}%
\pgfsetlinewidth{0.000000pt}%
\definecolor{currentstroke}{rgb}{1.000000,1.000000,1.000000}%
\pgfsetstrokecolor{currentstroke}%
\pgfsetstrokeopacity{0.000000}%
\pgfsetdash{}{0pt}%
\pgfpathmoveto{\pgfqpoint{0.000000in}{0.000000in}}%
\pgfpathlineto{\pgfqpoint{4.491960in}{0.000000in}}%
\pgfpathlineto{\pgfqpoint{4.491960in}{0.399973in}}%
\pgfpathlineto{\pgfqpoint{0.000000in}{0.399973in}}%
\pgfpathclose%
\pgfusepath{fill}%
\end{pgfscope}%
\begin{pgfscope}%
\pgfsetbuttcap%
\pgfsetmiterjoin%
\definecolor{currentfill}{rgb}{1.000000,1.000000,1.000000}%
\pgfsetfillcolor{currentfill}%
\pgfsetfillopacity{0.800000}%
\pgfsetlinewidth{1.003750pt}%
\definecolor{currentstroke}{rgb}{0.800000,0.800000,0.800000}%
\pgfsetstrokecolor{currentstroke}%
\pgfsetstrokeopacity{0.800000}%
\pgfsetdash{}{0pt}%
\pgfpathmoveto{\pgfqpoint{0.122222in}{0.100000in}}%
\pgfpathlineto{\pgfqpoint{4.369738in}{0.100000in}}%
\pgfpathquadraticcurveto{\pgfqpoint{4.391960in}{0.100000in}}{\pgfqpoint{4.391960in}{0.122222in}}%
\pgfpathlineto{\pgfqpoint{4.391960in}{0.277751in}}%
\pgfpathquadraticcurveto{\pgfqpoint{4.391960in}{0.299973in}}{\pgfqpoint{4.369738in}{0.299973in}}%
\pgfpathlineto{\pgfqpoint{0.122222in}{0.299973in}}%
\pgfpathquadraticcurveto{\pgfqpoint{0.100000in}{0.299973in}}{\pgfqpoint{0.100000in}{0.277751in}}%
\pgfpathlineto{\pgfqpoint{0.100000in}{0.122222in}}%
\pgfpathquadraticcurveto{\pgfqpoint{0.100000in}{0.100000in}}{\pgfqpoint{0.122222in}{0.100000in}}%
\pgfpathclose%
\pgfusepath{stroke,fill}%
\end{pgfscope}%
\begin{pgfscope}%
\pgfsetroundcap%
\pgfsetroundjoin%
\pgfsetlinewidth{1.003750pt}%
\definecolor{currentstroke}{rgb}{0.298039,0.447059,0.690196}%
\pgfsetstrokecolor{currentstroke}%
\pgfsetdash{}{0pt}%
\pgfpathmoveto{\pgfqpoint{0.144444in}{0.211104in}}%
\pgfpathlineto{\pgfqpoint{0.366667in}{0.211104in}}%
\pgfusepath{stroke}%
\end{pgfscope}%
\begin{pgfscope}%
\definecolor{textcolor}{rgb}{0.150000,0.150000,0.150000}%
\pgfsetstrokecolor{textcolor}%
\pgfsetfillcolor{textcolor}%
\pgftext[x=0.455556in,y=0.172215in,left,base]{\color{textcolor}\rmfamily\fontsize{8.000000}{9.600000}\selectfont Test nodes (clean)}%
\end{pgfscope}%
\begin{pgfscope}%
\pgfsetroundcap%
\pgfsetroundjoin%
\pgfsetlinewidth{1.003750pt}%
\definecolor{currentstroke}{rgb}{0.866667,0.517647,0.321569}%
\pgfsetstrokecolor{currentstroke}%
\pgfsetdash{}{0pt}%
\pgfpathmoveto{\pgfqpoint{1.623303in}{0.211104in}}%
\pgfpathlineto{\pgfqpoint{1.845525in}{0.211104in}}%
\pgfusepath{stroke}%
\end{pgfscope}%
\begin{pgfscope}%
\definecolor{textcolor}{rgb}{0.150000,0.150000,0.150000}%
\pgfsetstrokecolor{textcolor}%
\pgfsetfillcolor{textcolor}%
\pgftext[x=1.934414in,y=0.172215in,left,base]{\color{textcolor}\rmfamily\fontsize{8.000000}{9.600000}\selectfont Att. nodes (clean)}%
\end{pgfscope}%
\begin{pgfscope}%
\pgfsetroundcap%
\pgfsetroundjoin%
\pgfsetlinewidth{1.003750pt}%
\definecolor{currentstroke}{rgb}{0.333333,0.658824,0.407843}%
\pgfsetstrokecolor{currentstroke}%
\pgfsetdash{}{0pt}%
\pgfpathmoveto{\pgfqpoint{3.104669in}{0.211104in}}%
\pgfpathlineto{\pgfqpoint{3.326891in}{0.211104in}}%
\pgfusepath{stroke}%
\end{pgfscope}%
\begin{pgfscope}%
\definecolor{textcolor}{rgb}{0.150000,0.150000,0.150000}%
\pgfsetstrokecolor{textcolor}%
\pgfsetfillcolor{textcolor}%
\pgftext[x=3.415780in,y=0.172215in,left,base]{\color{textcolor}\rmfamily\fontsize{8.000000}{9.600000}\selectfont Att. nodes (pert.)}%
\end{pgfscope}%
\end{pgfpicture}%
\makeatother%
\endgroup%
}}
  \vspace{-10pt}
  \makebox[\linewidth][c]{
    \(\begin{array}{ccc}
      \subfloat[CE, $\epsilon=0.001$, directed]{\resizebox{0.28\linewidth}{!}{\input{assets/global_prbcd_kde_nl_pubmed_False_0.001_CE_margin_attacked.pgf}}} &
      \subfloat[CE, $\epsilon=0.01$, directed]{\resizebox{0.28\linewidth}{!}{\input{assets/global_prbcd_kde_nl_pubmed_False_0.01_CE_margin_attacked.pgf}}} &
      \subfloat[CE, $\epsilon=0.1$, directed]{\resizebox{0.28\linewidth}{!}{\input{assets/global_prbcd_kde_nl_pubmed_False_0.1_CE_margin_attacked.pgf}}} \\
      \subfloat[CW, $\epsilon=0.001$, directed]{\resizebox{0.28\linewidth}{!}{\input{assets/global_prbcd_kde_nl_pubmed_False_0.001_CW_margin_attacked.pgf}}} &
      \subfloat[CW, $\epsilon=0.01$, directed]{\resizebox{0.28\linewidth}{!}{\input{assets/global_prbcd_kde_nl_pubmed_False_0.01_CW_margin_attacked.pgf}}} &
      \subfloat[CW, $\epsilon=0.1$, directed]{\resizebox{0.28\linewidth}{!}{\input{assets/global_prbcd_kde_nl_pubmed_False_0.1_CW_margin_attacked.pgf}}} \\
      \subfloat[tanh margin, $\epsilon=0.001$, dir.]{\resizebox{0.28\linewidth}{!}{\input{assets/global_prbcd_kde_nl_pubmed_False_0.001_tanhMargin_margin_attacked.pgf}}} &
      \subfloat[tanh margin, $\epsilon=0.01$, dir.]{\resizebox{0.28\linewidth}{!}{\input{assets/global_prbcd_kde_nl_pubmed_False_0.01_tanhMargin_margin_attacked.pgf}}} &
      \subfloat[tanh margin, $\epsilon=0.1$, dir.]{\resizebox{0.28\linewidth}{!}{\input{assets/global_prbcd_kde_nl_pubmed_False_0.1_tanhMargin_margin_attacked.pgf}}} \\
      \subfloat[CE, $\epsilon=0.001$, undirected]{\resizebox{0.28\linewidth}{!}{\input{assets/global_prbcd_kde_nl_pubmed_True_0.001_CE_margin_attacked.pgf}}} &
      \subfloat[CE, $\epsilon=0.01$, undirected]{\resizebox{0.28\linewidth}{!}{\input{assets/global_prbcd_kde_nl_pubmed_True_0.01_CE_margin_attacked.pgf}}} &
      \subfloat[CE, $\epsilon=0.1$, undirected]{\resizebox{0.28\linewidth}{!}{\input{assets/global_prbcd_kde_nl_pubmed_True_0.1_CE_margin_attacked.pgf}}} \\
      \subfloat[CW, $\epsilon=0.001$, undirected]{\resizebox{0.28\linewidth}{!}{\input{assets/global_prbcd_kde_nl_pubmed_True_0.001_CW_margin_attacked.pgf}}} &
      \subfloat[CW, $\epsilon=0.01$, undirected]{\resizebox{0.28\linewidth}{!}{\input{assets/global_prbcd_kde_nl_pubmed_True_0.01_CW_margin_attacked.pgf}}} &
      \subfloat[CW, $\epsilon=0.1$, undirected]{\resizebox{0.28\linewidth}{!}{\input{assets/global_prbcd_kde_nl_pubmed_True_0.1_CW_margin_attacked.pgf}}} \\
      \subfloat[tanh margin, $\epsilon=0.001$, un.]{\resizebox{0.28\linewidth}{!}{\input{assets/global_prbcd_kde_nl_pubmed_True_0.001_tanhMargin_margin_attacked.pgf}}} &
      \subfloat[tanh margin, $\epsilon=0.01$, und.]{\resizebox{0.28\linewidth}{!}{\input{assets/global_prbcd_kde_nl_pubmed_True_0.01_tanhMargin_margin_attacked.pgf}}} &
      \subfloat[tanh margin, $\epsilon=0.1$, und.]{\resizebox{0.28\linewidth}{!}{\input{assets/global_prbcd_kde_nl_pubmed_True_0.1_tanhMargin_margin_attacked.pgf}}} \\
    \end{array}\)
  }
  \caption{Distribution of nodes attacked before/after PGD attack on Vanilla GCN on Pubmed.}%
  \label{fig:appendix_margin_densities_pubmed}
\end{figure}

\begin{figure}[H]
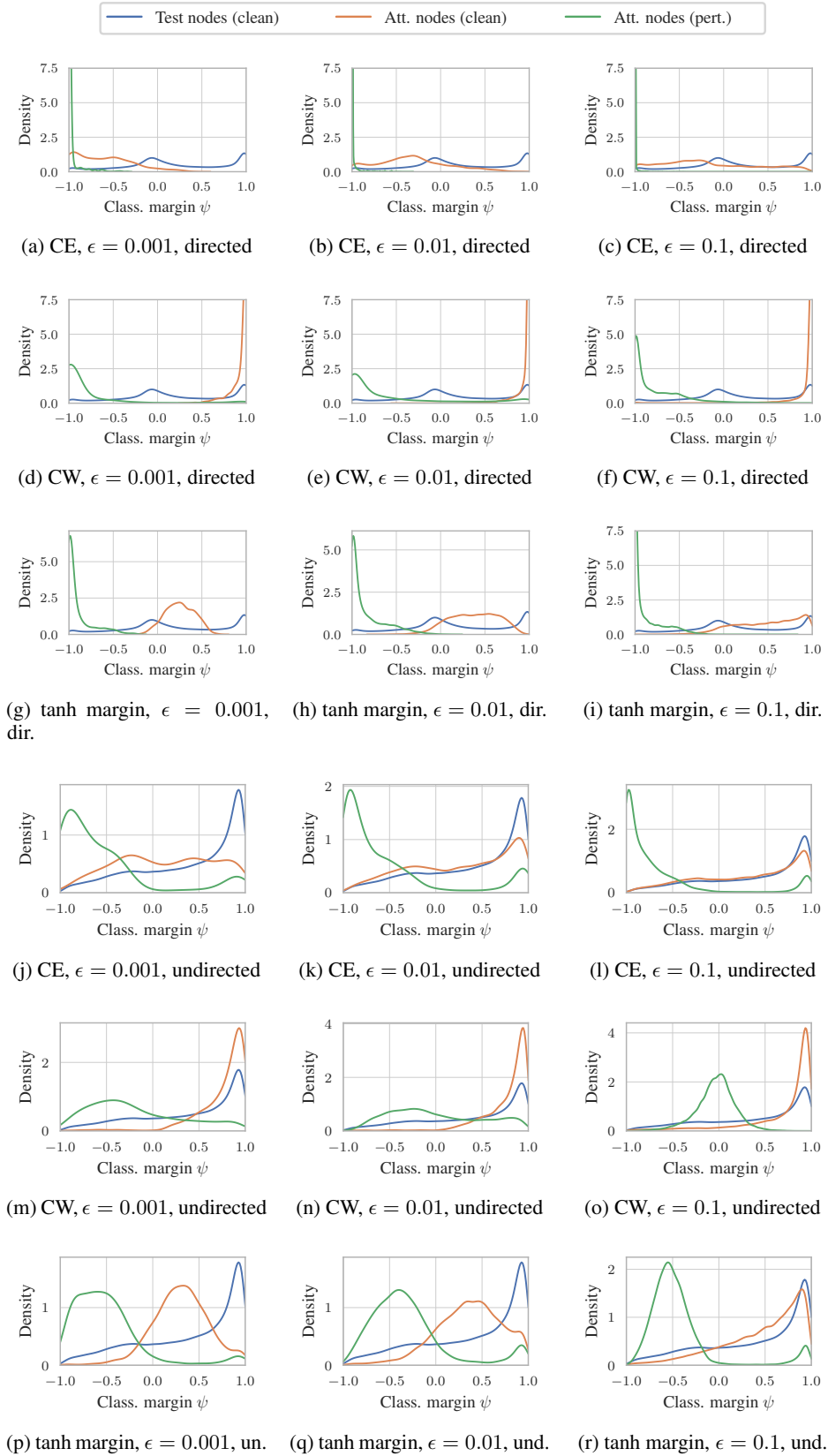

  \centering
  \hbox{\hspace{55pt}\resizebox{0.75\linewidth}{!}{
\begingroup%
\makeatletter%
\begin{pgfpicture}%
\pgfpathrectangle{\pgfpointorigin}{\pgfqpoint{4.491960in}{0.399973in}}%
\pgfusepath{use as bounding box, clip}%
\begin{pgfscope}%
\pgfsetbuttcap%
\pgfsetmiterjoin%
\definecolor{currentfill}{rgb}{1.000000,1.000000,1.000000}%
\pgfsetfillcolor{currentfill}%
\pgfsetlinewidth{0.000000pt}%
\definecolor{currentstroke}{rgb}{1.000000,1.000000,1.000000}%
\pgfsetstrokecolor{currentstroke}%
\pgfsetstrokeopacity{0.000000}%
\pgfsetdash{}{0pt}%
\pgfpathmoveto{\pgfqpoint{0.000000in}{0.000000in}}%
\pgfpathlineto{\pgfqpoint{4.491960in}{0.000000in}}%
\pgfpathlineto{\pgfqpoint{4.491960in}{0.399973in}}%
\pgfpathlineto{\pgfqpoint{0.000000in}{0.399973in}}%
\pgfpathclose%
\pgfusepath{fill}%
\end{pgfscope}%
\begin{pgfscope}%
\pgfsetbuttcap%
\pgfsetmiterjoin%
\definecolor{currentfill}{rgb}{1.000000,1.000000,1.000000}%
\pgfsetfillcolor{currentfill}%
\pgfsetfillopacity{0.800000}%
\pgfsetlinewidth{1.003750pt}%
\definecolor{currentstroke}{rgb}{0.800000,0.800000,0.800000}%
\pgfsetstrokecolor{currentstroke}%
\pgfsetstrokeopacity{0.800000}%
\pgfsetdash{}{0pt}%
\pgfpathmoveto{\pgfqpoint{0.122222in}{0.100000in}}%
\pgfpathlineto{\pgfqpoint{4.369738in}{0.100000in}}%
\pgfpathquadraticcurveto{\pgfqpoint{4.391960in}{0.100000in}}{\pgfqpoint{4.391960in}{0.122222in}}%
\pgfpathlineto{\pgfqpoint{4.391960in}{0.277751in}}%
\pgfpathquadraticcurveto{\pgfqpoint{4.391960in}{0.299973in}}{\pgfqpoint{4.369738in}{0.299973in}}%
\pgfpathlineto{\pgfqpoint{0.122222in}{0.299973in}}%
\pgfpathquadraticcurveto{\pgfqpoint{0.100000in}{0.299973in}}{\pgfqpoint{0.100000in}{0.277751in}}%
\pgfpathlineto{\pgfqpoint{0.100000in}{0.122222in}}%
\pgfpathquadraticcurveto{\pgfqpoint{0.100000in}{0.100000in}}{\pgfqpoint{0.122222in}{0.100000in}}%
\pgfpathclose%
\pgfusepath{stroke,fill}%
\end{pgfscope}%
\begin{pgfscope}%
\pgfsetroundcap%
\pgfsetroundjoin%
\pgfsetlinewidth{1.003750pt}%
\definecolor{currentstroke}{rgb}{0.298039,0.447059,0.690196}%
\pgfsetstrokecolor{currentstroke}%
\pgfsetdash{}{0pt}%
\pgfpathmoveto{\pgfqpoint{0.144444in}{0.211104in}}%
\pgfpathlineto{\pgfqpoint{0.366667in}{0.211104in}}%
\pgfusepath{stroke}%
\end{pgfscope}%
\begin{pgfscope}%
\definecolor{textcolor}{rgb}{0.150000,0.150000,0.150000}%
\pgfsetstrokecolor{textcolor}%
\pgfsetfillcolor{textcolor}%
\pgftext[x=0.455556in,y=0.172215in,left,base]{\color{textcolor}\rmfamily\fontsize{8.000000}{9.600000}\selectfont Test nodes (clean)}%
\end{pgfscope}%
\begin{pgfscope}%
\pgfsetroundcap%
\pgfsetroundjoin%
\pgfsetlinewidth{1.003750pt}%
\definecolor{currentstroke}{rgb}{0.866667,0.517647,0.321569}%
\pgfsetstrokecolor{currentstroke}%
\pgfsetdash{}{0pt}%
\pgfpathmoveto{\pgfqpoint{1.623303in}{0.211104in}}%
\pgfpathlineto{\pgfqpoint{1.845525in}{0.211104in}}%
\pgfusepath{stroke}%
\end{pgfscope}%
\begin{pgfscope}%
\definecolor{textcolor}{rgb}{0.150000,0.150000,0.150000}%
\pgfsetstrokecolor{textcolor}%
\pgfsetfillcolor{textcolor}%
\pgftext[x=1.934414in,y=0.172215in,left,base]{\color{textcolor}\rmfamily\fontsize{8.000000}{9.600000}\selectfont Att. nodes (clean)}%
\end{pgfscope}%
\begin{pgfscope}%
\pgfsetroundcap%
\pgfsetroundjoin%
\pgfsetlinewidth{1.003750pt}%
\definecolor{currentstroke}{rgb}{0.333333,0.658824,0.407843}%
\pgfsetstrokecolor{currentstroke}%
\pgfsetdash{}{0pt}%
\pgfpathmoveto{\pgfqpoint{3.104669in}{0.211104in}}%
\pgfpathlineto{\pgfqpoint{3.326891in}{0.211104in}}%
\pgfusepath{stroke}%
\end{pgfscope}%
\begin{pgfscope}%
\definecolor{textcolor}{rgb}{0.150000,0.150000,0.150000}%
\pgfsetstrokecolor{textcolor}%
\pgfsetfillcolor{textcolor}%
\pgftext[x=3.415780in,y=0.172215in,left,base]{\color{textcolor}\rmfamily\fontsize{8.000000}{9.600000}\selectfont Att. nodes (pert.)}%
\end{pgfscope}%
\end{pgfpicture}%
\makeatother%
\endgroup%
}}
  \vspace{-10pt}
  \makebox[\linewidth][c]{
    \(\begin{array}{ccc}
      \subfloat[CE, $\epsilon=0.001$, directed]{\resizebox{0.28\linewidth}{!}{\input{assets/global_prbcd_kde_nl_ogbn-arxiv_False_0.001_CE_margin_attacked.pgf}}} &
      \subfloat[CE, $\epsilon=0.01$, directed]{\resizebox{0.28\linewidth}{!}{\input{assets/global_prbcd_kde_nl_ogbn-arxiv_False_0.01_CE_margin_attacked.pgf}}} &
      \subfloat[CE, $\epsilon=0.1$, directed]{\resizebox{0.28\linewidth}{!}{\input{assets/global_prbcd_kde_nl_ogbn-arxiv_False_0.1_CE_margin_attacked.pgf}}} \\
      \subfloat[CW, $\epsilon=0.001$, directed]{\resizebox{0.28\linewidth}{!}{\input{assets/global_prbcd_kde_nl_ogbn-arxiv_False_0.001_CW_margin_attacked.pgf}}} &
      \subfloat[CW, $\epsilon=0.01$, directed]{\resizebox{0.28\linewidth}{!}{\input{assets/global_prbcd_kde_nl_ogbn-arxiv_False_0.01_CW_margin_attacked.pgf}}} &
      \subfloat[CW, $\epsilon=0.1$, directed]{\resizebox{0.28\linewidth}{!}{\input{assets/global_prbcd_kde_nl_ogbn-arxiv_False_0.1_CW_margin_attacked.pgf}}} \\
      \subfloat[tanh margin, $\epsilon=0.001$, dir.]{\resizebox{0.28\linewidth}{!}{\input{assets/global_prbcd_kde_nl_ogbn-arxiv_False_0.001_tanhMargin_margin_attacked.pgf}}} &
      \subfloat[tanh margin, $\epsilon=0.01$, dir.]{\resizebox{0.28\linewidth}{!}{\input{assets/global_prbcd_kde_nl_ogbn-arxiv_False_0.01_tanhMargin_margin_attacked.pgf}}} &
      \subfloat[tanh margin, $\epsilon=0.1$, dir.]{\resizebox{0.28\linewidth}{!}{\input{assets/global_prbcd_kde_nl_ogbn-arxiv_False_0.1_tanhMargin_margin_attacked.pgf}}} \\
      \subfloat[CE, $\epsilon=0.001$, undirected]{\resizebox{0.28\linewidth}{!}{\input{assets/global_prbcd_kde_nl_ogbn-arxiv_True_0.001_CE_margin_attacked.pgf}}} &
      \subfloat[CE, $\epsilon=0.01$, undirected]{\resizebox{0.28\linewidth}{!}{\input{assets/global_prbcd_kde_nl_ogbn-arxiv_True_0.01_CE_margin_attacked.pgf}}} &
      \subfloat[CE, $\epsilon=0.1$, undirected]{\resizebox{0.28\linewidth}{!}{\input{assets/global_prbcd_kde_nl_ogbn-arxiv_True_0.1_CE_margin_attacked.pgf}}} \\
      \subfloat[CW, $\epsilon=0.001$, undirected]{\resizebox{0.28\linewidth}{!}{\input{assets/global_prbcd_kde_nl_ogbn-arxiv_True_0.001_CW_margin_attacked.pgf}}} &
      \subfloat[CW, $\epsilon=0.01$, undirected]{\resizebox{0.28\linewidth}{!}{\input{assets/global_prbcd_kde_nl_ogbn-arxiv_True_0.01_CW_margin_attacked.pgf}}} &
      \subfloat[CW, $\epsilon=0.1$, undirected]{\resizebox{0.28\linewidth}{!}{\input{assets/global_prbcd_kde_nl_ogbn-arxiv_True_0.1_CW_margin_attacked.pgf}}} \\
      \subfloat[tanh margin, $\epsilon=0.001$, un.]{\resizebox{0.28\linewidth}{!}{\input{assets/global_prbcd_kde_nl_ogbn-arxiv_True_0.001_tanhMargin_margin_attacked.pgf}}} &
      \subfloat[tanh margin, $\epsilon=0.01$, und.]{\resizebox{0.28\linewidth}{!}{\input{assets/global_prbcd_kde_nl_ogbn-arxiv_True_0.01_tanhMargin_margin_attacked.pgf}}} &
      \subfloat[tanh margin, $\epsilon=0.1$, und.]{\resizebox{0.28\linewidth}{!}{\input{assets/global_prbcd_kde_nl_ogbn-arxiv_True_0.1_tanhMargin_margin_attacked.pgf}}} \\
    \end{array}\)
  }
  \caption{Distribution of nodes attacked before/after PGD attack on Vanilla GCN on arXiv.}%
  \label{fig:appendix_margin_densities_arxiv}
\end{figure}

\subsection{Impact of Surrogate Losses on Attack Strength}\label{sec:appendix_surrogate_attack_strength}

In \autoref{tab:appendix_losscompare_cora_ml} and \autoref{tab:appendix_losscompare_citeseer}, we additionally evaluate how the different losses perform for other models than the Vanilla GCN. Baring a few exceptions, we conclude that our analysis and choice of losses is model agnostic and that our claims and observations hold also for the other architectures.

\begin{figure}[ht!]
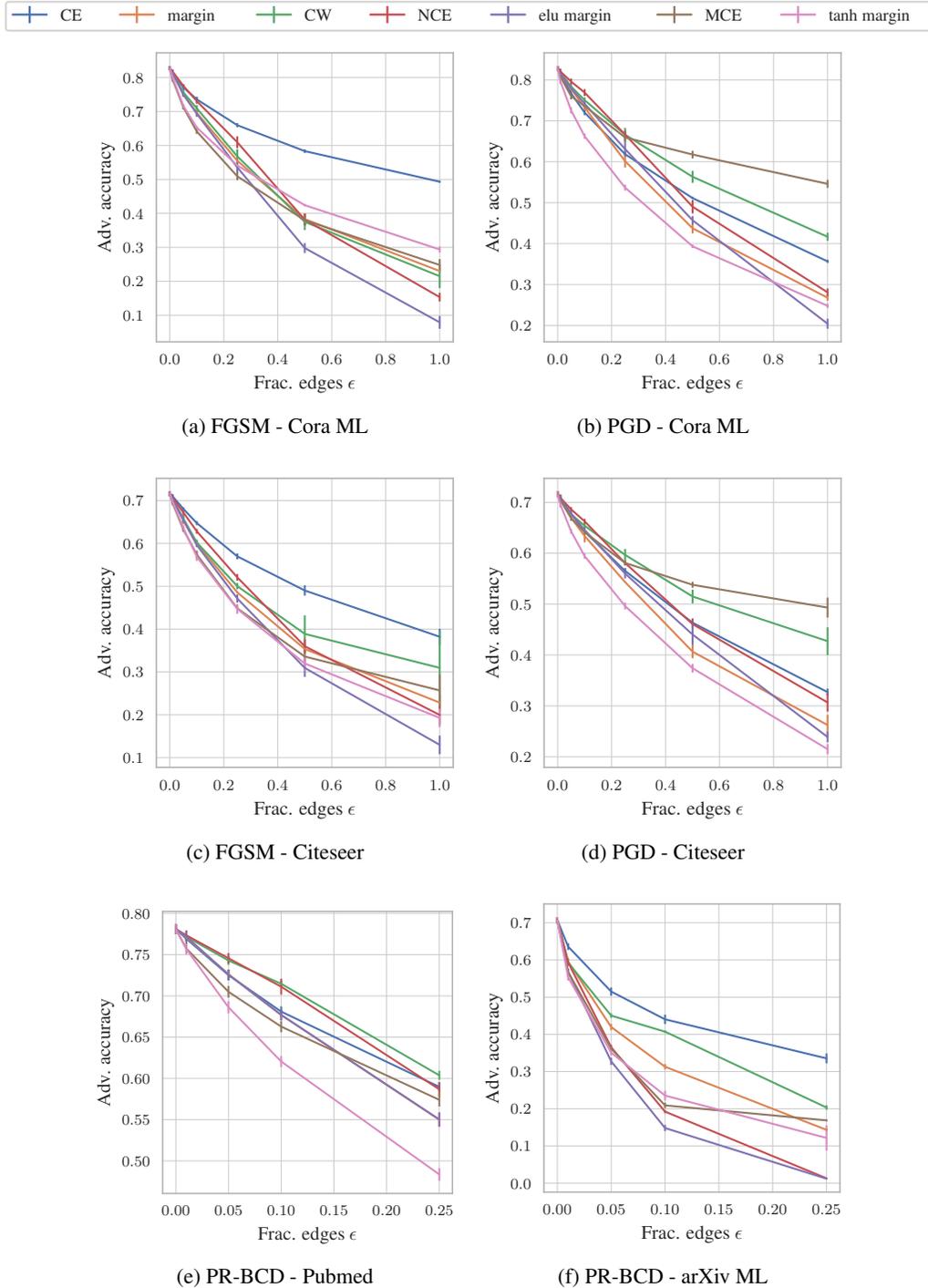

  \centering
  \hbox{\resizebox{\linewidth}{!}{
\begingroup%
\makeatletter%
\begin{pgfpicture}%
\pgfpathrectangle{\pgfpointorigin}{\pgfqpoint{6.247184in}{0.388266in}}%
\pgfusepath{use as bounding box, clip}%
\begin{pgfscope}%
\pgfsetbuttcap%
\pgfsetmiterjoin%
\definecolor{currentfill}{rgb}{1.000000,1.000000,1.000000}%
\pgfsetfillcolor{currentfill}%
\pgfsetlinewidth{0.000000pt}%
\definecolor{currentstroke}{rgb}{1.000000,1.000000,1.000000}%
\pgfsetstrokecolor{currentstroke}%
\pgfsetstrokeopacity{0.000000}%
\pgfsetdash{}{0pt}%
\pgfpathmoveto{\pgfqpoint{0.000000in}{0.000000in}}%
\pgfpathlineto{\pgfqpoint{6.247184in}{0.000000in}}%
\pgfpathlineto{\pgfqpoint{6.247184in}{0.388266in}}%
\pgfpathlineto{\pgfqpoint{0.000000in}{0.388266in}}%
\pgfpathclose%
\pgfusepath{fill}%
\end{pgfscope}%
\begin{pgfscope}%
\pgfsetbuttcap%
\pgfsetmiterjoin%
\definecolor{currentfill}{rgb}{1.000000,1.000000,1.000000}%
\pgfsetfillcolor{currentfill}%
\pgfsetfillopacity{0.800000}%
\pgfsetlinewidth{1.003750pt}%
\definecolor{currentstroke}{rgb}{0.800000,0.800000,0.800000}%
\pgfsetstrokecolor{currentstroke}%
\pgfsetstrokeopacity{0.800000}%
\pgfsetdash{}{0pt}%
\pgfpathmoveto{\pgfqpoint{0.122222in}{0.100000in}}%
\pgfpathlineto{\pgfqpoint{6.124962in}{0.100000in}}%
\pgfpathquadraticcurveto{\pgfqpoint{6.147184in}{0.100000in}}{\pgfqpoint{6.147184in}{0.122222in}}%
\pgfpathlineto{\pgfqpoint{6.147184in}{0.266044in}}%
\pgfpathquadraticcurveto{\pgfqpoint{6.147184in}{0.288266in}}{\pgfqpoint{6.124962in}{0.288266in}}%
\pgfpathlineto{\pgfqpoint{0.122222in}{0.288266in}}%
\pgfpathquadraticcurveto{\pgfqpoint{0.100000in}{0.288266in}}{\pgfqpoint{0.100000in}{0.266044in}}%
\pgfpathlineto{\pgfqpoint{0.100000in}{0.122222in}}%
\pgfpathquadraticcurveto{\pgfqpoint{0.100000in}{0.100000in}}{\pgfqpoint{0.122222in}{0.100000in}}%
\pgfpathclose%
\pgfusepath{stroke,fill}%
\end{pgfscope}%
\begin{pgfscope}%
\pgfsetbuttcap%
\pgfsetroundjoin%
\pgfsetlinewidth{1.003750pt}%
\definecolor{currentstroke}{rgb}{0.298039,0.447059,0.690196}%
\pgfsetstrokecolor{currentstroke}%
\pgfsetdash{}{0pt}%
\pgfpathmoveto{\pgfqpoint{0.255556in}{0.149378in}}%
\pgfpathlineto{\pgfqpoint{0.255556in}{0.260489in}}%
\pgfusepath{stroke}%
\end{pgfscope}%
\begin{pgfscope}%
\pgfsetroundcap%
\pgfsetroundjoin%
\pgfsetlinewidth{1.003750pt}%
\definecolor{currentstroke}{rgb}{0.298039,0.447059,0.690196}%
\pgfsetstrokecolor{currentstroke}%
\pgfsetdash{}{0pt}%
\pgfpathmoveto{\pgfqpoint{0.144444in}{0.204933in}}%
\pgfpathlineto{\pgfqpoint{0.366667in}{0.204933in}}%
\pgfusepath{stroke}%
\end{pgfscope}%
\begin{pgfscope}%
\definecolor{textcolor}{rgb}{0.150000,0.150000,0.150000}%
\pgfsetstrokecolor{textcolor}%
\pgfsetfillcolor{textcolor}%
\pgftext[x=0.455556in,y=0.166044in,left,base]{\color{textcolor}\rmfamily\fontsize{8.000000}{9.600000}\selectfont CE}%
\end{pgfscope}%
\begin{pgfscope}%
\pgfsetbuttcap%
\pgfsetroundjoin%
\pgfsetlinewidth{1.003750pt}%
\definecolor{currentstroke}{rgb}{0.866667,0.517647,0.321569}%
\pgfsetstrokecolor{currentstroke}%
\pgfsetdash{}{0pt}%
\pgfpathmoveto{\pgfqpoint{0.954396in}{0.149378in}}%
\pgfpathlineto{\pgfqpoint{0.954396in}{0.260489in}}%
\pgfusepath{stroke}%
\end{pgfscope}%
\begin{pgfscope}%
\pgfsetroundcap%
\pgfsetroundjoin%
\pgfsetlinewidth{1.003750pt}%
\definecolor{currentstroke}{rgb}{0.866667,0.517647,0.321569}%
\pgfsetstrokecolor{currentstroke}%
\pgfsetdash{}{0pt}%
\pgfpathmoveto{\pgfqpoint{0.843285in}{0.204933in}}%
\pgfpathlineto{\pgfqpoint{1.065507in}{0.204933in}}%
\pgfusepath{stroke}%
\end{pgfscope}%
\begin{pgfscope}%
\definecolor{textcolor}{rgb}{0.150000,0.150000,0.150000}%
\pgfsetstrokecolor{textcolor}%
\pgfsetfillcolor{textcolor}%
\pgftext[x=1.154396in,y=0.166044in,left,base]{\color{textcolor}\rmfamily\fontsize{8.000000}{9.600000}\selectfont margin}%
\end{pgfscope}%
\begin{pgfscope}%
\pgfsetbuttcap%
\pgfsetroundjoin%
\pgfsetlinewidth{1.003750pt}%
\definecolor{currentstroke}{rgb}{0.333333,0.658824,0.407843}%
\pgfsetstrokecolor{currentstroke}%
\pgfsetdash{}{0pt}%
\pgfpathmoveto{\pgfqpoint{1.848463in}{0.149378in}}%
\pgfpathlineto{\pgfqpoint{1.848463in}{0.260489in}}%
\pgfusepath{stroke}%
\end{pgfscope}%
\begin{pgfscope}%
\pgfsetroundcap%
\pgfsetroundjoin%
\pgfsetlinewidth{1.003750pt}%
\definecolor{currentstroke}{rgb}{0.333333,0.658824,0.407843}%
\pgfsetstrokecolor{currentstroke}%
\pgfsetdash{}{0pt}%
\pgfpathmoveto{\pgfqpoint{1.737352in}{0.204933in}}%
\pgfpathlineto{\pgfqpoint{1.959574in}{0.204933in}}%
\pgfusepath{stroke}%
\end{pgfscope}%
\begin{pgfscope}%
\definecolor{textcolor}{rgb}{0.150000,0.150000,0.150000}%
\pgfsetstrokecolor{textcolor}%
\pgfsetfillcolor{textcolor}%
\pgftext[x=2.048463in,y=0.166044in,left,base]{\color{textcolor}\rmfamily\fontsize{8.000000}{9.600000}\selectfont CW}%
\end{pgfscope}%
\begin{pgfscope}%
\pgfsetbuttcap%
\pgfsetroundjoin%
\pgfsetlinewidth{1.003750pt}%
\definecolor{currentstroke}{rgb}{0.768627,0.305882,0.321569}%
\pgfsetstrokecolor{currentstroke}%
\pgfsetdash{}{0pt}%
\pgfpathmoveto{\pgfqpoint{2.588228in}{0.149378in}}%
\pgfpathlineto{\pgfqpoint{2.588228in}{0.260489in}}%
\pgfusepath{stroke}%
\end{pgfscope}%
\begin{pgfscope}%
\pgfsetroundcap%
\pgfsetroundjoin%
\pgfsetlinewidth{1.003750pt}%
\definecolor{currentstroke}{rgb}{0.768627,0.305882,0.321569}%
\pgfsetstrokecolor{currentstroke}%
\pgfsetdash{}{0pt}%
\pgfpathmoveto{\pgfqpoint{2.477117in}{0.204933in}}%
\pgfpathlineto{\pgfqpoint{2.699339in}{0.204933in}}%
\pgfusepath{stroke}%
\end{pgfscope}%
\begin{pgfscope}%
\definecolor{textcolor}{rgb}{0.150000,0.150000,0.150000}%
\pgfsetstrokecolor{textcolor}%
\pgfsetfillcolor{textcolor}%
\pgftext[x=2.788228in,y=0.166044in,left,base]{\color{textcolor}\rmfamily\fontsize{8.000000}{9.600000}\selectfont NCE}%
\end{pgfscope}%
\begin{pgfscope}%
\pgfsetbuttcap%
\pgfsetroundjoin%
\pgfsetlinewidth{1.003750pt}%
\definecolor{currentstroke}{rgb}{0.505882,0.447059,0.701961}%
\pgfsetstrokecolor{currentstroke}%
\pgfsetdash{}{0pt}%
\pgfpathmoveto{\pgfqpoint{3.375473in}{0.149378in}}%
\pgfpathlineto{\pgfqpoint{3.375473in}{0.260489in}}%
\pgfusepath{stroke}%
\end{pgfscope}%
\begin{pgfscope}%
\pgfsetroundcap%
\pgfsetroundjoin%
\pgfsetlinewidth{1.003750pt}%
\definecolor{currentstroke}{rgb}{0.505882,0.447059,0.701961}%
\pgfsetstrokecolor{currentstroke}%
\pgfsetdash{}{0pt}%
\pgfpathmoveto{\pgfqpoint{3.264362in}{0.204933in}}%
\pgfpathlineto{\pgfqpoint{3.486584in}{0.204933in}}%
\pgfusepath{stroke}%
\end{pgfscope}%
\begin{pgfscope}%
\definecolor{textcolor}{rgb}{0.150000,0.150000,0.150000}%
\pgfsetstrokecolor{textcolor}%
\pgfsetfillcolor{textcolor}%
\pgftext[x=3.575473in,y=0.166044in,left,base]{\color{textcolor}\rmfamily\fontsize{8.000000}{9.600000}\selectfont elu margin}%
\end{pgfscope}%
\begin{pgfscope}%
\pgfsetbuttcap%
\pgfsetroundjoin%
\pgfsetlinewidth{1.003750pt}%
\definecolor{currentstroke}{rgb}{0.576471,0.470588,0.376471}%
\pgfsetstrokecolor{currentstroke}%
\pgfsetdash{}{0pt}%
\pgfpathmoveto{\pgfqpoint{4.459694in}{0.149378in}}%
\pgfpathlineto{\pgfqpoint{4.459694in}{0.260489in}}%
\pgfusepath{stroke}%
\end{pgfscope}%
\begin{pgfscope}%
\pgfsetroundcap%
\pgfsetroundjoin%
\pgfsetlinewidth{1.003750pt}%
\definecolor{currentstroke}{rgb}{0.576471,0.470588,0.376471}%
\pgfsetstrokecolor{currentstroke}%
\pgfsetdash{}{0pt}%
\pgfpathmoveto{\pgfqpoint{4.348583in}{0.204933in}}%
\pgfpathlineto{\pgfqpoint{4.570805in}{0.204933in}}%
\pgfusepath{stroke}%
\end{pgfscope}%
\begin{pgfscope}%
\definecolor{textcolor}{rgb}{0.150000,0.150000,0.150000}%
\pgfsetstrokecolor{textcolor}%
\pgfsetfillcolor{textcolor}%
\pgftext[x=4.659694in,y=0.166044in,left,base]{\color{textcolor}\rmfamily\fontsize{8.000000}{9.600000}\selectfont MCE}%
\end{pgfscope}%
\begin{pgfscope}%
\pgfsetbuttcap%
\pgfsetroundjoin%
\pgfsetlinewidth{1.003750pt}%
\definecolor{currentstroke}{rgb}{0.854902,0.545098,0.764706}%
\pgfsetstrokecolor{currentstroke}%
\pgfsetdash{}{0pt}%
\pgfpathmoveto{\pgfqpoint{5.266610in}{0.149378in}}%
\pgfpathlineto{\pgfqpoint{5.266610in}{0.260489in}}%
\pgfusepath{stroke}%
\end{pgfscope}%
\begin{pgfscope}%
\pgfsetroundcap%
\pgfsetroundjoin%
\pgfsetlinewidth{1.003750pt}%
\definecolor{currentstroke}{rgb}{0.854902,0.545098,0.764706}%
\pgfsetstrokecolor{currentstroke}%
\pgfsetdash{}{0pt}%
\pgfpathmoveto{\pgfqpoint{5.155499in}{0.204933in}}%
\pgfpathlineto{\pgfqpoint{5.377721in}{0.204933in}}%
\pgfusepath{stroke}%
\end{pgfscope}%
\begin{pgfscope}%
\definecolor{textcolor}{rgb}{0.150000,0.150000,0.150000}%
\pgfsetstrokecolor{textcolor}%
\pgfsetfillcolor{textcolor}%
\pgftext[x=5.466610in,y=0.166044in,left,base]{\color{textcolor}\rmfamily\fontsize{8.000000}{9.600000}\selectfont tanh margin}%
\end{pgfscope}%
\end{pgfpicture}%
\makeatother%
\endgroup%
}}
  \vspace{-14pt}
  \makebox[\linewidth][c]{
    \(\arraycolsep=1pt\def\arraystretch{2}\begin{array}{cc}
      \subfloat[FGSM - Cora ML]{\resizebox{0.4\linewidth}{!}{\input{assets/global_transfer_FGSM_cora_ml_VanillaGCN_surrloss_large_no_legend.pgf}}} &
      \subfloat[PGD - Cora ML]{\resizebox{0.4\linewidth}{!}{\input{assets/global_transfer_PGD_cora_ml_VanillaGCN_surrloss_large_no_legend.pgf}}}\\
      \subfloat[FGSM - Citeseer]{\resizebox{0.4\linewidth}{!}{\input{assets/global_transfer_FGSM_citeseer_VanillaGCN_surrloss_large_no_legend.pgf}}} &
      \subfloat[PGD - Citeseer]{\resizebox{0.4\linewidth}{!}{\input{assets/global_transfer_PGD_citeseer_VanillaGCN_surrloss_large_no_legend.pgf}}} \\
      \subfloat[PR-BCD - Pubmed]{\resizebox{0.4\linewidth}{!}{\input{assets/global_transfer_PRBCD_pubmed_VanillaGCN_surrloss_large_no_legend.pgf}}} &
      \subfloat[PR-BCD - arXiv ML]{\resizebox{0.4\linewidth}{!}{\input{assets/global_transfer_PRBCD_ogbn-arxiv_VanillaGCN_surrloss_large_no_legend.pgf}}}\\
    \end{array}\)
  }
  \caption{Comparison of the losses on a Vanilla GCN attacked with a greedy attack and a projected gradient / coordinate descent algorithm. \(\epsilon\) denotes the fraction of edges perturbed (relative to the clean graph). The lower the adversarial accuracy the better the loss.\label{fig:appendix_empsurrogate}}
\end{figure}

\begin{table}[H]
\centering
\caption{Adversarial accuracy comparing the conventional losses with our losses over the different architectures on Cora ML (transfer attack). \(\epsilon\) denotes the fraction of edges perturbed. %
}
\label{tab:appendix_losscompare_cora_ml}
\vskip 0.11in
\resizebox{\linewidth}{!}{
\begin{tabular}{lcl|cccccccccc|cccccccccc}
\toprule
                             &      & \textbf{Architecture} & \makecell{\underline{Soft}\\\underline{Median}\\\underline{GDC}} & \makecell{\underline{Soft}\\\underline{Median}\\\underline{PPRGo}} &     \makecell{Vanilla\\GCN} &     \makecell{Vanilla\\GDC} &   \makecell{Vanilla\\PPRGo} & \makecell{Soft\\Medoid\\GDC} &     \makecell{Jaccard\\GCN} &             \makecell{RGCN} \\
\midrule
\multirow{28}{*}{\rotatebox{90}{\textbf{FGSM}}} & \multirow{7}{*}{\rotatebox{90}{$\epsilon=0.01$}} & CE &                                  0.813 $\pm$ 0.002 &                                  0.816 $\pm$ 0.000 &           0.814 $\pm$ 0.004 &           0.826 $\pm$ 0.002 &           0.818 $\pm$ 0.002 &            0.810 $\pm$ 0.003 &           0.806 $\pm$ 0.003 &           0.807 $\pm$ 0.002 \\
                             &                                 & margin &                                  0.820 $\pm$ 0.001 &                                  0.820 $\pm$ 0.001 &           0.813 $\pm$ 0.003 &           0.825 $\pm$ 0.003 &           0.818 $\pm$ 0.002 &            0.816 $\pm$ 0.002 &           0.804 $\pm$ 0.003 &           0.804 $\pm$ 0.002 \\
                             &                                 & CW &                                  0.820 $\pm$ 0.001 &                                  0.819 $\pm$ 0.001 &           0.814 $\pm$ 0.003 &           0.826 $\pm$ 0.003 &           0.818 $\pm$ 0.001 &            0.816 $\pm$ 0.002 &           0.805 $\pm$ 0.003 &           0.804 $\pm$ 0.003 \\
                             &                                 & NCE &                                  0.822 $\pm$ 0.001 &                                  0.820 $\pm$ 0.001 &           0.818 $\pm$ 0.003 &           0.831 $\pm$ 0.003 &           0.822 $\pm$ 0.002 &            0.818 $\pm$ 0.002 &           0.809 $\pm$ 0.002 &           0.807 $\pm$ 0.003 \\
                             &                                 & elu margin &                                  0.821 $\pm$ 0.001 &                                  0.819 $\pm$ 0.001 &           0.814 $\pm$ 0.003 &           0.826 $\pm$ 0.003 &           0.817 $\pm$ 0.002 &            0.817 $\pm$ 0.002 &           0.804 $\pm$ 0.003 &           0.804 $\pm$ 0.002 \\
                             &                                 & \underline{MCE} &                                  0.811 $\pm$ 0.002 &                                  0.813 $\pm$ 0.001 &  \textbf{0.795 $\pm$ 0.004} &           0.811 $\pm$ 0.003 &           0.807 $\pm$ 0.001 &            0.808 $\pm$ 0.002 &  \textbf{0.791 $\pm$ 0.003} &  \textbf{0.794 $\pm$ 0.000} \\
                             &                                 & \underline{tanh margin} &                         \textbf{0.806 $\pm$ 0.001} &                         \textbf{0.811 $\pm$ 0.001} &           0.801 $\pm$ 0.003 &  \textbf{0.810 $\pm$ 0.003} &  \textbf{0.803 $\pm$ 0.002} &   \textbf{0.807 $\pm$ 0.003} &           0.794 $\pm$ 0.002 &           0.796 $\pm$ 0.001 \\
\cline{2-11}
                             & \multirow{7}{*}{\rotatebox{90}{$\epsilon=0.05$}} & CE &                                  0.779 $\pm$ 0.001 &                                  0.789 $\pm$ 0.000 &           0.771 $\pm$ 0.004 &           0.776 $\pm$ 0.001 &           0.781 $\pm$ 0.002 &            0.776 $\pm$ 0.001 &           0.768 $\pm$ 0.003 &           0.764 $\pm$ 0.001 \\
                             &                                 & margin &                                  0.799 $\pm$ 0.002 &                                  0.803 $\pm$ 0.001 &           0.751 $\pm$ 0.004 &           0.763 $\pm$ 0.004 &           0.774 $\pm$ 0.001 &            0.799 $\pm$ 0.002 &           0.748 $\pm$ 0.003 &           0.745 $\pm$ 0.003 \\
                             &                                 & CW &                                  0.804 $\pm$ 0.002 &                                  0.804 $\pm$ 0.000 &           0.757 $\pm$ 0.004 &           0.775 $\pm$ 0.005 &           0.779 $\pm$ 0.001 &            0.805 $\pm$ 0.003 &           0.754 $\pm$ 0.003 &           0.752 $\pm$ 0.002 \\
                             &                                 & NCE &                                  0.811 $\pm$ 0.002 &                                  0.812 $\pm$ 0.000 &           0.776 $\pm$ 0.002 &           0.794 $\pm$ 0.006 &           0.791 $\pm$ 0.001 &            0.809 $\pm$ 0.002 &           0.770 $\pm$ 0.002 &           0.767 $\pm$ 0.004 \\
                             &                                 & elu margin &                                  0.801 $\pm$ 0.001 &                                  0.802 $\pm$ 0.001 &           0.750 $\pm$ 0.003 &           0.766 $\pm$ 0.005 &           0.773 $\pm$ 0.001 &            0.803 $\pm$ 0.002 &           0.746 $\pm$ 0.002 &           0.745 $\pm$ 0.001 \\
                             &                                 & \underline{MCE} &                                  0.784 $\pm$ 0.001 &                                  0.792 $\pm$ 0.000 &  \textbf{0.713 $\pm$ 0.003} &           0.738 $\pm$ 0.004 &           0.754 $\pm$ 0.002 &            0.784 $\pm$ 0.003 &           0.722 $\pm$ 0.002 &           0.719 $\pm$ 0.004 \\
                             &                                 & \underline{tanh margin} &                         \textbf{0.767 $\pm$ 0.001} &                         \textbf{0.783 $\pm$ 0.001} &           0.717 $\pm$ 0.002 &  \textbf{0.726 $\pm$ 0.001} &  \textbf{0.737 $\pm$ 0.003} &   \textbf{0.772 $\pm$ 0.002} &  \textbf{0.720 $\pm$ 0.002} &  \textbf{0.718 $\pm$ 0.005} \\
\cline{2-11}
                             & \multirow{7}{*}{\rotatebox{90}{$\epsilon=0.1$}} & CE &                                  0.753 $\pm$ 0.002 &                         \textbf{0.764 $\pm$ 0.001} &           0.736 $\pm$ 0.004 &           0.740 $\pm$ 0.002 &           0.749 $\pm$ 0.003 &            0.751 $\pm$ 0.001 &           0.735 $\pm$ 0.003 &           0.729 $\pm$ 0.002 \\
                             &                                 & margin &                                  0.776 $\pm$ 0.001 &                                  0.780 $\pm$ 0.001 &           0.696 $\pm$ 0.004 &           0.708 $\pm$ 0.006 &           0.728 $\pm$ 0.001 &            0.780 $\pm$ 0.003 &           0.703 $\pm$ 0.004 &           0.691 $\pm$ 0.003 \\
                             &                                 & CW &                                  0.792 $\pm$ 0.002 &                                  0.787 $\pm$ 0.002 &           0.709 $\pm$ 0.005 &           0.735 $\pm$ 0.007 &           0.744 $\pm$ 0.001 &            0.792 $\pm$ 0.003 &           0.710 $\pm$ 0.004 &           0.704 $\pm$ 0.003 \\
                             &                                 & NCE &                                  0.793 $\pm$ 0.002 &                                  0.792 $\pm$ 0.001 &           0.731 $\pm$ 0.004 &           0.751 $\pm$ 0.007 &           0.760 $\pm$ 0.002 &            0.796 $\pm$ 0.003 &           0.731 $\pm$ 0.003 &           0.727 $\pm$ 0.005 \\
                             &                                 & elu margin &                                  0.788 $\pm$ 0.002 &                                  0.783 $\pm$ 0.002 &           0.693 $\pm$ 0.004 &           0.717 $\pm$ 0.007 &           0.732 $\pm$ 0.002 &            0.791 $\pm$ 0.003 &           0.698 $\pm$ 0.003 &           0.695 $\pm$ 0.002 \\
                             &                                 & \underline{MCE} &                                  0.769 $\pm$ 0.002 &                                  0.778 $\pm$ 0.001 &  \textbf{0.641 $\pm$ 0.003} &           0.672 $\pm$ 0.005 &           0.724 $\pm$ 0.003 &            0.773 $\pm$ 0.005 &  \textbf{0.661 $\pm$ 0.002} &  \textbf{0.654 $\pm$ 0.007} \\
                             &                                 & \underline{tanh margin} &                         \textbf{0.733 $\pm$ 0.001} &                                  0.765 $\pm$ 0.001 &           0.653 $\pm$ 0.002 &  \textbf{0.665 $\pm$ 0.000} &  \textbf{0.690 $\pm$ 0.003} &   \textbf{0.744 $\pm$ 0.003} &           0.662 $\pm$ 0.003 &           0.660 $\pm$ 0.006 \\
\cline{2-11}
                             & \multirow{7}{*}{\rotatebox{90}{$\epsilon=0.25$}} & CE &                                  0.687 $\pm$ 0.000 &                         \textbf{0.709 $\pm$ 0.002} &           0.660 $\pm$ 0.003 &           0.665 $\pm$ 0.002 &           0.681 $\pm$ 0.002 &   \textbf{0.687 $\pm$ 0.002} &           0.664 $\pm$ 0.002 &           0.657 $\pm$ 0.001 \\
                             &                                 & margin &                                  0.729 $\pm$ 0.003 &                                  0.741 $\pm$ 0.002 &           0.555 $\pm$ 0.008 &           0.580 $\pm$ 0.006 &           0.648 $\pm$ 0.004 &            0.738 $\pm$ 0.005 &           0.586 $\pm$ 0.007 &           0.557 $\pm$ 0.003 \\
                             &                                 & CW &                                  0.769 $\pm$ 0.003 &                                  0.765 $\pm$ 0.001 &           0.568 $\pm$ 0.010 &           0.625 $\pm$ 0.011 &           0.689 $\pm$ 0.003 &            0.777 $\pm$ 0.004 &           0.598 $\pm$ 0.006 &           0.577 $\pm$ 0.002 \\
                             &                                 & NCE &                                  0.764 $\pm$ 0.003 &                                  0.758 $\pm$ 0.002 &           0.609 $\pm$ 0.008 &           0.637 $\pm$ 0.008 &           0.687 $\pm$ 0.002 &            0.771 $\pm$ 0.003 &           0.629 $\pm$ 0.006 &           0.603 $\pm$ 0.008 \\
                             &                                 & elu margin &                                  0.756 $\pm$ 0.004 &                                  0.754 $\pm$ 0.002 &           0.535 $\pm$ 0.009 &           0.586 $\pm$ 0.008 &           0.664 $\pm$ 0.004 &            0.765 $\pm$ 0.003 &           0.576 $\pm$ 0.006 &           0.544 $\pm$ 0.001 \\
                             &                                 & \underline{MCE} &                                  0.750 $\pm$ 0.003 &                                  0.762 $\pm$ 0.001 &  \textbf{0.509 $\pm$ 0.005} &           0.575 $\pm$ 0.009 &           0.683 $\pm$ 0.003 &            0.762 $\pm$ 0.005 &           0.557 $\pm$ 0.001 &  \textbf{0.535 $\pm$ 0.015} \\
                             &                                 & \underline{tanh margin} &                         \textbf{0.679 $\pm$ 0.002} &                                  0.733 $\pm$ 0.002 &           0.541 $\pm$ 0.001 &  \textbf{0.554 $\pm$ 0.001} &  \textbf{0.610 $\pm$ 0.002} &            0.690 $\pm$ 0.005 &  \textbf{0.551 $\pm$ 0.001} &           0.553 $\pm$ 0.007 \\
\cline{1-11}
\cline{2-11}
\multirow{28}{*}{\rotatebox{90}{\textbf{PGD}}} & \multirow{7}{*}{\rotatebox{90}{$\epsilon=0.01$}} & CE &                                  0.813 $\pm$ 0.002 &                                  0.814 $\pm$ 0.001 &           0.815 $\pm$ 0.004 &           0.824 $\pm$ 0.002 &           0.815 $\pm$ 0.001 &   \textbf{0.810 $\pm$ 0.002} &           0.805 $\pm$ 0.003 &           0.805 $\pm$ 0.002 \\
                             &                                 & margin &                                  0.820 $\pm$ 0.002 &                                  0.820 $\pm$ 0.001 &           0.816 $\pm$ 0.003 &           0.830 $\pm$ 0.003 &           0.819 $\pm$ 0.002 &            0.816 $\pm$ 0.002 &           0.806 $\pm$ 0.003 &           0.807 $\pm$ 0.002 \\
                             &                                 & CW &                                  0.821 $\pm$ 0.002 &                                  0.820 $\pm$ 0.001 &           0.818 $\pm$ 0.003 &           0.833 $\pm$ 0.003 &           0.820 $\pm$ 0.001 &            0.817 $\pm$ 0.002 &           0.809 $\pm$ 0.003 &           0.810 $\pm$ 0.002 \\
                             &                                 & NCE &                                  0.821 $\pm$ 0.001 &                                  0.820 $\pm$ 0.001 &           0.821 $\pm$ 0.003 &           0.834 $\pm$ 0.004 &           0.821 $\pm$ 0.001 &            0.818 $\pm$ 0.002 &           0.811 $\pm$ 0.003 &           0.812 $\pm$ 0.002 \\
                             &                                 & elu margin &                                  0.820 $\pm$ 0.002 &                                  0.820 $\pm$ 0.001 &           0.816 $\pm$ 0.003 &           0.832 $\pm$ 0.003 &           0.819 $\pm$ 0.001 &            0.818 $\pm$ 0.002 &           0.807 $\pm$ 0.003 &           0.807 $\pm$ 0.002 \\
                             &                                 & \underline{MCE} &                                  0.817 $\pm$ 0.001 &                                  0.815 $\pm$ 0.001 &           0.808 $\pm$ 0.002 &           0.824 $\pm$ 0.004 &           0.813 $\pm$ 0.001 &            0.813 $\pm$ 0.002 &           0.800 $\pm$ 0.002 &           0.803 $\pm$ 0.001 \\
                             &                                 & \underline{tanh margin} &                         \textbf{0.809 $\pm$ 0.001} &                         \textbf{0.811 $\pm$ 0.001} &  \textbf{0.797 $\pm$ 0.003} &  \textbf{0.808 $\pm$ 0.002} &  \textbf{0.800 $\pm$ 0.002} &            0.810 $\pm$ 0.002 &  \textbf{0.792 $\pm$ 0.003} &  \textbf{0.792 $\pm$ 0.000} \\
\cline{2-11}
                             & \multirow{7}{*}{\rotatebox{90}{$\epsilon=0.05$}} & CE &                         \textbf{0.775 $\pm$ 0.002} &                         \textbf{0.786 $\pm$ 0.001} &           0.767 $\pm$ 0.003 &           0.773 $\pm$ 0.002 &           0.771 $\pm$ 0.001 &   \textbf{0.777 $\pm$ 0.002} &           0.761 $\pm$ 0.002 &           0.759 $\pm$ 0.001 \\
                             &                                 & margin &                                  0.804 $\pm$ 0.001 &                                  0.808 $\pm$ 0.000 &           0.778 $\pm$ 0.003 &           0.790 $\pm$ 0.004 &           0.789 $\pm$ 0.002 &            0.803 $\pm$ 0.002 &           0.773 $\pm$ 0.003 &           0.773 $\pm$ 0.002 \\
                             &                                 & CW &                                  0.810 $\pm$ 0.001 &                                  0.810 $\pm$ 0.000 &           0.784 $\pm$ 0.003 &           0.795 $\pm$ 0.004 &           0.798 $\pm$ 0.001 &            0.808 $\pm$ 0.002 &           0.779 $\pm$ 0.003 &           0.773 $\pm$ 0.003 \\
                             &                                 & NCE &                                  0.814 $\pm$ 0.001 &                                  0.812 $\pm$ 0.001 &           0.796 $\pm$ 0.003 &           0.809 $\pm$ 0.004 &           0.806 $\pm$ 0.001 &            0.810 $\pm$ 0.002 &           0.790 $\pm$ 0.003 &           0.788 $\pm$ 0.003 \\
                             &                                 & elu margin &                                  0.808 $\pm$ 0.001 &                                  0.809 $\pm$ 0.001 &           0.780 $\pm$ 0.003 &           0.797 $\pm$ 0.006 &           0.793 $\pm$ 0.001 &            0.806 $\pm$ 0.003 &           0.777 $\pm$ 0.002 &           0.775 $\pm$ 0.002 \\
                             &                                 & \underline{MCE} &                                  0.800 $\pm$ 0.001 &                                  0.801 $\pm$ 0.002 &           0.760 $\pm$ 0.004 &           0.776 $\pm$ 0.004 &           0.783 $\pm$ 0.002 &            0.801 $\pm$ 0.002 &           0.762 $\pm$ 0.003 &           0.762 $\pm$ 0.000 \\
                             &                                 & \underline{tanh margin} &                                  0.779 $\pm$ 0.002 &                                  0.787 $\pm$ 0.001 &  \textbf{0.726 $\pm$ 0.004} &  \textbf{0.740 $\pm$ 0.002} &  \textbf{0.748 $\pm$ 0.002} &            0.782 $\pm$ 0.003 &  \textbf{0.730 $\pm$ 0.003} &  \textbf{0.725 $\pm$ 0.005} \\
\cline{2-11}
                             & \multirow{7}{*}{\rotatebox{90}{$\epsilon=0.1$}} & CE &                         \textbf{0.745 $\pm$ 0.002} &                         \textbf{0.762 $\pm$ 0.002} &           0.720 $\pm$ 0.003 &           0.724 $\pm$ 0.002 &           0.733 $\pm$ 0.002 &   \textbf{0.748 $\pm$ 0.001} &           0.720 $\pm$ 0.003 &           0.719 $\pm$ 0.001 \\
                             &                                 & margin &                                  0.786 $\pm$ 0.002 &                                  0.789 $\pm$ 0.001 &           0.733 $\pm$ 0.003 &           0.743 $\pm$ 0.006 &           0.753 $\pm$ 0.001 &            0.787 $\pm$ 0.002 &           0.730 $\pm$ 0.003 &           0.724 $\pm$ 0.004 \\
                             &                                 & CW &                                  0.799 $\pm$ 0.001 &                                  0.799 $\pm$ 0.000 &           0.751 $\pm$ 0.003 &           0.767 $\pm$ 0.006 &           0.776 $\pm$ 0.001 &            0.796 $\pm$ 0.002 &           0.752 $\pm$ 0.003 &           0.745 $\pm$ 0.004 \\
                             &                                 & NCE &                                  0.804 $\pm$ 0.001 &                                  0.805 $\pm$ 0.000 &           0.769 $\pm$ 0.004 &           0.786 $\pm$ 0.004 &           0.789 $\pm$ 0.002 &            0.801 $\pm$ 0.001 &           0.765 $\pm$ 0.003 &           0.758 $\pm$ 0.005 \\
                             &                                 & elu margin &                                  0.802 $\pm$ 0.001 &                                  0.797 $\pm$ 0.001 &           0.742 $\pm$ 0.004 &           0.765 $\pm$ 0.006 &           0.767 $\pm$ 0.001 &            0.797 $\pm$ 0.002 &           0.743 $\pm$ 0.003 &           0.732 $\pm$ 0.004 \\
                             &                                 & \underline{MCE} &                                  0.792 $\pm$ 0.002 &                                  0.791 $\pm$ 0.001 &           0.736 $\pm$ 0.003 &           0.751 $\pm$ 0.005 &           0.775 $\pm$ 0.000 &            0.793 $\pm$ 0.002 &           0.737 $\pm$ 0.003 &           0.732 $\pm$ 0.002 \\
                             &                                 & \underline{tanh margin} &                                  0.758 $\pm$ 0.002 &                                  0.769 $\pm$ 0.001 &  \textbf{0.662 $\pm$ 0.003} &  \textbf{0.679 $\pm$ 0.002} &  \textbf{0.704 $\pm$ 0.001} &            0.759 $\pm$ 0.003 &  \textbf{0.673 $\pm$ 0.002} &  \textbf{0.671 $\pm$ 0.007} \\
\cline{2-11}
                             & \multirow{7}{*}{\rotatebox{90}{$\epsilon=0.25$}} & CE &                         \textbf{0.684 $\pm$ 0.001} &                         \textbf{0.692 $\pm$ 0.001} &           0.618 $\pm$ 0.003 &           0.626 $\pm$ 0.002 &           0.641 $\pm$ 0.003 &   \textbf{0.688 $\pm$ 0.002} &           0.624 $\pm$ 0.003 &           0.617 $\pm$ 0.001 \\
                             &                                 & margin &                                  0.744 $\pm$ 0.003 &                                  0.748 $\pm$ 0.003 &           0.601 $\pm$ 0.007 &           0.638 $\pm$ 0.005 &           0.671 $\pm$ 0.003 &            0.752 $\pm$ 0.004 &           0.621 $\pm$ 0.004 &           0.599 $\pm$ 0.004 \\
                             &                                 & CW &                                  0.779 $\pm$ 0.003 &                                  0.778 $\pm$ 0.002 &           0.666 $\pm$ 0.008 &           0.707 $\pm$ 0.007 &           0.744 $\pm$ 0.002 &            0.778 $\pm$ 0.004 &           0.681 $\pm$ 0.007 &           0.661 $\pm$ 0.006 \\
                             &                                 & NCE &                                  0.774 $\pm$ 0.002 &                                  0.773 $\pm$ 0.000 &           0.667 $\pm$ 0.005 &           0.700 $\pm$ 0.007 &           0.731 $\pm$ 0.000 &            0.777 $\pm$ 0.002 &           0.679 $\pm$ 0.004 &           0.659 $\pm$ 0.006 \\
                             &                                 & elu margin &                                  0.776 $\pm$ 0.003 &                                  0.774 $\pm$ 0.001 &           0.631 $\pm$ 0.007 &           0.679 $\pm$ 0.008 &           0.725 $\pm$ 0.002 &            0.775 $\pm$ 0.004 &           0.656 $\pm$ 0.005 &           0.635 $\pm$ 0.004 \\
                             &                                 & \underline{MCE} &                                  0.772 $\pm$ 0.003 &                                  0.777 $\pm$ 0.002 &           0.660 $\pm$ 0.006 &           0.694 $\pm$ 0.007 &           0.752 $\pm$ 0.002 &            0.778 $\pm$ 0.003 &           0.666 $\pm$ 0.005 &           0.664 $\pm$ 0.003 \\
                             &                                 & \underline{tanh margin} &                                  0.716 $\pm$ 0.004 &                                  0.739 $\pm$ 0.001 &  \textbf{0.537 $\pm$ 0.003} &  \textbf{0.567 $\pm$ 0.005} &  \textbf{0.632 $\pm$ 0.002} &            0.727 $\pm$ 0.004 &  \textbf{0.560 $\pm$ 0.001} &  \textbf{0.547 $\pm$ 0.011} \\
\bottomrule
\end{tabular}
}
\end{table}

\begin{table}[H]
\centering
\caption{Adversarial accuracy comparing the conventional losses with our losses over the different architectures on Citeseer (transfer attack). \(\epsilon\) denotes the fraction of edges perturbed. 
}
\label{tab:appendix_losscompare_citeseer}
\vskip 0.11in
\resizebox{1\linewidth}{!}{
\begin{tabular}{lcl|ccccccccc}
\toprule
                             &      & \textbf{Architecture} & \makecell{\underline{Soft}\\\underline{Median}\\\underline{GDC}} & \makecell{\underline{Soft}\\\underline{Median}\\\underline{PPRGo}} &     \makecell{Vanilla\\GCN} &     \makecell{Vanilla\\GDC} &   \makecell{Vanilla\\PPRGo} & \makecell{Soft\\Medoid\\GDC} &     \makecell{Jaccard\\GCN} &             \makecell{RGCN} \\
\midrule
\multirow{28}{*}{\rotatebox{90}{\textbf{FGSM}}} & \multirow{7}{*}{\rotatebox{90}{$\epsilon=0.01$}} & CE &                                  0.705 $\pm$ 0.002 &                                  0.712 $\pm$ 0.006 &           0.710 $\pm$ 0.002 &           0.699 $\pm$ 0.001 &           0.720 $\pm$ 0.005 &            0.705 $\pm$ 0.003 &           0.716 $\pm$ 0.004 &           0.681 $\pm$ 0.005 \\
                             &                                 & margin &                                  0.708 $\pm$ 0.002 &                                  0.712 $\pm$ 0.006 &           0.704 $\pm$ 0.003 &           0.694 $\pm$ 0.002 &           0.720 $\pm$ 0.006 &            0.707 $\pm$ 0.003 &           0.712 $\pm$ 0.005 &           0.673 $\pm$ 0.005 \\
                             &                                 & CW &                                  0.708 $\pm$ 0.002 &                                  0.712 $\pm$ 0.006 &           0.705 $\pm$ 0.003 &           0.694 $\pm$ 0.002 &           0.720 $\pm$ 0.006 &            0.707 $\pm$ 0.002 &           0.711 $\pm$ 0.005 &           0.673 $\pm$ 0.005 \\
                             &                                 & NCE &                                  0.709 $\pm$ 0.002 &                                  0.714 $\pm$ 0.006 &           0.707 $\pm$ 0.003 &           0.696 $\pm$ 0.002 &           0.722 $\pm$ 0.006 &            0.708 $\pm$ 0.003 &           0.714 $\pm$ 0.005 &           0.675 $\pm$ 0.006 \\
                             &                                 & elu margin &                                  0.708 $\pm$ 0.002 &                                  0.712 $\pm$ 0.006 &           0.704 $\pm$ 0.003 &           0.694 $\pm$ 0.002 &           0.719 $\pm$ 0.006 &            0.706 $\pm$ 0.003 &           0.711 $\pm$ 0.005 &           0.673 $\pm$ 0.005 \\
                             &                                 & \underline{MCE} &                                  0.703 $\pm$ 0.003 &                                  0.712 $\pm$ 0.006 &  \textbf{0.695 $\pm$ 0.003} &           0.686 $\pm$ 0.001 &           0.715 $\pm$ 0.006 &            0.702 $\pm$ 0.002 &  \textbf{0.707 $\pm$ 0.004} &  \textbf{0.672 $\pm$ 0.004} \\
                             &                                 & \underline{tanh margin} &                         \textbf{0.702 $\pm$ 0.002} &                         \textbf{0.710 $\pm$ 0.006} &           0.698 $\pm$ 0.003 &  \textbf{0.685 $\pm$ 0.000} &  \textbf{0.710 $\pm$ 0.005} &   \textbf{0.701 $\pm$ 0.003} &           0.708 $\pm$ 0.005 &           0.672 $\pm$ 0.004 \\
\cline{2-11}
                             & \multirow{7}{*}{\rotatebox{90}{$\epsilon=0.05$}} & CE &                                  0.688 $\pm$ 0.003 &                                  0.699 $\pm$ 0.006 &           0.681 $\pm$ 0.002 &           0.664 $\pm$ 0.002 &           0.698 $\pm$ 0.004 &            0.688 $\pm$ 0.003 &           0.693 $\pm$ 0.004 &           0.654 $\pm$ 0.004 \\
                             &                                 & margin &                                  0.701 $\pm$ 0.002 &                                  0.701 $\pm$ 0.008 &           0.654 $\pm$ 0.004 &           0.639 $\pm$ 0.003 &           0.691 $\pm$ 0.008 &            0.700 $\pm$ 0.003 &           0.672 $\pm$ 0.007 &           0.622 $\pm$ 0.004 \\
                             &                                 & CW &                                  0.702 $\pm$ 0.002 &                                  0.703 $\pm$ 0.008 &           0.658 $\pm$ 0.004 &           0.646 $\pm$ 0.002 &           0.692 $\pm$ 0.008 &            0.702 $\pm$ 0.004 &           0.677 $\pm$ 0.005 &           0.626 $\pm$ 0.004 \\
                             &                                 & NCE &                                  0.706 $\pm$ 0.002 &                                  0.705 $\pm$ 0.008 &           0.673 $\pm$ 0.004 &           0.661 $\pm$ 0.001 &           0.699 $\pm$ 0.007 &            0.706 $\pm$ 0.003 &           0.687 $\pm$ 0.006 &           0.635 $\pm$ 0.005 \\
                             &                                 & elu margin &                                  0.703 $\pm$ 0.002 &                                  0.702 $\pm$ 0.008 &           0.655 $\pm$ 0.004 &           0.642 $\pm$ 0.002 &           0.689 $\pm$ 0.008 &            0.703 $\pm$ 0.003 &           0.674 $\pm$ 0.005 &           0.624 $\pm$ 0.004 \\
                             &                                 & \underline{MCE} &                                  0.695 $\pm$ 0.002 &                                  0.697 $\pm$ 0.006 &  \textbf{0.633 $\pm$ 0.003} &  \textbf{0.622 $\pm$ 0.002} &           0.677 $\pm$ 0.008 &            0.694 $\pm$ 0.004 &  \textbf{0.663 $\pm$ 0.004} &  \textbf{0.622 $\pm$ 0.003} \\
                             &                                 & \underline{tanh margin} &                         \textbf{0.681 $\pm$ 0.002} &                         \textbf{0.693 $\pm$ 0.006} &           0.636 $\pm$ 0.003 &           0.630 $\pm$ 0.002 &  \textbf{0.667 $\pm$ 0.005} &   \textbf{0.685 $\pm$ 0.003} &           0.664 $\pm$ 0.005 &           0.622 $\pm$ 0.003 \\
\cline{2-11}
                             & \multirow{7}{*}{\rotatebox{90}{$\epsilon=0.1$}} & CE &                                  0.666 $\pm$ 0.003 &                                  0.684 $\pm$ 0.007 &           0.648 $\pm$ 0.002 &           0.631 $\pm$ 0.003 &           0.669 $\pm$ 0.005 &   \textbf{0.670 $\pm$ 0.004} &           0.666 $\pm$ 0.003 &           0.621 $\pm$ 0.004 \\
                             &                                 & margin &                                  0.688 $\pm$ 0.003 &                                  0.692 $\pm$ 0.007 &           0.600 $\pm$ 0.003 &           0.579 $\pm$ 0.004 &           0.648 $\pm$ 0.008 &            0.689 $\pm$ 0.004 &           0.632 $\pm$ 0.006 &           0.574 $\pm$ 0.003 \\
                             &                                 & CW &                                  0.693 $\pm$ 0.003 &                                  0.694 $\pm$ 0.007 &           0.602 $\pm$ 0.003 &           0.589 $\pm$ 0.004 &           0.660 $\pm$ 0.008 &            0.695 $\pm$ 0.003 &           0.639 $\pm$ 0.005 &           0.581 $\pm$ 0.004 \\
                             &                                 & NCE &                                  0.699 $\pm$ 0.002 &                                  0.697 $\pm$ 0.008 &           0.628 $\pm$ 0.003 &           0.621 $\pm$ 0.001 &           0.678 $\pm$ 0.008 &            0.700 $\pm$ 0.003 &           0.656 $\pm$ 0.005 &           0.593 $\pm$ 0.005 \\
                             &                                 & elu margin &                                  0.693 $\pm$ 0.002 &                                  0.693 $\pm$ 0.008 &           0.596 $\pm$ 0.002 &           0.581 $\pm$ 0.004 &           0.655 $\pm$ 0.008 &            0.696 $\pm$ 0.003 &           0.634 $\pm$ 0.004 &           0.575 $\pm$ 0.003 \\
                             &                                 & \underline{MCE} &                                  0.676 $\pm$ 0.002 &                                  0.685 $\pm$ 0.007 &           0.574 $\pm$ 0.004 &  \textbf{0.562 $\pm$ 0.003} &           0.644 $\pm$ 0.009 &            0.682 $\pm$ 0.003 &           0.622 $\pm$ 0.006 &  \textbf{0.568 $\pm$ 0.005} \\
                             &                                 & \underline{tanh margin} &                         \textbf{0.663 $\pm$ 0.003} &                         \textbf{0.683 $\pm$ 0.006} &  \textbf{0.569 $\pm$ 0.005} &           0.568 $\pm$ 0.004 &  \textbf{0.621 $\pm$ 0.006} &            0.672 $\pm$ 0.004 &  \textbf{0.609 $\pm$ 0.005} &           0.569 $\pm$ 0.006 \\
\cline{2-11}
                             & \multirow{7}{*}{\rotatebox{90}{$\epsilon=0.25$}} & CE &                         \textbf{0.616 $\pm$ 0.003} &                         \textbf{0.652 $\pm$ 0.007} &           0.570 $\pm$ 0.003 &           0.555 $\pm$ 0.005 &           0.614 $\pm$ 0.006 &   \textbf{0.624 $\pm$ 0.005} &           0.606 $\pm$ 0.004 &           0.551 $\pm$ 0.003 \\
                             &                                 & margin &                                  0.659 $\pm$ 0.004 &                                  0.663 $\pm$ 0.010 &           0.486 $\pm$ 0.001 &           0.463 $\pm$ 0.006 &           0.577 $\pm$ 0.008 &            0.668 $\pm$ 0.006 &           0.550 $\pm$ 0.000 &           0.473 $\pm$ 0.006 \\
                             &                                 & CW &                                  0.682 $\pm$ 0.002 &                                  0.674 $\pm$ 0.009 &           0.500 $\pm$ 0.003 &           0.495 $\pm$ 0.014 &           0.615 $\pm$ 0.007 &            0.686 $\pm$ 0.003 &           0.569 $\pm$ 0.002 &           0.478 $\pm$ 0.003 \\
                             &                                 & NCE &                                  0.681 $\pm$ 0.002 &                                  0.676 $\pm$ 0.008 &           0.521 $\pm$ 0.004 &           0.519 $\pm$ 0.012 &           0.613 $\pm$ 0.006 &            0.689 $\pm$ 0.003 &           0.585 $\pm$ 0.004 &           0.487 $\pm$ 0.004 \\
                             &                                 & elu margin &                                  0.677 $\pm$ 0.002 &                                  0.670 $\pm$ 0.008 &           0.471 $\pm$ 0.004 &           0.460 $\pm$ 0.012 &           0.595 $\pm$ 0.008 &            0.685 $\pm$ 0.004 &           0.553 $\pm$ 0.002 &  \textbf{0.455 $\pm$ 0.005} \\
                             &                                 & \underline{MCE} &                                  0.658 $\pm$ 0.002 &                                  0.671 $\pm$ 0.009 &           0.448 $\pm$ 0.003 &  \textbf{0.439 $\pm$ 0.005} &           0.605 $\pm$ 0.009 &            0.675 $\pm$ 0.004 &           0.545 $\pm$ 0.003 &           0.461 $\pm$ 0.008 \\
                             &                                 & \underline{tanh margin} &                                  0.631 $\pm$ 0.002 &                                  0.666 $\pm$ 0.007 &  \textbf{0.447 $\pm$ 0.005} &           0.456 $\pm$ 0.006 &  \textbf{0.548 $\pm$ 0.005} &            0.649 $\pm$ 0.003 &  \textbf{0.511 $\pm$ 0.002} &           0.462 $\pm$ 0.012 \\
\cline{1-11}
\cline{2-11}
\multirow{28}{*}{\rotatebox{90}{\textbf{PGD}}} & \multirow{7}{*}{\rotatebox{90}{$\epsilon=0.01$}} & CE &                                  0.705 $\pm$ 0.002 &                                  0.712 $\pm$ 0.006 &           0.710 $\pm$ 0.003 &           0.699 $\pm$ 0.001 &           0.720 $\pm$ 0.005 &   \textbf{0.703 $\pm$ 0.002} &           0.714 $\pm$ 0.005 &           0.680 $\pm$ 0.005 \\
                             &                                 & margin &                                  0.707 $\pm$ 0.002 &                                  0.712 $\pm$ 0.006 &           0.706 $\pm$ 0.004 &           0.694 $\pm$ 0.002 &           0.719 $\pm$ 0.006 &            0.707 $\pm$ 0.003 &           0.714 $\pm$ 0.006 &           0.675 $\pm$ 0.005 \\
                             &                                 & CW &                                  0.708 $\pm$ 0.002 &                                  0.712 $\pm$ 0.006 &           0.708 $\pm$ 0.003 &           0.697 $\pm$ 0.001 &           0.721 $\pm$ 0.006 &            0.707 $\pm$ 0.003 &           0.715 $\pm$ 0.005 &           0.677 $\pm$ 0.005 \\
                             &                                 & NCE &                                  0.708 $\pm$ 0.002 &                                  0.714 $\pm$ 0.006 &           0.709 $\pm$ 0.003 &           0.700 $\pm$ 0.002 &           0.723 $\pm$ 0.006 &            0.708 $\pm$ 0.003 &           0.716 $\pm$ 0.005 &           0.679 $\pm$ 0.006 \\
                             &                                 & elu margin &                                  0.708 $\pm$ 0.002 &                                  0.713 $\pm$ 0.006 &           0.707 $\pm$ 0.003 &           0.696 $\pm$ 0.002 &           0.720 $\pm$ 0.006 &            0.707 $\pm$ 0.003 &           0.714 $\pm$ 0.005 &           0.677 $\pm$ 0.006 \\
                             &                                 & \underline{MCE} &                                  0.706 $\pm$ 0.002 &                                  0.712 $\pm$ 0.006 &           0.704 $\pm$ 0.003 &           0.694 $\pm$ 0.000 &           0.720 $\pm$ 0.006 &            0.706 $\pm$ 0.003 &           0.713 $\pm$ 0.004 &           0.675 $\pm$ 0.005 \\
                             &                                 & \underline{tanh margin} &                         \textbf{0.703 $\pm$ 0.002} &                         \textbf{0.711 $\pm$ 0.006} &  \textbf{0.696 $\pm$ 0.003} &  \textbf{0.685 $\pm$ 0.000} &  \textbf{0.712 $\pm$ 0.006} &            0.703 $\pm$ 0.003 &  \textbf{0.706 $\pm$ 0.005} &  \textbf{0.670 $\pm$ 0.005} \\
\cline{2-11}
                             & \multirow{7}{*}{\rotatebox{90}{$\epsilon=0.05$}} & CE &                                  0.689 $\pm$ 0.002 &                         \textbf{0.697 $\pm$ 0.007} &           0.677 $\pm$ 0.002 &           0.661 $\pm$ 0.002 &           0.695 $\pm$ 0.005 &            0.689 $\pm$ 0.005 &           0.688 $\pm$ 0.004 &           0.647 $\pm$ 0.003 \\
                             &                                 & margin &                                  0.702 $\pm$ 0.002 &                                  0.702 $\pm$ 0.007 &           0.670 $\pm$ 0.003 &           0.654 $\pm$ 0.002 &           0.693 $\pm$ 0.007 &            0.701 $\pm$ 0.003 &           0.688 $\pm$ 0.004 &           0.642 $\pm$ 0.004 \\
                             &                                 & CW &                                  0.704 $\pm$ 0.002 &                                  0.707 $\pm$ 0.006 &           0.676 $\pm$ 0.002 &           0.661 $\pm$ 0.001 &           0.707 $\pm$ 0.007 &            0.702 $\pm$ 0.002 &           0.693 $\pm$ 0.004 &           0.646 $\pm$ 0.005 \\
                             &                                 & NCE &                                  0.706 $\pm$ 0.002 &                                  0.708 $\pm$ 0.007 &           0.686 $\pm$ 0.002 &           0.676 $\pm$ 0.001 &           0.708 $\pm$ 0.005 &            0.705 $\pm$ 0.003 &           0.700 $\pm$ 0.004 &           0.654 $\pm$ 0.005 \\
                             &                                 & elu margin &                                  0.704 $\pm$ 0.002 &                                  0.703 $\pm$ 0.007 &           0.678 $\pm$ 0.001 &           0.660 $\pm$ 0.002 &           0.698 $\pm$ 0.005 &            0.705 $\pm$ 0.002 &           0.693 $\pm$ 0.004 &           0.649 $\pm$ 0.005 \\
                             &                                 & \underline{MCE} &                                  0.696 $\pm$ 0.001 &                                  0.702 $\pm$ 0.006 &           0.669 $\pm$ 0.002 &           0.645 $\pm$ 0.003 &           0.700 $\pm$ 0.006 &            0.698 $\pm$ 0.003 &           0.688 $\pm$ 0.004 &           0.639 $\pm$ 0.006 \\
                             &                                 & \underline{tanh margin} &                         \textbf{0.686 $\pm$ 0.002} &                                  0.700 $\pm$ 0.005 &  \textbf{0.643 $\pm$ 0.002} &  \textbf{0.627 $\pm$ 0.001} &  \textbf{0.675 $\pm$ 0.004} &   \textbf{0.688 $\pm$ 0.002} &  \textbf{0.666 $\pm$ 0.004} &  \textbf{0.629 $\pm$ 0.003} \\
\cline{2-11}
                             & \multirow{7}{*}{\rotatebox{90}{$\epsilon=0.1$}} & CE &                         \textbf{0.663 $\pm$ 0.001} &                         \textbf{0.681 $\pm$ 0.006} &           0.643 $\pm$ 0.003 &           0.622 $\pm$ 0.001 &           0.664 $\pm$ 0.005 &   \textbf{0.665 $\pm$ 0.004} &           0.662 $\pm$ 0.003 &           0.621 $\pm$ 0.002 \\
                             &                                 & margin &                                  0.691 $\pm$ 0.002 &                                  0.696 $\pm$ 0.008 &           0.634 $\pm$ 0.006 &           0.615 $\pm$ 0.002 &           0.668 $\pm$ 0.008 &            0.690 $\pm$ 0.004 &           0.661 $\pm$ 0.006 &           0.606 $\pm$ 0.004 \\
                             &                                 & CW &                                  0.697 $\pm$ 0.001 &                                  0.701 $\pm$ 0.007 &           0.654 $\pm$ 0.002 &           0.635 $\pm$ 0.005 &           0.691 $\pm$ 0.005 &            0.698 $\pm$ 0.003 &           0.676 $\pm$ 0.005 &           0.620 $\pm$ 0.005 \\
                             &                                 & NCE &                                  0.697 $\pm$ 0.002 &                                  0.702 $\pm$ 0.008 &           0.663 $\pm$ 0.002 &           0.647 $\pm$ 0.004 &           0.691 $\pm$ 0.007 &            0.700 $\pm$ 0.002 &           0.681 $\pm$ 0.005 &           0.624 $\pm$ 0.006 \\
                             &                                 & elu margin &                                  0.700 $\pm$ 0.002 &                                  0.699 $\pm$ 0.008 &           0.646 $\pm$ 0.002 &           0.626 $\pm$ 0.003 &           0.684 $\pm$ 0.006 &            0.699 $\pm$ 0.003 &           0.670 $\pm$ 0.004 &           0.612 $\pm$ 0.006 \\
                             &                                 & \underline{MCE} &                                  0.687 $\pm$ 0.001 &                                  0.691 $\pm$ 0.007 &           0.641 $\pm$ 0.002 &           0.605 $\pm$ 0.006 &           0.684 $\pm$ 0.006 &            0.690 $\pm$ 0.003 &           0.666 $\pm$ 0.005 &           0.614 $\pm$ 0.005 \\
                             &                                 & \underline{tanh margin} &                                  0.675 $\pm$ 0.002 &                                  0.692 $\pm$ 0.007 &  \textbf{0.594 $\pm$ 0.002} &  \textbf{0.581 $\pm$ 0.003} &  \textbf{0.649 $\pm$ 0.003} &            0.677 $\pm$ 0.003 &  \textbf{0.630 $\pm$ 0.003} &  \textbf{0.589 $\pm$ 0.004} \\
\cline{2-11}
                             & \multirow{7}{*}{\rotatebox{90}{$\epsilon=0.25$}} & CE &                         \textbf{0.617 $\pm$ 0.005} &                         \textbf{0.653 $\pm$ 0.008} &           0.565 $\pm$ 0.003 &           0.544 $\pm$ 0.001 &           0.605 $\pm$ 0.008 &   \textbf{0.627 $\pm$ 0.006} &           0.594 $\pm$ 0.004 &           0.550 $\pm$ 0.002 \\
                             &                                 & margin &                                  0.660 $\pm$ 0.004 &                                  0.671 $\pm$ 0.009 &           0.543 $\pm$ 0.000 &           0.512 $\pm$ 0.004 &           0.610 $\pm$ 0.006 &            0.670 $\pm$ 0.005 &           0.593 $\pm$ 0.004 &           0.522 $\pm$ 0.005 \\
                             &                                 & CW &                                  0.682 $\pm$ 0.002 &                                  0.685 $\pm$ 0.009 &           0.597 $\pm$ 0.006 &           0.575 $\pm$ 0.010 &           0.670 $\pm$ 0.006 &            0.687 $\pm$ 0.004 &           0.636 $\pm$ 0.005 &           0.557 $\pm$ 0.005 \\
                             &                                 & NCE &                                  0.683 $\pm$ 0.001 &                                  0.681 $\pm$ 0.008 &           0.581 $\pm$ 0.002 &           0.571 $\pm$ 0.007 &           0.651 $\pm$ 0.006 &            0.688 $\pm$ 0.003 &           0.623 $\pm$ 0.002 &           0.541 $\pm$ 0.007 \\
                             &                                 & elu margin &                                  0.681 $\pm$ 0.002 &                                  0.681 $\pm$ 0.009 &           0.560 $\pm$ 0.005 &           0.541 $\pm$ 0.010 &           0.650 $\pm$ 0.007 &            0.687 $\pm$ 0.003 &           0.612 $\pm$ 0.003 &           0.537 $\pm$ 0.005 \\
                             &                                 & \underline{MCE} &                                  0.670 $\pm$ 0.004 &                                  0.681 $\pm$ 0.008 &           0.581 $\pm$ 0.002 &           0.548 $\pm$ 0.007 &           0.665 $\pm$ 0.007 &            0.677 $\pm$ 0.006 &           0.624 $\pm$ 0.004 &           0.552 $\pm$ 0.004 \\
                             &                                 & \underline{tanh margin} &                                  0.649 $\pm$ 0.002 &                                  0.671 $\pm$ 0.006 &  \textbf{0.496 $\pm$ 0.003} &  \textbf{0.486 $\pm$ 0.002} &  \textbf{0.590 $\pm$ 0.007} &            0.658 $\pm$ 0.004 &  \textbf{0.553 $\pm$ 0.004} &  \textbf{0.497 $\pm$ 0.006} \\
\bottomrule
\end{tabular}
}
\end{table}

\subsection{Proof of \autoref{proposition:goodsurrogate}}\label{sec:appendix_proofprop1}

\setcounter{proposition}{0}
\begin{proposition}
  Let \(\mathcal{L}'\) be the surrogate for the 0/1 loss \(\mathcal{L}_{0/1}\) used to attack a node classification algorithm \(f_{\theta}(\adj, \features)\) with a global budget \(\Delta\). %
  Suppose we greedily attack nodes in order of \(\nicefrac{\partial \mathcal{L}'}{\partial \evz_{c^*}}(\psi_0) \le \nicefrac{\partial \mathcal{L}'}{\partial \evz_{c^*}}(\psi_1) \le \dots \le \nicefrac{\partial \mathcal{L}'}{\partial \evz_{c^*}}(\psi_l)\) until the budget is exhausted \(\Delta < \sum_{i=0}^{l+1} \Delta_i\).
  Under Assumptions 1 \& 2, we then obtain the global optimum of
  \(
    \max_{\tilde{\adj}\text{ s.t.\ }\|\tilde{\adj} - \adj\|_0 < \Delta} \mathcal{L}_{0/1}(f_{\theta}(\tilde{\adj}, \features))\,
  \)
  if \(\mathcal{L}'\) has the properties \textbf{\emph{(I)}} \(\nicefrac{\partial \mathcal{L}'}{\partial \evz_{c^*}} |_{\psi < 0} = 0\) and \textbf{\emph{(II)}} \(\nicefrac{\partial \mathcal{L}'}{\partial \evz_{c^*}} |_{\psi_0} < \nicefrac{\partial \mathcal{L}'}{\partial \evz_{c^*}} |_{\psi_1}\) for any \(0 < \psi_0 < \psi_1\).\looseness=-1
\end{proposition}

\begin{proof}
  We can easily see that the greedy algorithm obeying the attack order \(0 \le \psi_0 \le \psi_1 \le \dots \le \psi_u \le \dots\) obtains the optimal solution by an exchange argument. Since \(g(\psi_i)\) is strictly increasing and does not change the order, we can simply omit it. Let us suppose we are given the optimal plan \(\rho^*\) and the greedy solution has the plan \(\rho\). Suppose \(\rho^*\) would contain one or more nodes for that \(w > u\) instead of \(q \le u\). We know that \(\psi_w \ge \psi_q\) and hence \(\Delta_w \ge \Delta_q\). Thus, replacing \(w\) by \(q\) would either lead to the an equally good or even better solution (contradiction!). Hence, the greedy plan \(\rho\) is at least as good as the optimal plan \(\rho^*\). The solution is unique except for ties s.t.\ \(0 \le \psi_i = \psi_j \le \psi_u\).
  
  Consequently, a surrogate loss \(\mathcal{L}'\) that leads to the order above will yield the global optimum as well. The order is preserved if (compare with \autoref{definition:budgetaware}):
  \begin{enumerate}[label=(\Roman*)]
      \item \(\nicefrac{\partial \mathcal{L}'}{\partial \evz_{c^*}} |_{\psi < 0} = 0\)
      \item \(\nicefrac{\partial \mathcal{L}'}{\partial \evz_{c^*}} |_{\psi_0} < \nicefrac{\partial \mathcal{L}'}{\partial \evz_{c^*}} |_{\psi_1}\) for any \(0 < \psi_0 < \psi_1\) (i.e.\ \(\nicefrac{\partial \mathcal{L}'}{\partial \evz_{c^*}}\) is strictly concave for positive inputs)
  \end{enumerate}
  From property (II) follows that \(\nicefrac{\partial \mathcal{L}'}{\partial \evz_{c^*}}\) is minimal for \(\psi \to 0^+\).
\end{proof}

\subsection{Alternative Version of \autoref{proposition:goodsurrogate}}\label{sec:appendix_alternativeprop1}

Here we state an alternative version of \autoref{proposition:goodsurrogate} if we relax Assumption 2 s.t.\ it only needs to hold in expectation.

\textbf{Assumption 2} The \emph{expected} budget required to change the prediction of node $i$ increases with the margin: $\mathbb{E}[\Delta_i | \psi_i] = g(|\psi_i|)$ for some increasing function $g(|\psi_i|) \ge 1.$

\textbf{Proposition 1}
Let $\mathcal{L}'$ be the surrogate for the 0/1 loss $\mathcal{L}_{0/1}$ used to attack a node classification algorithm $f_{\theta}(\mathbf{A}, \mathbf{X})$ with a global budget $\Delta$.
Additionally to Assumptions 1 and 2, suppose the adversary perturbs the chosen node until it is misclassified. We then obtain the global optimum of 
\[
\max_{\tilde{\mathbf{A}}\text{ s.t. }\|\tilde{\mathbf{A}} - \mathbf{A}\|_0 \le \Delta} \mathbb{E}[\mathcal{L}_{0/1}(f_{\theta}(\tilde{\mathbf{A}}, \mathbf{X}))]
\]
through greedily attacking the nodes in order $\frac{\partial \mathcal{L}'}{\partial \mathbf{z}_{c^*}}(\psi_0) \le \frac{\partial \mathcal{L}'}{\partial \mathbf{z}_{c^*}}(\psi_1) \le \dots \le \frac{\partial \mathcal{L}'}{\partial \mathbf{z}_{c^*}}(\psi_l)$ until the budget is exhausted $\Delta \le \sum_{i=0}^{l+1} \Delta_i$, if $\mathcal{L}'$ has the properties \textbf{(I)} $\frac{\partial \mathcal{L}'}{\partial \mathbf{z}_{c^*}} |_{\psi < 0} = 0$ and \textbf{(II)} $\frac{\partial \mathcal{L}'}{\partial \mathbf{z}_{c^*}} |_{\psi_0} < \frac{\partial \mathcal{L}'}{\partial \mathbf{z}_{c^*}} |_{\psi_1}$ for any $0 \le \psi_0 < \psi_1$.

Assumption 2 only needs to hold for a small fraction of nodes with low $\psi_i$. For the empirical distribution of a two-layer GCN on Cora ML, $\mathbb{E}[\Delta_i | \psi_i] = 1$ and $\text{Var}[\Delta_i | \psi_i] = 0$ for the $22.9\%$ nodes with lowest margin $\psi_i$. Hence, 
\[
\max_{\tilde{\mathbf{A}}\text{ s.t. }\|\tilde{\mathbf{A}} - \mathbf{A}\|_0 \le \Delta} \mathbb{E}[\mathcal{L}_{0/1}(f_{\theta}(\tilde{\mathbf{A}}, \mathbf{X}))] \approx \max_{\tilde{\mathbf{A}}\text{ s.t. }\|\tilde{\mathbf{A}} - \mathbf{A}\|_0 \le \Delta} \mathcal{L}_{0/1}(f_{\theta}(\tilde{\mathbf{A}}, \mathbf{X}))
\]
for small $\Delta$.

\section{Scalable Attacks}\label{sec:appendix_attacks}

We start with some general remarks on \(L_0\) Projected Gradient Descent in \autoref{sec:appendix_l0pgd}. Then we give more details on our attacks PR-BCD and GR-BCD (\autoref{sec:appendix_pgrbcd}). In \autoref{sec:appendix_localpprupdate}, we conclude this section with the derivation and complexity of the update of PPR scores (required for attacking PPRGo).

\subsection{\(L_0\) Projected Gradient Descent}\label{sec:appendix_l0pgd}

For \(L_0\) Projected Gradient Descent (\(L_0\)-PGD) we largely follow \citet{Xu2019a}. In fact, theirs is a special case of our PR-BCD (with the exceptions detailed below). To recover the (\(L_0\)-PGD), one solely needs to select all possible indices in line 3 in Algo.~\ref{algo:prbcd} and drop lines 10-14.

As discussed in \autoref{sec:attack}, we aim to solve:
\begin{equation}\label{eq:appendix_pgd}
  \max_{\mP\,\,\text{s.t.}\, \mP \in \{0, 1\}^{n \times n} \text{, } \sum \mP \le \Delta} \mathcal{L}(f_{\theta}(\adj \oplus \mP, \features))\,.
\end{equation}
where we explicitly model the perturbations \(\mP \in \{0, 1\}^{n \times n}\) ($\mP_{ij} = 1$ denotes an edge flip). For the sake of optimizing \(\mP\) with first-order/gradient methods we relax it from \(\{0, 1\}^{(n \times n)}\) to \([0, 1]^{(n \times n)}\). In words, during the attack we allow a \emph{weighted adjacency matrix} where the weights of \(\mP\) at the same time represent the probability to flip this edge in the last step of the attack. The sampling \(\mP \sim \text{Bernoulli}(\vp_{t})\) s.t.\ \(\sum \mP \le \Delta\) is required to obtain a binary perturbed adjacency matrix in the end: \(\tilde{\adj} \in \{0, 1\}^{(n \times n)}\). Note that we overload \(\oplus\) (besides its binary XOR meaning) s.t.\ \(\adj_{ij} \oplus p_{ij} = \adj_{ij} + p_{ij}\) if \(\adj_{ij} = 0\) and \(\adj_{ij} - p_{ij}\) otherwise.

\textbf{Projection.} Recall that after each gradient update, the projection \(\Pi_{\E[\text{Bernoulli}(\vp)] \le \Delta} (\vp)\) adjusts the probability mass such that \(\E[\text{Bernoulli}(\vp)] = \sum_{i \in b} \evp_i \le \Delta\) and that \(\vp \in [0, 1]\). Specifically, the projection operation
\begin{equation}\label{eq:appendix_project}
    \Pi_{\E[\text{Bernoulli}(\vp)] \le \Delta}(\vp) = \begin{cases}
      \Pi_{[0, 1]}(\vp) & \text{if } \mathbf{1}^\top \Pi_{[0, 1]}(\vp) \le \Delta \\
      \Pi_{[0, 1]}(\vp - \lambda \mathbf{1}) \text{ s.t. } \mathbf{1}^\top \Pi_{[0, 1]}(\vp - \lambda \mathbf{1}) = \Delta & \text{otherwise} \\
    \end{cases}
\end{equation}
where \(\Pi_{[0, 1]}(\vp)\) is simply clamping the values to the interval \([0, 1]\) and \(\lambda\) originates from the Lagrange formulation of the constrained optimization problem. \(\lambda\) can be efficiently calculated with the bisection method in \(\log_2[\nicefrac{\max(\vp) - \min(\vp - 1)}{\xi}]\) steps with the admissible error \(\xi\). In contrast to \citet{Xu2019a}, we additionally limit the number of steps to account for numerical instabilities on very large graphs.

\textbf{Sampling solution.} To retrieve a discrete and valid perturbed adjacency matrix in the last step, we sample \(\mP \sim \text{Bernoulli}(\vp_{t})\) s.t.\ \(\sum \mP \le \Delta\). \citet{Xu2019a} propose to sample for 20 times and reject all samples that violate the constraint. To eliminate the case that no solution was found and for improved attack strength (at the cost of a potential bias), we take the top-\(\Delta\) values of \(\vp\) instead of sampling in the first iteration of this ``rejection sampling'' procedure. In case of ties, we take the preceding sample.

\textbf{Learning rate.} To obtain a constant learning rate regardless of the budget, we scale a ``base'' learning rate (hyperparameter) by the budget. When using the PR-BCD attack (which we discuss next), we use the block size \(\nicefrac{b}{n^2}\) as an additional scaling factor and then apply the square root.

\subsection{Projected and Greedy Randomized Block Coordinate Descent}\label{sec:appendix_pgrbcd}

We first give some implementation details on our Projected Randomized Block Coordinate Descent (PR-BCD). Then we formally introduce Greedy Randomized Block Coordinate Descent (GR-BCD).

\textbf{Sampling w/o replacement.} As it turns out, even sampling w/o replacement \(\vi_0 \in \{0, 1, \dots, n^2 - 1\}^b\), which we need to determine the current block, is not easily parallelizable if one just has \(\mathcal{O}(b)\) memory and, hence, rather slow on modern GPUs. For this reason, we simply sample with replacement and afterward drop the duplicates. This comes at the cost of not having a block with exactly \(b\) elements. Especially on large graphs, the difference is rather small, although, collisions do exist. For a proper analysis, we refer to well-studied problems such as the Birthday Paradox or hash sets/tables.

\textbf{Representing zeros.} We require that all elements in \(\vp\) have a small, negligible non-zero value, i.e.\ must not be exactly zero. This affects the initialization and the projection procedure. We require it for two reasons: (1) we can easily ``detect'' edge removals (\(\evp_i\) must be subtracted instead of added) and (2) some sparse operations implicitly remove edges of zero weight.

\textbf{GR-BCD.} The biggest pitfall while aiming for maximum scalability is that we do not desire a runtime of \(\mathbf{O}(m)\). Instead, we solely want to iterate a constant number of steps (epochs) \(E\). We simply achieve this through defining a schedule \(\Delta_t\) for \(t \in \{1,2, \dots, E\}\) where \(\sum_{t=1}^E \Delta_t = \Delta\). In our experiments, we distribute the budget evenly and leave more complicated alternatives for future work. For the pseudo code of GR-BCD see \autoref{algo:appendix_grbcd}. 

\textbf{Advantages and limitations.} Our GR-BCD shares many commonalities with PR-BCD but does not require a learning rate \(\alpha_t\), heuristic \(h(\dots)\), and \(E_{\text{res.}}\) since we resample in each epoch. Another advantage is that we do not require \(b > \Delta\), which makes it more scalable than PR-BCD. However, since PR-BCD is scalable itself, in our experiments, we always kept the same block sizes for both attacks for improved comparability of results. GR-BCD's biggest drawback is its much slower learning dynamics. That is if an edge is flipped it is rarely flipped back. This is particularly important if one does not attack a GCN / designs an adaptive attack (see \autoref{sec:appendix_global}).

\begin{algorithm}[ht]
  \small 
  \caption{Greedy Randomized Block Coordinate Descent (GR-BCD)}
  \label{algo:appendix_grbcd}
  \begin{algorithmic}[1]
    \STATE {\bfseries Input:} Gr.\ \((\adj, \features)\), lab.\ \(\vy\), GNN \(f_{\theta}(\cdot)\), loss \(\mathcal{L}\)
    \STATE {\bfseries Parameter:} block size \(b\), schedule \(\Delta_t\) for \(t \in \{1,2, \dots, E\}\)
    \STATE Draw w/o replacement \(\vi_0 {\in} \{0, 1, \dots, n^2 - 1\}^b\)\hspace{-0.5em}
    \STATE Initialize zeros for \(\vp_0 \in \R^b\)
    \STATE initialize \(\hat{\adj} \leftarrow \adj\)
    \FOR{\(t \in \{1,2, \dots, E\}\)}
    \STATE \(\hat{\vy} \leftarrow f_{\theta}(\hat{\adj} \oplus \vp_{t-1}, \features)\)
    \STATE Flip \(\arg\mathrm{top}\text{-}\Delta_t(\nabla_{\vi_{t-1}}\mathcal{L}(\hat{\vy}, \vy))\) edges in \(\hat{\adj}\)
    \STATE \(\text{mask}_{\text{res.}} \leftarrow h(\vp_{t})\)
    \STATE \(\vp_t[\text{mask}_{\text{res.}}] \leftarrow \mathbf{0}\)
    \STATE Resample \(\vi_{t}[\text{mask}_{\text{res.}}]\)
    \ENDFOR
    \STATE Return \(\hat{\adj}\)
  \end{algorithmic}
\end{algorithm}

\subsection{Derivation and Complexity of Personalized Page Rank Update}\label{sec:appendix_localpprupdate}

In the following, we discuss how we can attack a single node on PPRGo using PR-BCD (i.e.\ a local attack). Since PPRGo avoids a recursive message passing scheme, relying on the PPR scores, we need an efficient, differentiable procedure to update the PPR scores given the perturbation of the adjacency matrix. We further limit the perturbations to the incoming edges. Perturbing adjacent edges is the most effective attack~\cite{Zugner2018}. To update the PPR scores of a directed graph for a node in \(\boldsymbol{\Pi}\), we use the Sherman-Morrison formula
\begin{equation}\label{eq:appendix_shermanmorrison}
    (\mB + \vu\vv^\top)^{-1} = \mB^{-1} - \frac{\mB^{-1}\vu\vv^\top\mB^{-1}}{1 + \vv^\top\mB^{-1}\vu}
\end{equation}
for rank one update \(\vu\vv^\top\) of the inverse of an invertible matrix \(\mB \in \sR^{n \times n}\). %
The rank one \(\vu\vv^\top\) update in general has shape \([n \times n]\) and therefore comes with space complexity \(\mathcal{O}(n^2)\) and the update via the Sherman-Morrison formula has \(\mathcal{O}(n^3)\). Since we use row normalization with PPRGo, we can attack the PPR scores via updating a single row \(\tilde{\boldsymbol{\Pi}}_i\) of the adjacency matrix \(\mA\) (including normalization) and obtain the gradient for the \(b\) potentially non-zero entries in \(\vp\).

We can write the closed-form local PPR update as:
\begin{equation}\label{eq:appendix_pprrowupdate}
    \tilde{\boldsymbol{\Pi}}_i
    = \alpha \Big[ \mI - (1-\alpha)\mD^{-1}\mA +  \vu\vv^\top \Big]_i^{-1} 
    = \alpha \left( \boldsymbol{\Pi}'_i - \frac{\boldsymbol{\Pi}'_{ii} \vv \boldsymbol{\Pi}'}{1 + \vv \boldsymbol{\Pi}'_{:i}} \right)
\end{equation}
where \(\boldsymbol{\Pi}' = (\mI - (1-\alpha)\mD^{-1}\mA)^{-1} = \alpha^{-1}\boldsymbol{\Pi}\) and we choose \(\evu_j=0\,\forall j \ne i\) and \(\evu_i=1\). For PPR we need e.g.\ a row stochastic matrix and hence need to normalize the adjacency matrix, also accounting for the prospective update. This implies that through an alteration of \(b\) entries in the \(i\)-th row of the \emph{unnormalized} adjacency matrix, we need to adjust every entry of this row to obtain the \emph{normalized} adjacency matrix. We can simply achieve this through adding the normalized row \((\mD_{ii} + \sum \vp)^{-1} (\mA_i + \vp)\) after the alteration and subtract its original entries \(\mD^{-1}_{ii} \mA_i\). Putting this together, the rank one update of the \(i\)-th row results in \(\vv = (\mD_{ii} + \sum \vp)^{-1} (\mA_i + \vp) - \mD^{-1}_{ii} \mA_i\) where \(\vp\) is a sparse vector with at most \(b\) non-zero elements.

\begin{figure}[H]
    \centering
    \resizebox{1\linewidth}{!}{\input{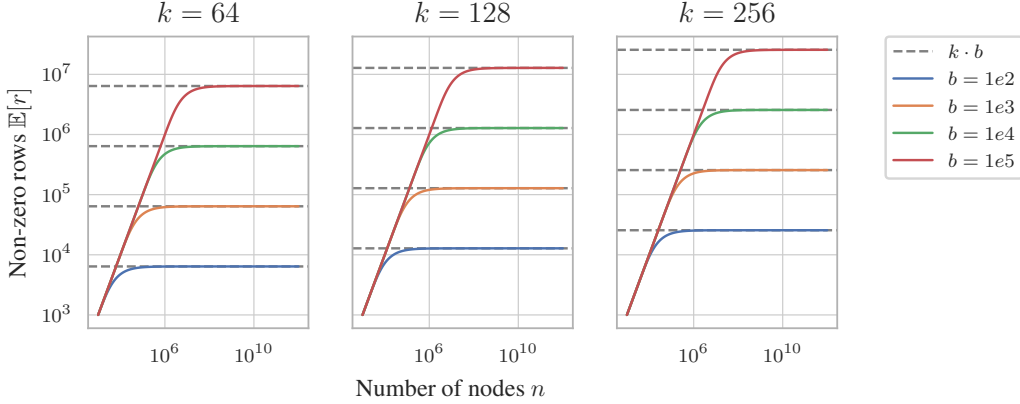}}
    \caption{\(\mathbb{E}[r] = \frac{n^b - (n - k)^b}{n^{b-1}}\) for different (practical) values of \(k\) and \(b\).}
    \label{fig:appendix_scalingppr}
\end{figure}

With dense matrices, this would leave us with a complexity of \(\mathcal{O}(b n)\) due to the vector-matrix product \(\vv \boldsymbol{\Pi}'\). We follow~\citet{Bojchevski2020a} and use the top-k-sparsified PPR \(\boldsymbol{\Pi}^{(k)}\) instead of \(\boldsymbol{\Pi}\) with at most \(k\) entries per row. Since \(\vv\) has at most \(b\) non-zero entries, most columns in the slice \(\boldsymbol{\Pi}^{(k)}_{\vv\ne0}\) only contain zero elements. Thus, we can equivalently write \(\vv \boldsymbol{\Pi}'\) as a dense vector matrix product of shapes \([1, b]\) and \([b, r]\), where r is the number of non-zero columns in the rows \(\boldsymbol{\Pi}'_{\vi_t}\). Recall that \(\vi_t\) are the indices of epoch \(t\) and that \(b \ge |\vi_t|\). If we assume randomly distributed ones, the probability of a non-zero entry is \(\nicefrac{k}{n}\). Hence we can model \(P(\sum \boldsymbol{\Pi}^{(k)}_{\vv\ne0, j}) = \text{Bin}(b, \nicefrac{k}{n})\) for column \(j\) and analogously
\begin{equation}
\begin{aligned}
      \mathbb{E}[r] 
      &= n \cdot P\left(\sum \boldsymbol{\Pi}^{(k)}_{\vv\ne0, j} > 0\right) \\
      &= n \left[ 1 - P\left(\sum \boldsymbol{\Pi}^{(k)}_{\vv\ne0, j} = 0\right) \right] \\
      &= n \left[1 - \left(1 - \frac{k}{n}\right)^b\right] \\
      &= \frac{n^b - (n - k)^b}{n^{b-1}}\,.
\end{aligned}
\end{equation}
For an appropriate choice of \(k \ll n\), the expected number of non-zero rows is \(\mathbb{E}[r] = \mathcal{O}(b k)\). The stronger asymptotic relation \(\mathbb{E}[r] = \Theta(b k)\) (and more strict alternatives) holds, but we omit this discussion for simplicity. Instead, we refer to \autoref{fig:appendix_scalingppr} for an illustration. In summary, the complexity of \(\vv \boldsymbol{\Pi}'\) and our local attack on PPRGo is \(\mathcal{O}(b k)\). Please note that in contrast to the global PR-BCD attack, this \emph{includes} the (Soft Median) PPRGo, and therefore is much more scalable. In practice, we observed slightly lower values for \(r\) than predicted by the relation above. We hypothesize that this is due to the fact that many rows contain less than \(k\) non-zero elements (depending on the approximation of the PPR scores) but that our assumption of randomly distributed non-zero elements holds.

To obtain a local attack we simply need to change lines 3 and 13 of Algo.~\ref{algo:prbcd} to sample only indices \(\vi_0 \in \{0, 1, \dots, n-1\}\). Further, we either keep line 6 if we attack e.g.\ GCN~\cite{Kipf2017} or update \(\tilde{\boldsymbol{\Pi}_i}\) as described in \autoref{eq:pprrowupdate}. We simply use a margin loss in logit space since we only have a local budget. After the attack and before applying the victim model the last time, we recalculate the PPR score for the target node based on the perturbed graph structure. The difference between the margin in the best epoch and after recalculating the PPR scores is usually negligible and shows that the approximation holds.\looseness=-1

\section{Scalable Defense}\label{sec:appendix_defense}

We first formally define the breakdown point. Then in \autoref{sec:appendix_prooftheo1}, we give the proof of \autoref{theorem:softmedianbreakdown} and in \autoref{sec:appendix_weightedbreakdown} we extend the discussion to the weighted Soft Median.

\textbf{Breakdown point.} Many metrics have been proposed that capture the robustness of a point estimate / aggregation with different flavors. One of the most widely used properties is the breakdown point. The (finite-sample) breakdown point captures the minimal fraction \(\epsilon = \nicefrac{m}{n}\) so that the result of the location estimator \(\mu(\features)\) can be arbitrarily placed~\citep{Donoho1983}:
\begin{equation}\label{eq:appendix_breakdown}
  \epsilon^*(t, \features) = \min_{1 \le m \le n} \left \{ \frac{m}{n}: \sup_{\pertm} \|\mu(\features)-\mu(\pertm)\| = \infty \right \}
\end{equation}
In this section we use \(m\) and \(n\) differently than in the rest of the paper: \(m\) denotes the number of perturbed inputs and \(n\) the number of inputs of the aggregation \(\mu(\features)\) (or number of rows in \(\features\)).

\subsection{Proof of \autoref{theorem:softmedianbreakdown}}\label{sec:appendix_prooftheo1}

\setcounter{theorem}{0}
\begin{theorem}
  Let \(\featset = \{ \mathbf{\mathbf{x}}_1, \dots, \mathbf{\mathbf{x}}_n\} \) be a collection of points in \(\mathbb{R}^d\) with finite coordinates and temperature \(T \in [0, \infty) \). Then the Soft Median location estimator (\autoref{eq:softmedian}) has the finite sample breakdown point of \(\epsilon^*(\mu_{\text{Soft Median}}, \features) = \nicefrac{1}{n} \lfloor \nicefrac{(n+1)}{2}\rfloor \) (asympt.\ \( \lim_{n \to \infty} \epsilon^*(\mu_{\text{SoftMedian}}, \features) = 0.5 \)).
\end{theorem}

\begin{proof}\label{proof:appendix_actual_soft_median}
  Let \( \pertmset \) be decomposable such that \(\pertmset = \pertmset^{(\text{c})} \cup \pertmset^{(\text{p})} \) and \(\pertmset^{(\text{c})} \cap \pertmset^{(\text{p})} = \emptyset\). Here \(\pertmset^{(\text{c})}\) denotes the clean and \(\pertmset^{(\text{p})}\) the perturbed inputs.  We now have to find the minimal fraction of outliers \(\epsilon\) for which \(\sup_{\pertm} \|\mu_{\text{SoftMedian}}(\pertm)\| < \infty\) does not hold anymore. According to \autoref{eq:appendix_breakdown}, if we now want to arbitrarily perturb the Soft Median, 
  we must \(\|\tilde{\vx}_{v}\| \to \infty ,\,\exists\, v \in \pertmset^{(\text{p})}\). Next we analyze the influence of this instance on \autoref{eq:softmedian} (w.l.o.g.\ we omit the factor \(\sqrt{D}\)):
  \[
    \begin{aligned}
      \hat{\softout}_{v} \tilde{\vx}_{v}
      &= \frac{\exp \left\{-\frac{1}{T} \|\bar{\vx} - \tilde{\vx}_{v}\| \right\} \tilde{\vx}_{v}}{\sum\nolimits_{i \in \pertmset^{(\text{c})}} \exp \left \{-\frac{1}{T} \|\bar{\vx} - \vx_{i}\| \right\} + \sum\nolimits_{j \in \pertmset^{(\text{p})}} \exp \left \{-\frac{1}{T} \|\bar{\vx} - \vx_{j}\| \right\}} \\
    \end{aligned}
  \]
  Instead of \(\lim_{\|\tilde{\vx}_{v}\| \to \infty} \hat{\softout}_{v} \vx_{v}\), we can equivalently derive the limit for the numerator and the denominator independently, as long as the denominator does not approach 0 which is easy to show (the denominator is \(> 0\) and \(\le |\pertmset|\)). With \(\lim_{\|\tilde{\vx}_{v}\| \to \infty}\) we denote the fact that the statements holds regardless how we achieve that the norm approaches infinity:
  \[
    \begin{aligned}
      \lim_{\|\tilde{\vx}_{v}\| \to \infty} \Big\| \exp \left\{-\frac{1}{T} \|\bar{\vx} - \tilde{\vx}_{v}\| \right\} \tilde{\vx}_{v} \Big\| = 
      \begin{cases}
        0 \text{, if } \lim_{\|\tilde{\vx}_{v}\| \to \infty} \|\bar{\vx} - \tilde{\vx}_{v}\| = 0\\
        \infty\text{, otherwise}
      \end{cases}
    \end{aligned}
  \]
  Please note that \(\lim_{x \to \infty} x e^{-x/a} = 0\) for \(a \in [0, \infty)\).
  As long as \(\epsilon < 0.5\), we know that for each dimension the perturbed dimension-wise median must be still within the range of the clean points. Or in other words, the perturbed median lays within the smallest possible hypercube around the original clean data \(\featset\). As long as \(\epsilon < 0.5\) we have that \(\lim_{\|\tilde{\vx}_{v}\| \to \infty} \|\bar{\vx} - \tilde{\vx}_{v}\| = 0\). Consequently, \(\|\mu(\features)-\mu(\pertm)\| = \infty\) can only be true if \(\nicefrac{m}{n} \ge 0.5\) for \(T \in [0, \infty)\).
\end{proof}

\subsection{Weighted Soft Median}\label{sec:appendix_weightedbreakdown}

It is easy to show that our proof also holds in the weighted case. Recall that we denote the weights of the neighbors with \(\mathbf{a}\). We extract from the weighted/normalized adjacency matrix (compare with \autoref{eq:appendix_messagepassing}). Given that a computer program represents numbers with limited precision, we do not provide an elaborate proof for weights in \(\R\). Instead, we can convert a weighted  problem into an unweighted, if we can find the greatest common divisor \(\text{gcd}(\mathbf{a}) = \text{gcd}([\begin{matrix}\mathbf{a}_1 & \dots & \mathbf{a}_n \end{matrix}])\). Once we found a \text{gcd}, we can use it to determine the factor of replications for each instance s.t.\ the relations to not change. For more details we refer to \citep{Geisler2020}.

\subsection{Empirical Error}\label{sec:appendix_empirical_error}

\begin{figure}[H]
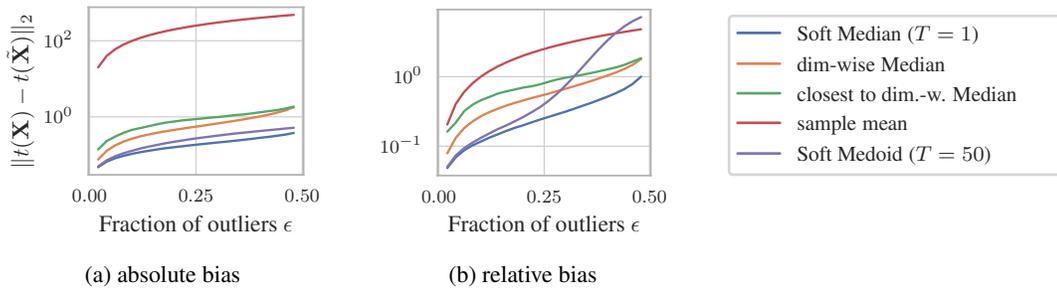

  \centering
  
  \(\begin{array}{ccc}
    \subfloat[absolute bias]{\resizebox{0.33\linewidth}{!}{\input{assets/emp_bias_curve_no_legend_1e3.pgf}}} & 
    \subfloat[relative bias]{\resizebox{0.305\linewidth}{!}{\input{assets/emp_bias_curve_no_leglab_1e1.pgf}}} &
    \raisebox{5ex}[0pt][0pt]{\begin{minipage}[t]{0.35\linewidth}
\begingroup%
\makeatletter%
\begin{pgfpicture}%
\pgfpathrectangle{\pgfpointorigin}{\pgfqpoint{1.942111in}{1.031412in}}%
\pgfusepath{use as bounding box, clip}%
\begin{pgfscope}%
\pgfsetbuttcap%
\pgfsetmiterjoin%
\definecolor{currentfill}{rgb}{1.000000,1.000000,1.000000}%
\pgfsetfillcolor{currentfill}%
\pgfsetlinewidth{0.000000pt}%
\definecolor{currentstroke}{rgb}{1.000000,1.000000,1.000000}%
\pgfsetstrokecolor{currentstroke}%
\pgfsetstrokeopacity{0.000000}%
\pgfsetdash{}{0pt}%
\pgfpathmoveto{\pgfqpoint{0.000000in}{0.000000in}}%
\pgfpathlineto{\pgfqpoint{1.942111in}{0.000000in}}%
\pgfpathlineto{\pgfqpoint{1.942111in}{1.031412in}}%
\pgfpathlineto{\pgfqpoint{0.000000in}{1.031412in}}%
\pgfpathclose%
\pgfusepath{fill}%
\end{pgfscope}%
\begin{pgfscope}%
\pgfsetbuttcap%
\pgfsetmiterjoin%
\definecolor{currentfill}{rgb}{1.000000,1.000000,1.000000}%
\pgfsetfillcolor{currentfill}%
\pgfsetfillopacity{0.800000}%
\pgfsetlinewidth{1.003750pt}%
\definecolor{currentstroke}{rgb}{0.800000,0.800000,0.800000}%
\pgfsetstrokecolor{currentstroke}%
\pgfsetstrokeopacity{0.800000}%
\pgfsetdash{}{0pt}%
\pgfpathmoveto{\pgfqpoint{0.122222in}{0.100000in}}%
\pgfpathlineto{\pgfqpoint{1.819888in}{0.100000in}}%
\pgfpathquadraticcurveto{\pgfqpoint{1.842111in}{0.100000in}}{\pgfqpoint{1.842111in}{0.122222in}}%
\pgfpathlineto{\pgfqpoint{1.842111in}{0.909189in}}%
\pgfpathquadraticcurveto{\pgfqpoint{1.842111in}{0.931412in}}{\pgfqpoint{1.819888in}{0.931412in}}%
\pgfpathlineto{\pgfqpoint{0.122222in}{0.931412in}}%
\pgfpathquadraticcurveto{\pgfqpoint{0.100000in}{0.931412in}}{\pgfqpoint{0.100000in}{0.909189in}}%
\pgfpathlineto{\pgfqpoint{0.100000in}{0.122222in}}%
\pgfpathquadraticcurveto{\pgfqpoint{0.100000in}{0.100000in}}{\pgfqpoint{0.122222in}{0.100000in}}%
\pgfpathclose%
\pgfusepath{stroke,fill}%
\end{pgfscope}%
\begin{pgfscope}%
\pgfsetroundcap%
\pgfsetroundjoin%
\pgfsetlinewidth{1.003750pt}%
\definecolor{currentstroke}{rgb}{0.298039,0.447059,0.690196}%
\pgfsetstrokecolor{currentstroke}%
\pgfsetdash{}{0pt}%
\pgfpathmoveto{\pgfqpoint{0.144444in}{0.842543in}}%
\pgfpathlineto{\pgfqpoint{0.366667in}{0.842543in}}%
\pgfusepath{stroke}%
\end{pgfscope}%
\begin{pgfscope}%
\definecolor{textcolor}{rgb}{0.150000,0.150000,0.150000}%
\pgfsetstrokecolor{textcolor}%
\pgfsetfillcolor{textcolor}%
\pgftext[x=0.455556in,y=0.803654in,left,base]{\color{textcolor}\rmfamily\fontsize{8.000000}{9.600000}\selectfont Soft Median (\(\displaystyle T=1\))}%
\end{pgfscope}%
\begin{pgfscope}%
\pgfsetroundcap%
\pgfsetroundjoin%
\pgfsetlinewidth{1.003750pt}%
\definecolor{currentstroke}{rgb}{0.866667,0.517647,0.321569}%
\pgfsetstrokecolor{currentstroke}%
\pgfsetdash{}{0pt}%
\pgfpathmoveto{\pgfqpoint{0.144444in}{0.681439in}}%
\pgfpathlineto{\pgfqpoint{0.366667in}{0.681439in}}%
\pgfusepath{stroke}%
\end{pgfscope}%
\begin{pgfscope}%
\definecolor{textcolor}{rgb}{0.150000,0.150000,0.150000}%
\pgfsetstrokecolor{textcolor}%
\pgfsetfillcolor{textcolor}%
\pgftext[x=0.455556in,y=0.642550in,left,base]{\color{textcolor}\rmfamily\fontsize{8.000000}{9.600000}\selectfont dim-wise Median}%
\end{pgfscope}%
\begin{pgfscope}%
\pgfsetroundcap%
\pgfsetroundjoin%
\pgfsetlinewidth{1.003750pt}%
\definecolor{currentstroke}{rgb}{0.333333,0.658824,0.407843}%
\pgfsetstrokecolor{currentstroke}%
\pgfsetdash{}{0pt}%
\pgfpathmoveto{\pgfqpoint{0.144444in}{0.526506in}}%
\pgfpathlineto{\pgfqpoint{0.366667in}{0.526506in}}%
\pgfusepath{stroke}%
\end{pgfscope}%
\begin{pgfscope}%
\definecolor{textcolor}{rgb}{0.150000,0.150000,0.150000}%
\pgfsetstrokecolor{textcolor}%
\pgfsetfillcolor{textcolor}%
\pgftext[x=0.455556in,y=0.487617in,left,base]{\color{textcolor}\rmfamily\fontsize{8.000000}{9.600000}\selectfont closest to dim.-w. Median}%
\end{pgfscope}%
\begin{pgfscope}%
\pgfsetroundcap%
\pgfsetroundjoin%
\pgfsetlinewidth{1.003750pt}%
\definecolor{currentstroke}{rgb}{0.768627,0.305882,0.321569}%
\pgfsetstrokecolor{currentstroke}%
\pgfsetdash{}{0pt}%
\pgfpathmoveto{\pgfqpoint{0.144444in}{0.371573in}}%
\pgfpathlineto{\pgfqpoint{0.366667in}{0.371573in}}%
\pgfusepath{stroke}%
\end{pgfscope}%
\begin{pgfscope}%
\definecolor{textcolor}{rgb}{0.150000,0.150000,0.150000}%
\pgfsetstrokecolor{textcolor}%
\pgfsetfillcolor{textcolor}%
\pgftext[x=0.455556in,y=0.332684in,left,base]{\color{textcolor}\rmfamily\fontsize{8.000000}{9.600000}\selectfont sample mean}%
\end{pgfscope}%
\begin{pgfscope}%
\pgfsetroundcap%
\pgfsetroundjoin%
\pgfsetlinewidth{1.003750pt}%
\definecolor{currentstroke}{rgb}{0.505882,0.447059,0.701961}%
\pgfsetstrokecolor{currentstroke}%
\pgfsetdash{}{0pt}%
\pgfpathmoveto{\pgfqpoint{0.144444in}{0.211104in}}%
\pgfpathlineto{\pgfqpoint{0.366667in}{0.211104in}}%
\pgfusepath{stroke}%
\end{pgfscope}%
\begin{pgfscope}%
\definecolor{textcolor}{rgb}{0.150000,0.150000,0.150000}%
\pgfsetstrokecolor{textcolor}%
\pgfsetfillcolor{textcolor}%
\pgftext[x=0.455556in,y=0.172215in,left,base]{\color{textcolor}\rmfamily\fontsize{8.000000}{9.600000}\selectfont Soft Medoid (\(\displaystyle T=50\))}%
\end{pgfscope}%
\end{pgfpicture}%
\makeatother%
\endgroup%
\end{minipage}}
    \\
  \end{array}\)
  \caption{Empirical error for a point mass perturbation. We reproduce Figure 2 in~\citep{Geisler2020} and add our Soft Median.\label{fig:appendix_empirical_error}}
\end{figure}

Similar to the finding in~\citep{Geisler2020}, we observe our Soft Median comes with a lower error if facing perturbed inputs (see \autoref{fig:appendix_empirical_error} which reproduce and complement Fig.\ 2 in~\citep{Geisler2020}). Here we plot the error $\|t(\mathbf{X}) - t(\tilde{\mathbf{X}})\|_2$, for 50 samples from a centered ($t_{\text{SoftMedian}}(\mathbf{X}) = 0$) bivariate normal distribution. The adversary is a point mass perturbation on the first axis over increasing fraction of outlier $\epsilon$.

\subsection{Improving Provable Robustness}\label{sec:appendix_proovable_robustness}

Similarly to the Soft Medoid in~\citep{Schuchardt2021, Geisler2020}, in \autoref{tab:appendix_certified_robustness}, we show that the Soft Median can improve the certified robustness. Here we apply randomized smoothing~\citep{Bojchevski2020} and obtain a significantly greater provable adversarial robustness. In the subsequent table, we show the "Accumulated certificates" obtained by randomized smoothing (same setup as in Table 2 in \citep{Geisler2020}). Even though our defense does not come with an adversarial robustness guarantee, we show that it can lead to increased \emph{provable} robustness.

\begin{table}[ht]
\centering
\caption{Certified robustness with randomized smoothing~\citep{Bojchevski2020} following the setup of \citep{Geisler2020}.\label{tab:appendix_certified_robustness}}
\begin{tabular}{llllll}
\toprule
         & \textbf{Accumulated certificate} & \textbf{Add \& Del.} & \textbf{Add.} & \textbf{Del.} & \textbf{Accuracy} \\
\midrule
\multirow{3}{*}{\textbf{Cora ML}}  & \underline{Soft Median GDC}                         & \textbf{5.7}             & \textbf{0.66}      & \textbf{4.9}       & \textbf{0.833}         \\
  & Vanilla GCN                                    & 1.84                & 0.21          & 4.41          & 0.823             \\
  & Soft Medoid GDC                                & 5.5                 & 0.64          & 4.78          & 0.814             \\
\multirow{3}{*}{\textbf{Citeseer}}  &  \underline{Soft Median GDC}                         & \textbf{4.43}            & \textbf{0.57}      & 4.31          & \textbf{0.728}         \\
 & Vanilla GCN                                    & 1.24                & 0.11          & 3.88          & 0.710             \\
 & Soft Medoid GDC                                & 3.64                & 0.49          & \textbf{4.33}      & 0.705     \\
\bottomrule
\end{tabular}
\end{table}

\section{Theoretical Complexities}\label{sec:appendix_theoretical_complexities}

In the following, we summarize the theoretical complexities approaches we use in our experiments. We assume that the number of features and hidden neurons is negligible in comparison to e.g.\ the number of nodes. $k$ denotes the GDC/PPRGo hyperparameter for top-$k$-sparsification of the PPR matrix, $n$ is the number of nodes, and $m$ is the number of edges. We try to keep the overview simple and e.g.\ only list the most important hyperparameters. If a model preprocesses the adjacency matrix, we report the time complexities for preprocessing and GNN separately. For the attacks, we report the \emph{additional} complexity (i.e. GNN excluded). We also use $b$ for the block size, $\Delta$ as the budget, and $E$ for the number of epochs. We chose $b=\mathcal{O}(m)$ for global attacks. In \autoref{tab:appenidx_complexities_models} and \autoref{tab:appenidx_complexities_global_attacks}, we list the theoretical complexities of the used models and global attacks respectively.

\begin{table}[ht]
\centering
\caption{Theoretical complexities of the models and defenses.\label{tab:appenidx_complexities_models}}
\begin{tabular}{lll}
\toprule
\textbf{Architecture}    & \textbf{Memory Complexity} & \textbf{Time Complexity (all nodes)}        \\
\midrule
\underline{Soft Median GDC}   & $\mathcal{O}(k \cdot n)$   & $\mathcal{O}(n) + \mathcal{O}(k \cdot n)$   \\
\underline{Soft Median PPRGo} & $\mathcal{O}(k)$           & $\mathcal{O}(n) + \mathcal{O}(k \cdot n)$   \\
Vanilla GCN              & $\mathcal{O}(m)$           & $\mathcal{O}(m)$                            \\
Vanilla GDC              & $\mathcal{O}(k \cdot n)$   & $\mathcal{O}(n) + \mathcal{O}(k \cdot n)$   \\
Vanilla PPRGo            & $\mathcal{O}(k)$           & $\mathcal{O}(n) + \mathcal{O}(k \cdot n)$   \\
Soft Medoid GDC          & $\mathcal{O}(k^2 \cdot n)$ & $\mathcal{O}(n) + \mathcal{O}(k^2 \cdot n)$ \\
SVD GCN                  & $\mathcal{O}(n^2)$         & $\mathcal{O}(m) + \mathcal{O}(n^2)$         \\
Jaccard GCN              & $\mathcal{O}(m)$           & $\mathcal{O}(n^2) + \mathcal{O}(m)$         \\
RGCN                     & $\mathcal{O}(m)$           & $\mathcal{O}(m)$ \\                         
\bottomrule
\end{tabular}
\end{table}

\begin{table}[ht]
\centering
\caption{Theoretical complexities of the global attacks.\label{tab:appenidx_complexities_global_attacks}}
\begin{tabular}{llll}
\toprule
\textbf{Global Attack} & \textbf{\begin{tabular}[c]{@{}l@{}}Memory\\ Complexity\end{tabular}} & \textbf{Time Complexity}         & \textbf{Details}         \\
\midrule
\underline{PR-BCD} & $\mathcal{O}(b)$                                            & $\mathcal{O}(E \cdot b \log(b))$ & $b \ge \Delta$           \\
\underline{GR-BCD} & $\mathcal{O}(b)$                                            & $\mathcal{O}(E \cdot b)$         & $b \ge \frac{\Delta}{E}$ \\
FGSM          & $\mathcal{O}(n^2)$                                          & $\mathcal{O}(\Delta \cdot n^2)$  &                          \\
PGD           & $\mathcal{O}(n^2)$                                          & $\mathcal{O}(E \cdot n^2)$       &                          \\
DICE          & $\mathcal{O}(\Delta)$                                       & $\mathcal{O}(\Delta)$            &                          \\ 
\bottomrule
\end{tabular}
\end{table}

For the local complexities in \autoref{tab:appenidx_complexities_local_attacks} we distinguish between the complexities \emph{with and without} GNN.

\begin{table}[ht]
\centering
\caption{Theoretical complexities of the local attacks.\label{tab:appenidx_complexities_local_attacks}}
\begin{tabular}{lcccc}
\toprule
\textbf{Local Attack}                  & \multicolumn{2}{c}{\textbf{Memory Complexity}}   & \multicolumn{2}{c}{\textbf{Time Complexity}}                        \\
\textbf{Complexity of GNN}             & excluded              & included                 & excluded                         & included                         \\
\midrule
\underline{PR-BCD}                     & $\mathcal{O}(b)$      & $\mathcal{O}(m)$         & $\mathcal{O}(E \cdot b \log(b))$ & $\mathcal{O}(E \cdot m)$         \\
\underline{PR-BCD} (with PPRGo) & -                     & $\mathcal{O}(b \cdot k)$ & -                                & $\mathcal{O}(E \cdot b \log(b))$ \\
Nettack                                & $\mathcal{O}(\Delta)$ & $\mathcal{O}(m)$         & $\mathcal{O}(\Delta \cdot n)$    & $\mathcal{O}(E \cdot m)$         \\
SGA (with SGC)                  & -                     & $\mathcal{O}(m)$         & -                                & $\mathcal{O}(E \cdot m)$         \\
DICE                                   & $\mathcal{O}(\Delta)$ & $\mathcal{O}(m)$         & $\mathcal{O}(\Delta)$            & $\mathcal{O}(E \cdot m)$         \\
DICE (with PPRGo)               & -                     & $\mathcal{O}(\Delta)$    & -                                & $\mathcal{O}(E \cdot\Delta)$ \\

\bottomrule
\end{tabular}
\end{table}

We now continue with a discussion of SGA's complexity. The space complexity of SGA is largely driven by two factors: (a) edge deletions are considered for the entire receptive field of the attacked node, and (b) the edge insertions are considered to the top \(\Delta\) nodes of the second most likely class.

To determine the top \(\Delta\) for (b) one needs to obtain the gradient w.r.t. to the edge connecting to the nodes of the second most likely class. For some graphs this requires to check \(\mathcal{O}(n)\) nodes. Even though this can be done iteratively or in batches.

It is even more challenging to obtain the gradient towards the edges for (a). This is simply due to the recursive nature of GNNs. For example, in a graph that approximately follows a power-law distribution, there will be some nodes with very high degree. Thus, for a moderate number of message passing steps and with high-degree nodes in the neighborhood, this subgraph might span a large fraction of nodes. Hence, in the worst case, the number of edges we need to consider scales with \(\mathcal{O}(m)\). Note this is the same limitation we describe in \autoref{sec:attack} when we say that ``we are limited by the scalability of the attacked GNN''. Hence, we also experience this if attacking a GCN locally on datasets larger than Products. This was our main motivation to also consider PPRGo.

\section{Empirical Evaluation}\label{sec:appendix_empirical}

In \autoref{sec:appendix_empiricalsetup} we start with a more thorough description of the experiment setup and in \autoref{sec:appendix_timememorycost} we give insights into the time and memory cost. We conclude with additional experiments using our global attack and local attack in \autoref{sec:appendix_global} and \autoref{sec:appendix_local}, respectively. While our local PR-BCD attack is adaptive since it does not rely on a surrogate model, our global attacks are not adaptive and transfer the attack from a Vanilla GCN (common practice presumably because many defenses/baselines are not differentiable, e.g.\ see \citep{Entezari2020, Geisler2020, Wu2019, Zhu2019}). Nevertheless, we also experiment with direct, adaptive attacks on our defense Soft Median GCN in \autoref{sec:appendix_global} or more specifically \autoref{tab:appendix_direct}.

\subsection{Setup}\label{sec:appendix_empiricalsetup}

\textbf{Datasets.} We use the common Cora ML~\citep{Bojchevski2018} and Citeseer~\citep{McCallum2000} for a comprehensive comparison of the state-of-the-art attacks and defenses (most baselines do not scale further). For large scale experiments, we use PubMed~\citep{Sen2008} as well as arXiv, Products and Papers 100M of the recent Open Graph Benchmark~\citep{Hu2020}. Since most approaches rely on graphs of similar size as PubMed for their ``large-scale'' experiments, we scale the global attack by more than 100 times (number of nodes), or by factor 15,000 if counting the possible adjacency matrix entries (see \autoref{tab:appendix_datasets}).
We scale our local attack to Papers 100M which has 111 million nodes, outscaling previous local attacks by a factor of 500. For a detailed overview see \autoref{tab:appendix_datasets}.

\begin{table}[hb]
\centering
\caption{Statistics of the used datasets (extension of \autoref{tab:datasets}). For the dense adjacency matrix we assume that each elements is represented by 4 bytes. In the sparse case we use two 8 byte integer pointers and a 4 bytes float value. For Cora ML, Citeseer and PubMed we extract the largest connected component.}
\label{tab:appendix_datasets}
\vskip 0.11in
\resizebox{\linewidth}{!}{
\begin{tabular}{llrrrrrrr}
\toprule
\textbf{Dataset}    &                                                                \textbf{License} & \textbf{\#Features $d$} & \textbf{\#Nodes $n$} & \textbf{\#Edges $e$} & \makecell{\textbf{\#Possible}\\\textbf{edges $n^2$}} & \makecell{\textbf{Average}\\\textbf{degree $\nicefrac{e}{n}$}} & \textbf{Size (dense)} & \textbf{Size (sparse)} \\
\midrule
\textbf{Cora ML}~\citep{Bojchevski2018}     &                                                                             N/A &                   2,879 &                2,810 &               15,962 &                                             7.896E+06 &                                                           5.68 &              31.58 MB &              319.24 kB \\
\textbf{Citeseer}~\citep{McCallum2000}    &                                                                             N/A &                   3,703 &                2,110 &                7,336 &                                             4.452E+06 &                                                           3.48 &              17.81 MB &              146.72 kB \\
\textbf{PubMed}~\citep{Sen2008}      &                                                                             N/A &                     500 &               19,717 &               88,648 &                                             3.888E+08 &                                                           4.50 &               1.56 GB &                1.77 MB \\
\textbf{arXiv}~\citep{Hu2020}       &                         \href{https://opendatacommons.org/licenses/by/}{ODC-BY} &                     128 &              169,343 &            1,166,243 &                                             2.868E+10 &                                                           6.89 &             114.71 GB &               23.32 MB \\
\textbf{Products}~\citep{Hu2020}    &  \href{https://s3.amazonaws.com/amazon-reviews-pds/license.txt}{Amazon} &                     100 &            2,449,029 &          123,718,280 &                                             5.998E+12 &                                                          50.52 &              23.99 TB &                2.47 GB \\
\textbf{Papers 100M}~\citep{Hu2020} &                         \href{https://opendatacommons.org/licenses/by/}{ODC-BY} &                     128 &          111,059,956 &        1,615,685,872 &                                             1.233E+16 &                                                          14.55 &              49.34 PB &               32.31 GB \\
\bottomrule
\end{tabular}
}
\end{table}

\textbf{Attacks.} We compare our global attacks PR-BCD and GR-BCD (\autoref{sec:attack}) with PGD~\citep{Xu2019a}, and greedy FGSM (similar to~\citet{Dai2018}) attacks. The greedy FGSM-like attack is the dense equivalent of our GR-BCD attack with the exception of flipping one edge at a time. Regardless of the scale of the datasets, we also compare to the global DICE~\citep{Waniek2018} attack. DICE is a greedy, randomized black-box attack that flips one randomly determined entry in the adjacency matrix at a time. An edge is deleted if both nodes share the same label and an edge is added if the labels of the nodes differ. We ensure that a single node does not become disconnected. Moreover, we use 60\% of the budget to add new edges and otherwise remove edges. We compare our local PR-BCD (\autoref{sec:attack}) with Nettack
~\cite{Zugner2018}. Nettack perturbs the adjacency matrix greedily exploiting the properties of the linearized GCN surrogate.

\textbf{Defenses.} We compare our Soft Median architectures with state-of-the-art defenses \citep{Entezari2020, Geisler2020, Wu2019, Zhu2019}. Following~\cite{Geisler2020}, we use the GDC/PPR preprocessing~\cite{Klicpera2019a} in combination with our Soft Median.
The SVD GCN~\citep{Entezari2020} uses a (dense) low-rank approximation (here rank 50) of the adjacency matrix to filter adversarial perturbations. RGCN~\citep{Zhu2019} models the neighborhood aggregation via Gaussian distribution to filter outliers, and Jaccard GCN~\citep{Wu2019} filters edges based on attribute dissimilarity (here threshold 0.01). For the Soft Medoid GDC, we use the temperature \(T=0.5\) as it is a good compromise between accuracy and robustness (except for arXiv where we choose \(T=5.0\)). For more details about the Soft Medoid GDC see \autoref{sec:defense}.

\textbf{Checkpointing.} Empirically, almost 30 GB are required to train a three-layer GCN on Products (our largest dataset for global attacks) using sparse matrices. However, obtaining the gradient, e.g.\ towards the perturbation vector/matrix, requires extra memory. We notice that most operations in modern GNNs only depend on the neighborhood size (i.e.\ a row in the adjacency matrix). As proposed by~\citet{Chen2016}, the gradient is obtainable with sublinear memory cost via checkpointing. The idea is to discard some intermediate results in the forward phase (that would be cached) and recompute them in the backward phase. Specifically, we chunk some operations (e.g. matrix multiplication) within the message passing step to successfully scale to larger graphs. This allows us to attack a three-layer GCN on Products with full GPU acceleration.

\textbf{Hyperparameters.} For full details we refer to our code and configuration. We run every experiment for three random seeds (unless otherwise stated) except for the largest dataset Papers100M.

\emph{Models:} We typically train for at most 3000 epochs with early stopping and patience of 300. We use a learning rate of 0.01 and a weight decay of 0.001 for all models (except PPRGo and SGC). On all datasets, we use a restart probability of \(0.15\) (except arXiv 0.1) for GDC and sparsify the adjacency by selecting the top 64 edges in each row. For the remaining configuration, we closely follow the setup of~\citet{Geisler2020} on Cora, Citeseer, and Pubmed. On arXiv as well as Products we follow~\citet{Hu2020} but still train for 3000 epochs with early-stopping patience of 300 and use three layers / message passing steps. We only deviate from the standard configuration for the PPRGo models as they are more sensitive to the hyperparameter choice. We use checkpointing for the Soft Medoid and Soft Median GDC on arXiv and for all models on Products. For the Soft Medoid and Soft Median GDC, we lower the number of layers and hidden dimensions such that they fit in the GPU memory. We determine the optimal temperature/parameters of our Soft Median GDC/PPRGo through a rudimentary grid search. On the small datasets, we typically end up with either \(T=0.2\) or \(T=0.5\). The larger the dataset, the larger the best temperature becomes (this effect is even stronger in combination with PPRGo). For the largest dataset Papers100M, we plot and analyze the influence of the temperature in \autoref{fig:appendix_paperstemperature}. 

\emph{Attacks:} For the global attacks GR-BCD and PR-BCD, we run the attack for 500 epochs (100 epochs fine-tuning with PR-BCD). We choose a block size \(b\) of 1,000,000, 2,500,000, 10,000,000 for Cora ML/Citeseer, Pubmed and arXiv/Products, respectively. For our local PR-BCD, we also attack for 500 epochs on Cora ML and Citeseer but observe that for Products and Papers100M 30 epochs are sufficient. Respectively, we choose a block size \(b\) of 10,000, 10,000, 20,000 and 2,500. We select the learning rates such that the budget requirement is met. 

\subsection{Time and Memory Cost}\label{sec:appendix_timememorycost}

\begin{table}[ht]
\centering
\caption{Time cost and memory cost of our \emph{global} PR-BCD attack on a Vanilla GCN.}
\label{tab:appendix_globaltimememory}
\vskip 0.11in
\begin{tabular}{llrr}
\toprule
\textbf{Dataset} & \textbf{Block size $b$} &  \textbf{Max GPU memory \textbackslash \hspace{0.025cm} GB} &  \textbf{Duration of epoch \textbackslash \hspace{0.025cm} s} \\
\midrule
Cora ML &               1,000,000 &                          0.34 &                            0.12 \\
PubMed &               2,500,000 &                          0.89 &                            0.32 \\
arXiv &              10,000,000 &                          4.29 &                            1.55 \\
\bottomrule
\end{tabular}
\end{table}

\begin{table}[ht]
\centering
\caption{Time cost and memory cost of our \emph{local} PR-BCD attack on PPRGo.}
\label{tab:appendix_localtimememory}
\vskip 0.11in
\begin{tabular}{lllrr}
\toprule
\textbf{Dataset} & \textbf{Architecture (victim)} & \textbf{Block size $b$} &  \makecell{\textbf{Max GPU}\\\textbf{memory \textbackslash \hspace{0.025cm} GB}} &  \makecell{\textbf{Duration of}\\\textbf{epoch \textbackslash \hspace{0.025cm} s}} \\
\midrule
\multirow{2}{*}{Cora ML} & Soft Median PPRGo &                  10,000 &                          0.08 &                            0.49 \\
                     & Vanilla PPRGo &                  10,000 &                          0.12 &                            0.52 \\
\multirow{2}{*}{Citeseer} & Soft Median PPRGo &                  10,000 &                          0.08 &                            0.39 \\
                     & Vanilla PPRGo &                  10,000 &                          0.14 &                            0.47 \\
\multirow{2}{*}{Products} & Soft Median PPRGo &                  20,000 &                          5.58 &                            4.29 \\
                     & Vanilla PPRGo &                  20,000 &                          6.07 &                            3.26 \\
\multirow{2}{*}{Papers 100M} & Soft Median PPRGo &                   2,500 &                          2.85 &                           14.66 \\
                     & Vanilla PPRGo &                   2,500 &                          2.81 &                           14.82 \\
\bottomrule
\end{tabular}
\end{table}

We present the time and memory cost for our global PR-BCD in \autoref{tab:appendix_globaltimememory} and for our local PR-BCD in \autoref{tab:appendix_localtimememory}\footnote{Note that the numbers presented here are slightly more pessimistic than in the main part since here we report the runtime/memory consumption including preprocessing and postprocessing steps. We will make this consistent in a future version of this paper.}. In all cases, our attack is reasonably fast and comes with a low memory footprint. We refer to \autoref{fig:memorycomparison} for a comparison with the dense PGD attack. Note that a global PGD attack on arXiv would require around 1 TB (see \autoref{fig:memorycomparison}) while our PR-BCD only requires around 4 GB while still being effective (see \autoref{tab:appendix_globaltimememory}). In \autoref{tab:appendix_localtimememory} we see that with a proper choice of the block size \(b\) once can attack a massive dataset on a reasonably sized GPU. Moreover, our Soft Medoid PPRGo and the Vanilla PPRGo come with a comparable runtime and memory consumption.

\subsection{Global Attacks}\label{sec:appendix_global}

\begin{figure}[ht]
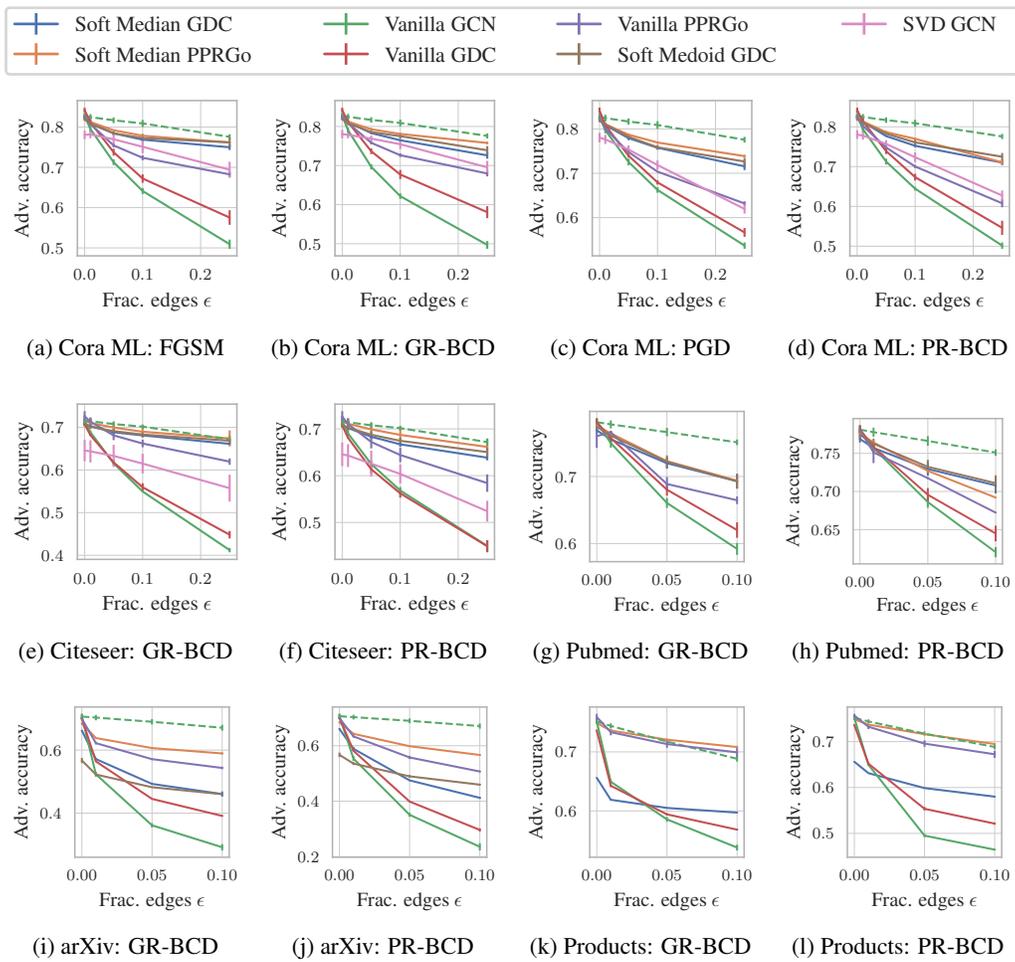

  \centering
  \hbox{\resizebox{\linewidth}{!}{
\begingroup%
\makeatletter%
\begin{pgfpicture}%
\pgfpathrectangle{\pgfpointorigin}{\pgfqpoint{5.418466in}{0.543199in}}%
\pgfusepath{use as bounding box, clip}%
\begin{pgfscope}%
\pgfsetbuttcap%
\pgfsetmiterjoin%
\definecolor{currentfill}{rgb}{1.000000,1.000000,1.000000}%
\pgfsetfillcolor{currentfill}%
\pgfsetlinewidth{0.000000pt}%
\definecolor{currentstroke}{rgb}{1.000000,1.000000,1.000000}%
\pgfsetstrokecolor{currentstroke}%
\pgfsetstrokeopacity{0.000000}%
\pgfsetdash{}{0pt}%
\pgfpathmoveto{\pgfqpoint{0.000000in}{0.000000in}}%
\pgfpathlineto{\pgfqpoint{5.418466in}{0.000000in}}%
\pgfpathlineto{\pgfqpoint{5.418466in}{0.543199in}}%
\pgfpathlineto{\pgfqpoint{0.000000in}{0.543199in}}%
\pgfpathclose%
\pgfusepath{fill}%
\end{pgfscope}%
\begin{pgfscope}%
\pgfsetbuttcap%
\pgfsetmiterjoin%
\definecolor{currentfill}{rgb}{1.000000,1.000000,1.000000}%
\pgfsetfillcolor{currentfill}%
\pgfsetfillopacity{0.800000}%
\pgfsetlinewidth{1.003750pt}%
\definecolor{currentstroke}{rgb}{0.800000,0.800000,0.800000}%
\pgfsetstrokecolor{currentstroke}%
\pgfsetstrokeopacity{0.800000}%
\pgfsetdash{}{0pt}%
\pgfpathmoveto{\pgfqpoint{0.122222in}{0.100000in}}%
\pgfpathlineto{\pgfqpoint{5.296243in}{0.100000in}}%
\pgfpathquadraticcurveto{\pgfqpoint{5.318466in}{0.100000in}}{\pgfqpoint{5.318466in}{0.122222in}}%
\pgfpathlineto{\pgfqpoint{5.318466in}{0.420977in}}%
\pgfpathquadraticcurveto{\pgfqpoint{5.318466in}{0.443199in}}{\pgfqpoint{5.296243in}{0.443199in}}%
\pgfpathlineto{\pgfqpoint{0.122222in}{0.443199in}}%
\pgfpathquadraticcurveto{\pgfqpoint{0.100000in}{0.443199in}}{\pgfqpoint{0.100000in}{0.420977in}}%
\pgfpathlineto{\pgfqpoint{0.100000in}{0.122222in}}%
\pgfpathquadraticcurveto{\pgfqpoint{0.100000in}{0.100000in}}{\pgfqpoint{0.122222in}{0.100000in}}%
\pgfpathclose%
\pgfusepath{stroke,fill}%
\end{pgfscope}%
\begin{pgfscope}%
\pgfsetbuttcap%
\pgfsetroundjoin%
\pgfsetlinewidth{1.003750pt}%
\definecolor{currentstroke}{rgb}{0.298039,0.447059,0.690196}%
\pgfsetstrokecolor{currentstroke}%
\pgfsetdash{}{0pt}%
\pgfpathmoveto{\pgfqpoint{0.255556in}{0.304311in}}%
\pgfpathlineto{\pgfqpoint{0.255556in}{0.415422in}}%
\pgfusepath{stroke}%
\end{pgfscope}%
\begin{pgfscope}%
\pgfsetroundcap%
\pgfsetroundjoin%
\pgfsetlinewidth{1.003750pt}%
\definecolor{currentstroke}{rgb}{0.298039,0.447059,0.690196}%
\pgfsetstrokecolor{currentstroke}%
\pgfsetdash{}{0pt}%
\pgfpathmoveto{\pgfqpoint{0.144444in}{0.359866in}}%
\pgfpathlineto{\pgfqpoint{0.366667in}{0.359866in}}%
\pgfusepath{stroke}%
\end{pgfscope}%
\begin{pgfscope}%
\definecolor{textcolor}{rgb}{0.150000,0.150000,0.150000}%
\pgfsetstrokecolor{textcolor}%
\pgfsetfillcolor{textcolor}%
\pgftext[x=0.455556in,y=0.320977in,left,base]{\color{textcolor}\rmfamily\fontsize{8.000000}{9.600000}\selectfont Soft Median GDC}%
\end{pgfscope}%
\begin{pgfscope}%
\pgfsetbuttcap%
\pgfsetroundjoin%
\pgfsetlinewidth{1.003750pt}%
\definecolor{currentstroke}{rgb}{0.866667,0.517647,0.321569}%
\pgfsetstrokecolor{currentstroke}%
\pgfsetdash{}{0pt}%
\pgfpathmoveto{\pgfqpoint{0.255556in}{0.149378in}}%
\pgfpathlineto{\pgfqpoint{0.255556in}{0.260489in}}%
\pgfusepath{stroke}%
\end{pgfscope}%
\begin{pgfscope}%
\pgfsetroundcap%
\pgfsetroundjoin%
\pgfsetlinewidth{1.003750pt}%
\definecolor{currentstroke}{rgb}{0.866667,0.517647,0.321569}%
\pgfsetstrokecolor{currentstroke}%
\pgfsetdash{}{0pt}%
\pgfpathmoveto{\pgfqpoint{0.144444in}{0.204933in}}%
\pgfpathlineto{\pgfqpoint{0.366667in}{0.204933in}}%
\pgfusepath{stroke}%
\end{pgfscope}%
\begin{pgfscope}%
\definecolor{textcolor}{rgb}{0.150000,0.150000,0.150000}%
\pgfsetstrokecolor{textcolor}%
\pgfsetfillcolor{textcolor}%
\pgftext[x=0.455556in,y=0.166044in,left,base]{\color{textcolor}\rmfamily\fontsize{8.000000}{9.600000}\selectfont Soft Median PPRGo}%
\end{pgfscope}%
\begin{pgfscope}%
\pgfsetbuttcap%
\pgfsetroundjoin%
\pgfsetlinewidth{1.003750pt}%
\definecolor{currentstroke}{rgb}{0.333333,0.658824,0.407843}%
\pgfsetstrokecolor{currentstroke}%
\pgfsetdash{}{0pt}%
\pgfpathmoveto{\pgfqpoint{1.853270in}{0.304311in}}%
\pgfpathlineto{\pgfqpoint{1.853270in}{0.415422in}}%
\pgfusepath{stroke}%
\end{pgfscope}%
\begin{pgfscope}%
\pgfsetroundcap%
\pgfsetroundjoin%
\pgfsetlinewidth{1.003750pt}%
\definecolor{currentstroke}{rgb}{0.333333,0.658824,0.407843}%
\pgfsetstrokecolor{currentstroke}%
\pgfsetdash{}{0pt}%
\pgfpathmoveto{\pgfqpoint{1.742159in}{0.359866in}}%
\pgfpathlineto{\pgfqpoint{1.964381in}{0.359866in}}%
\pgfusepath{stroke}%
\end{pgfscope}%
\begin{pgfscope}%
\definecolor{textcolor}{rgb}{0.150000,0.150000,0.150000}%
\pgfsetstrokecolor{textcolor}%
\pgfsetfillcolor{textcolor}%
\pgftext[x=2.053270in,y=0.320977in,left,base]{\color{textcolor}\rmfamily\fontsize{8.000000}{9.600000}\selectfont Vanilla GCN}%
\end{pgfscope}%
\begin{pgfscope}%
\pgfsetbuttcap%
\pgfsetroundjoin%
\pgfsetlinewidth{1.003750pt}%
\definecolor{currentstroke}{rgb}{0.768627,0.305882,0.321569}%
\pgfsetstrokecolor{currentstroke}%
\pgfsetdash{}{0pt}%
\pgfpathmoveto{\pgfqpoint{1.853270in}{0.149378in}}%
\pgfpathlineto{\pgfqpoint{1.853270in}{0.260489in}}%
\pgfusepath{stroke}%
\end{pgfscope}%
\begin{pgfscope}%
\pgfsetroundcap%
\pgfsetroundjoin%
\pgfsetlinewidth{1.003750pt}%
\definecolor{currentstroke}{rgb}{0.768627,0.305882,0.321569}%
\pgfsetstrokecolor{currentstroke}%
\pgfsetdash{}{0pt}%
\pgfpathmoveto{\pgfqpoint{1.742159in}{0.204933in}}%
\pgfpathlineto{\pgfqpoint{1.964381in}{0.204933in}}%
\pgfusepath{stroke}%
\end{pgfscope}%
\begin{pgfscope}%
\definecolor{textcolor}{rgb}{0.150000,0.150000,0.150000}%
\pgfsetstrokecolor{textcolor}%
\pgfsetfillcolor{textcolor}%
\pgftext[x=2.053270in,y=0.166044in,left,base]{\color{textcolor}\rmfamily\fontsize{8.000000}{9.600000}\selectfont Vanilla GDC}%
\end{pgfscope}%
\begin{pgfscope}%
\pgfsetbuttcap%
\pgfsetroundjoin%
\pgfsetlinewidth{1.003750pt}%
\definecolor{currentstroke}{rgb}{0.505882,0.447059,0.701961}%
\pgfsetstrokecolor{currentstroke}%
\pgfsetdash{}{0pt}%
\pgfpathmoveto{\pgfqpoint{3.054399in}{0.304311in}}%
\pgfpathlineto{\pgfqpoint{3.054399in}{0.415422in}}%
\pgfusepath{stroke}%
\end{pgfscope}%
\begin{pgfscope}%
\pgfsetroundcap%
\pgfsetroundjoin%
\pgfsetlinewidth{1.003750pt}%
\definecolor{currentstroke}{rgb}{0.505882,0.447059,0.701961}%
\pgfsetstrokecolor{currentstroke}%
\pgfsetdash{}{0pt}%
\pgfpathmoveto{\pgfqpoint{2.943288in}{0.359866in}}%
\pgfpathlineto{\pgfqpoint{3.165511in}{0.359866in}}%
\pgfusepath{stroke}%
\end{pgfscope}%
\begin{pgfscope}%
\definecolor{textcolor}{rgb}{0.150000,0.150000,0.150000}%
\pgfsetstrokecolor{textcolor}%
\pgfsetfillcolor{textcolor}%
\pgftext[x=3.254399in,y=0.320977in,left,base]{\color{textcolor}\rmfamily\fontsize{8.000000}{9.600000}\selectfont Vanilla PPRGo}%
\end{pgfscope}%
\begin{pgfscope}%
\pgfsetbuttcap%
\pgfsetroundjoin%
\pgfsetlinewidth{1.003750pt}%
\definecolor{currentstroke}{rgb}{0.576471,0.470588,0.376471}%
\pgfsetstrokecolor{currentstroke}%
\pgfsetdash{}{0pt}%
\pgfpathmoveto{\pgfqpoint{3.054399in}{0.149378in}}%
\pgfpathlineto{\pgfqpoint{3.054399in}{0.260489in}}%
\pgfusepath{stroke}%
\end{pgfscope}%
\begin{pgfscope}%
\pgfsetroundcap%
\pgfsetroundjoin%
\pgfsetlinewidth{1.003750pt}%
\definecolor{currentstroke}{rgb}{0.576471,0.470588,0.376471}%
\pgfsetstrokecolor{currentstroke}%
\pgfsetdash{}{0pt}%
\pgfpathmoveto{\pgfqpoint{2.943288in}{0.204933in}}%
\pgfpathlineto{\pgfqpoint{3.165511in}{0.204933in}}%
\pgfusepath{stroke}%
\end{pgfscope}%
\begin{pgfscope}%
\definecolor{textcolor}{rgb}{0.150000,0.150000,0.150000}%
\pgfsetstrokecolor{textcolor}%
\pgfsetfillcolor{textcolor}%
\pgftext[x=3.254399in,y=0.166044in,left,base]{\color{textcolor}\rmfamily\fontsize{8.000000}{9.600000}\selectfont Soft Medoid GDC}%
\end{pgfscope}%
\begin{pgfscope}%
\pgfsetbuttcap%
\pgfsetroundjoin%
\pgfsetlinewidth{1.003750pt}%
\definecolor{currentstroke}{rgb}{0.854902,0.545098,0.764706}%
\pgfsetstrokecolor{currentstroke}%
\pgfsetdash{}{0pt}%
\pgfpathmoveto{\pgfqpoint{4.524367in}{0.304311in}}%
\pgfpathlineto{\pgfqpoint{4.524367in}{0.415422in}}%
\pgfusepath{stroke}%
\end{pgfscope}%
\begin{pgfscope}%
\pgfsetroundcap%
\pgfsetroundjoin%
\pgfsetlinewidth{1.003750pt}%
\definecolor{currentstroke}{rgb}{0.854902,0.545098,0.764706}%
\pgfsetstrokecolor{currentstroke}%
\pgfsetdash{}{0pt}%
\pgfpathmoveto{\pgfqpoint{4.413256in}{0.359866in}}%
\pgfpathlineto{\pgfqpoint{4.635479in}{0.359866in}}%
\pgfusepath{stroke}%
\end{pgfscope}%
\begin{pgfscope}%
\definecolor{textcolor}{rgb}{0.150000,0.150000,0.150000}%
\pgfsetstrokecolor{textcolor}%
\pgfsetfillcolor{textcolor}%
\pgftext[x=4.724367in,y=0.320977in,left,base]{\color{textcolor}\rmfamily\fontsize{8.000000}{9.600000}\selectfont SVD GCN}%
\end{pgfscope}%
\end{pgfpicture}%
\makeatother%
\endgroup%
}}
  \vspace{-14pt}
  \makebox[\linewidth][c]{
    \(\arraycolsep=1pt\def\arraystretch{2}\begin{array}{cccc}
      \subfloat[Cora ML: FGSM]{\resizebox{0.24\linewidth}{!}{\input{assets/global_transfer_FGSM_cora_ml_pertaccuracy_no_legend.pgf}}} &
      \subfloat[Cora ML: GR-BCD ]{\resizebox{0.24\linewidth}{!}{\input{assets/global_transfer_GreedyRBCD_cora_ml_pertaccuracy_no_legend.pgf}}} &
      \subfloat[Cora ML: PGD]{\resizebox{0.24\linewidth}{!}{\input{assets/global_transfer_PGD_cora_ml_pertaccuracy_no_legend.pgf}}} &
      \subfloat[Cora ML: PR-BCD]{\resizebox{0.24\linewidth}{!}{\input{assets/global_transfer_PRBCD_cora_ml_pertaccuracy_no_legend.pgf}}} \\
      \subfloat[Citeseer: GR-BCD]{\resizebox{0.24\linewidth}{!}{\input{assets/global_transfer_GreedyRBCD_citeseer_pertaccuracy_no_legend.pgf}}} &
      \subfloat[Citeseer: PR-BCD]{\resizebox{0.24\linewidth}{!}{\input{assets/global_transfer_PRBCD_citeseer_pertaccuracy_no_legend.pgf}}} &
      \subfloat[Pubmed: GR-BCD]{\resizebox{0.24\linewidth}{!}{\input{assets/global_transfer_GreedyRBCD_pubmed_pertaccuracy_no_legend.pgf}}} &
      \subfloat[Pubmed: PR-BCD]{\resizebox{0.24\linewidth}{!}{\input{assets/global_transfer_PRBCD_pubmed_pertaccuracy_no_legend.pgf}}} \\
      \subfloat[arXiv: GR-BCD]{\resizebox{0.24\linewidth}{!}{\input{assets/global_transfer_GreedyRBCD_ogbn-arxiv_pertaccuracy_no_legend.pgf}}} &
      \subfloat[arXiv: PR-BCD]{\resizebox{0.24\linewidth}{!}{\input{assets/global_transfer_PRBCD_ogbn-arxiv_pertaccuracy_no_legend.pgf}}} &
      \subfloat[Products: GR-BCD]{\resizebox{0.24\linewidth}{!}{\input{assets/global_transfer_GreedyRBCD_ogbn-products_pertaccuracy_no_legend.pgf}}} &
      \subfloat[Products: PR-BCD]{\resizebox{0.24\linewidth}{!}{\input{assets/global_transfer_PRBCD_ogbn-products_pertaccuracy_no_legend.pgf}}} \\
    \end{array}\)
  }
  \caption{Adversarial accuracy for the budget \(\epsilon\) as a fraction of edges. The lower the adversarial accuracy, the stronger the attack or weaker the defense. The higher the adversarial accuracy, the stronger the defense or weaker the attack. Here we provide more detail than in \autoref{fig:prbcdlarge}.\label{fig:appendix_globaltransfer}}
\end{figure}

We compare the robustness of the models over different attack budgets in \autoref{fig:appendix_globaltransfer} (similarly to \autoref{fig:prbcdlarge}). In \autoref{tab:appendix_global_transfer}, we give a detailed overview of how each model's accuracy declines for all benchmarked attacks. More importantly, in \autoref{tab:appendix_direct}, we provide the results for an adaptive attack.

\begin{table}[hb]
\centering
\caption{Adversarial accuracy using \emph{direct, adaptive} attacks and our Soft Median GDC with the vanilla baselines. We show the adversarial accuracy and the clean test accuracy (last column). We highlight the \textbf{strongest defense} in bold as the attacks perform similarly. \underline{Our approaches} are underlined.}
\label{tab:appendix_direct}
\vskip 0.11in
\resizebox{\linewidth}{!}{
\begin{tabular}{llccccccccccccc}
\toprule
\textbf{Dataset}                       & \textbf{Attack} & \multicolumn{6}{c}{\underline{\textbf{GR-BCD}}} \\
                                  & Frac. edges \(\boldsymbol{\epsilon}\) &                        0.01 &                        0.05 &                        0.10 &                        0.25 &                        0.50 &                        1.00 \\
\midrule
\multirow{3}{*}{\textbf{Cora ML}} & \underline{Soft Median GDC} &           0.807 $\pm$ 0.002 &  \textbf{0.773 $\pm$ 0.002} &  \textbf{0.749 $\pm$ 0.001} &  \textbf{0.692 $\pm$ 0.004} &  \textbf{0.648 $\pm$ 0.002} &  \textbf{0.603 $\pm$ 0.001} \\
                  & Vanilla GCN &           0.789 $\pm$ 0.003 &           0.699 $\pm$ 0.003 &           0.619 $\pm$ 0.004 &           0.475 $\pm$ 0.004 &           0.333 $\pm$ 0.003 &           0.148 $\pm$ 0.005 \\
                  & Vanilla GDC &  \textbf{0.808 $\pm$ 0.002} &           0.749 $\pm$ 0.003 &           0.703 $\pm$ 0.003 &           0.623 $\pm$ 0.005 &           0.513 $\pm$ 0.005 &           0.396 $\pm$ 0.007 \\
\cline{1-8}
\multirow{3}{*}{\textbf{Citeseer}} & \underline{Soft Median GDC} &  \textbf{0.705 $\pm$ 0.001} &  \textbf{0.687 $\pm$ 0.002} &  \textbf{0.664 $\pm$ 0.001} &  \textbf{0.626 $\pm$ 0.003} &  \textbf{0.582 $\pm$ 0.001} &  \textbf{0.536 $\pm$ 0.004} \\
                  & Vanilla GCN &           0.689 $\pm$ 0.002 &           0.618 $\pm$ 0.001 &           0.554 $\pm$ 0.001 &           0.410 $\pm$ 0.003 &           0.265 $\pm$ 0.002 &           0.105 $\pm$ 0.008 \\
                  & Vanilla GDC &           0.679 $\pm$ 0.001 &           0.626 $\pm$ 0.002 &           0.588 $\pm$ 0.006 &           0.504 $\pm$ 0.003 &           0.421 $\pm$ 0.003 &           0.309 $\pm$ 0.007 \\
\bottomrule
\toprule
                                  & \textbf{Attack} & \multicolumn{6}{c}{\underline{\textbf{PR-BCD}}} \\
                                  & Frac. edges \(\boldsymbol{\epsilon}\) &                        0.01 &                        0.05 &                        0.10 &                        0.25 &                        0.50 &                        1.00 \\
\midrule
\multirow{3}{*}{\textbf{Cora ML}} & \underline{Soft Median GDC} &           0.796 $\pm$ 0.002 &  \textbf{0.735 $\pm$ 0.002} &  \textbf{0.690 $\pm$ 0.002} &  \textbf{0.615 $\pm$ 0.003} &  \textbf{0.564 $\pm$ 0.005} &  \textbf{0.523 $\pm$ 0.005} \\
                  & Vanilla GCN &           0.792 $\pm$ 0.003 &           0.704 $\pm$ 0.004 &           0.635 $\pm$ 0.005 &           0.478 $\pm$ 0.003 &           0.309 $\pm$ 0.005 &           0.141 $\pm$ 0.005 \\
                  & Vanilla GDC &  \textbf{0.799 $\pm$ 0.002} &           0.711 $\pm$ 0.003 &           0.645 $\pm$ 0.005 &           0.532 $\pm$ 0.006 &           0.457 $\pm$ 0.005 &           0.400 $\pm$ 0.006 \\
\cline{1-8}
\multirow{3}{*}{\textbf{Citeseer}} & \underline{Soft Median GDC} &  \textbf{0.692 $\pm$ 0.002} &  \textbf{0.650 $\pm$ 0.002} &  \textbf{0.615 $\pm$ 0.003} &  \textbf{0.548 $\pm$ 0.005} &  \textbf{0.494 $\pm$ 0.006} &  \textbf{0.446 $\pm$ 0.008} \\
                  & Vanilla GCN &           0.689 $\pm$ 0.002 &           0.621 $\pm$ 0.003 &           0.560 $\pm$ 0.004 &           0.429 $\pm$ 0.007 &           0.282 $\pm$ 0.012 &           0.127 $\pm$ 0.009 \\
                  & Vanilla GDC &           0.670 $\pm$ 0.001 &           0.591 $\pm$ 0.004 &           0.515 $\pm$ 0.002 &           0.374 $\pm$ 0.003 &           0.264 $\pm$ 0.007 &           0.194 $\pm$ 0.006 \\
\bottomrule
\end{tabular}
}

\end{table}

\textbf{Adaptive, global attack.} In \autoref{tab:appendix_global_transfer}, we compare our Soft Median GDC to a Vanilla GCN (and Vanilla GDC as ablation). We make two major observations: (1) our Soft Median GDC outperforms the baselines by a large margin over all budgets \(\epsilon > 0.01\); (2) while in the previous experiments the greedy attacks (FGSM or GR-BCD) seemed to perform on par (sometimes even stronger) with PGD as well as PR-BCD, this does not hold if we attack Vanilla GDC or our Soft Median GDC. This implies that for a Vanilla GCN the gradient on the clean edges is an excellent indicator for the effectiveness of an edge flip. Moreover and with no surprise, transfer attacks can provide a false impression of robustness. We want to emphasize that for GNNs most defenses were not studied using an adaptive attack (e.g.\ see \citep{Entezari2020, Geisler2020, Wu2019, Zhu2019}). We now compare the performance with an MLP that achieves around 60\% on Cora ML~\citep{shchur_pitfalls_2019}. With GR-BCD we hit 60\% adversarial accuracy for \(\epsilon=1\) but for the PR-BCD we already drop below it somewhere in \([0.25, 0.5]\). Note that the Vanilla GCN already drops below the MLP performance \(\epsilon \approx 0.1\). For large budgets (e.g.\ \(\epsilon = 1.0\)), our Soft Median GDC has a four to five times higher adversarial accuracy than a Vanilla GCN.

Unfortunately, we are not aware of an efficient PPR implementation (large fraction of nodes at once) that allows us to backpropagate through it. Moreover, our Soft Median would lose a lot of its robustness, if we removed the PPR diffusion (GDC). Hence, we are limited to a calculation of the PPR scores with the matrix inverse. This is not scalable (runtime complexity \(\mathbf{O}(n^3)\)) and the inverse of a sparse matrix is not sparse in general (space complexity \(\mathbf{O}(n^2)\)). For this reason, we can only use an adaptive, global attack on Soft Median GDC on sufficiently small datasets. For adaptive attacks at scale, we refer to the local attacks in \autoref{sec:empirical} and \autoref{sec:appendix_local}.

\begin{sidewaystable}
\centering
\caption{Comparing attacks (transfer from Vanilla GCN) and defenses. We show the adversarial accuracy and the clean test accuracy (last column).
We highlight the \textbf{strongest defense} in bold as the attacks perform similarly. A more \emph{nuanced highlighting} is given for the strongest attack for each architecture and budget. \underline{Our approaches} are underlined.}
\label{tab:appendix_global_transfer}
\vskip 0.11in
\resizebox{.975\linewidth}{!}{
\begin{tabular}{llcccccccccccccccc}
\toprule
                                  & \textbf{Attack} & \multicolumn{2}{c}{\textbf{DICE}} & \multicolumn{2}{c}{\textbf{FGSM}} & \multicolumn{2}{c}{\underline{\textbf{GR-BCD}}} & \multicolumn{2}{c}{\textbf{PGD}} & \multicolumn{2}{c}{\underline{\textbf{PR-BCD}}} &               \textbf{Acc.} \\
                                  & Frac. edges \(\boldsymbol{\epsilon}\) &                        0.05 &                         0.1 &                                 0.05 &                                  0.1 &                                 0.05 &                         0.1 &                                 0.05 &                                  0.1 &                                 0.05 & 0.1 & \\
\midrule
\multirow{10}{*}{\rotatebox{90}{\textbf{Cora ML}}} & \underline{Soft Median GDC} &           0.816 $\pm$ 0.002 &           0.813 $\pm$ 0.002 &                    0.784 $\pm$ 0.001 &                    0.769 $\pm$ 0.002 &                    0.783 $\pm$ 0.001 &           0.765 $\pm$ 0.001 &                    0.779 $\pm$ 0.002 &                    0.758 $\pm$ 0.002 &           \textit{0.777 $\pm$ 0.000} &           \textit{0.752 $\pm$ 0.002} &           0.824 $\pm$ 0.002 \\
                                  & \underline{Soft Median PPRGo} &           0.814 $\pm$ 0.002 &           0.804 $\pm$ 0.001 &           \textbf{0.792 $\pm$ 0.000} &           \textbf{0.778 $\pm$ 0.001} &           \textbf{0.793 $\pm$ 0.002} &  \textbf{0.781 $\pm$ 0.002} &  \textbf{\textit{0.787 $\pm$ 0.001}} &  \textbf{\textit{0.769 $\pm$ 0.001}} &           \textbf{0.787 $\pm$ 0.002} &           \textbf{0.770 $\pm$ 0.001} &           0.821 $\pm$ 0.001 \\
                                  & Vanilla GCN &           0.817 $\pm$ 0.003 &           0.809 $\pm$ 0.004 &                    0.713 $\pm$ 0.003 &                    0.641 $\pm$ 0.003 &           \textit{0.697 $\pm$ 0.003} &  \textit{0.622 $\pm$ 0.003} &                    0.726 $\pm$ 0.004 &                    0.662 $\pm$ 0.003 &                    0.713 $\pm$ 0.003 &                    0.645 $\pm$ 0.002 &           0.827 $\pm$ 0.003 \\
                                  & Vanilla GDC &  \textbf{0.830 $\pm$ 0.003} &  \textbf{0.819 $\pm$ 0.002} &                    0.738 $\pm$ 0.004 &           \textit{0.672 $\pm$ 0.005} &           \textit{0.737 $\pm$ 0.003} &           0.677 $\pm$ 0.005 &                    0.740 $\pm$ 0.002 &                    0.679 $\pm$ 0.002 &                    0.739 $\pm$ 0.003 &                    0.674 $\pm$ 0.004 &  \textbf{0.842 $\pm$ 0.003} \\
                                  & Vanilla PPRGo &           0.816 $\pm$ 0.002 &           0.807 $\pm$ 0.001 &                    0.754 $\pm$ 0.002 &                    0.724 $\pm$ 0.003 &                    0.758 $\pm$ 0.002 &           0.726 $\pm$ 0.002 &           \textit{0.748 $\pm$ 0.002} &                    0.704 $\pm$ 0.001 &                    0.748 $\pm$ 0.003 &           \textit{0.700 $\pm$ 0.002} &           0.826 $\pm$ 0.002 \\
                                  & Vanilla GAT &           0.763 $\pm$ 0.002 &           0.725 $\pm$ 0.003 &                    0.741 $\pm$ 0.001 &                    0.688 $\pm$ 0.002 &                    0.743 $\pm$ 0.000 &           0.699 $\pm$ 0.001 &           \textit{0.731 $\pm$ 0.002} &                    0.683 $\pm$ 0.003 &                    0.738 $\pm$ 0.001 &           \textit{0.677 $\pm$ 0.002} &           0.806 $\pm$ 0.001 \\
                                  & Soft Medoid GDC &           0.814 $\pm$ 0.002 &           0.809 $\pm$ 0.002 &                    0.784 $\pm$ 0.003 &                    0.773 $\pm$ 0.005 &                    0.786 $\pm$ 0.002 &           0.775 $\pm$ 0.003 &           \textit{0.782 $\pm$ 0.003} &           \textit{0.759 $\pm$ 0.003} &                    0.783 $\pm$ 0.001 &                    0.761 $\pm$ 0.003 &           0.819 $\pm$ 0.002 \\
                                  & SVD GCN &           0.766 $\pm$ 0.005 &           0.752 $\pm$ 0.003 &                    0.770 $\pm$ 0.006 &                    0.751 $\pm$ 0.007 &                    0.769 $\pm$ 0.004 &           0.755 $\pm$ 0.006 &           \textit{0.753 $\pm$ 0.004} &           \textit{0.719 $\pm$ 0.005} &                    0.757 $\pm$ 0.004 &                    0.724 $\pm$ 0.006 &           0.781 $\pm$ 0.005 \\
                                  & Jaccard GCN &           0.809 $\pm$ 0.003 &           0.803 $\pm$ 0.003 &                    0.722 $\pm$ 0.002 &           \textit{0.661 $\pm$ 0.002} &           \textit{0.719 $\pm$ 0.001} &           0.664 $\pm$ 0.001 &                    0.730 $\pm$ 0.003 &                    0.673 $\pm$ 0.002 &                    0.725 $\pm$ 0.001 &                    0.667 $\pm$ 0.003 &           0.818 $\pm$ 0.003 \\
                                  & RGCN &           0.808 $\pm$ 0.002 &           0.796 $\pm$ 0.003 &           \textit{0.719 $\pm$ 0.004} &           \textit{0.654 $\pm$ 0.007} &                    0.725 $\pm$ 0.002 &           0.665 $\pm$ 0.005 &                    0.725 $\pm$ 0.005 &                    0.671 $\pm$ 0.007 &                    0.724 $\pm$ 0.003 &                    0.664 $\pm$ 0.004 &           0.819 $\pm$ 0.002 \\
\cline{1-13}
\multirow{10}{*}{\rotatebox{90}{\textbf{Citeseer}}} & \underline{Soft Median GDC} &           0.706 $\pm$ 0.001 &           0.699 $\pm$ 0.001 &                    0.695 $\pm$ 0.002 &                    0.676 $\pm$ 0.002 &                    0.688 $\pm$ 0.002 &           0.681 $\pm$ 0.002 &                    0.686 $\pm$ 0.002 &                    0.675 $\pm$ 0.002 &           \textit{0.683 $\pm$ 0.002} &           \textit{0.667 $\pm$ 0.003} &           0.708 $\pm$ 0.002 \\
                                  & \underline{Soft Median PPRGo} &           0.709 $\pm$ 0.006 &           0.700 $\pm$ 0.006 &  \textbf{\textit{0.697 $\pm$ 0.006}} &  \textbf{\textit{0.685 $\pm$ 0.007}} &           \textbf{0.699 $\pm$ 0.006} &  \textbf{0.690 $\pm$ 0.006} &           \textbf{0.700 $\pm$ 0.005} &           \textbf{0.692 $\pm$ 0.007} &           \textbf{0.699 $\pm$ 0.007} &           \textbf{0.687 $\pm$ 0.006} &           0.716 $\pm$ 0.006 \\
                                  & Vanilla GCN &           0.708 $\pm$ 0.003 &           0.702 $\pm$ 0.002 &                    0.633 $\pm$ 0.003 &                    0.574 $\pm$ 0.004 &           \textit{0.616 $\pm$ 0.001} &  \textit{0.550 $\pm$ 0.001} &                    0.643 $\pm$ 0.002 &                    0.594 $\pm$ 0.002 &                    0.625 $\pm$ 0.004 &                    0.568 $\pm$ 0.004 &           0.716 $\pm$ 0.003 \\
                                  & Vanilla GDC &           0.694 $\pm$ 0.001 &           0.687 $\pm$ 0.002 &                    0.622 $\pm$ 0.002 &                    0.562 $\pm$ 0.003 &                    0.618 $\pm$ 0.003 &  \textit{0.560 $\pm$ 0.004} &                    0.627 $\pm$ 0.001 &                    0.581 $\pm$ 0.003 &           \textit{0.614 $\pm$ 0.005} &                    0.562 $\pm$ 0.004 &           0.707 $\pm$ 0.001 \\
                                  & Vanilla PPRGo &  \textbf{0.719 $\pm$ 0.005} &  \textbf{0.708 $\pm$ 0.006} &                    0.677 $\pm$ 0.008 &                    0.644 $\pm$ 0.009 &                    0.681 $\pm$ 0.005 &           0.662 $\pm$ 0.004 &                    0.675 $\pm$ 0.004 &                    0.649 $\pm$ 0.003 &           \textit{0.672 $\pm$ 0.006} &           \textit{0.644 $\pm$ 0.007} &  \textbf{0.726 $\pm$ 0.006} \\
                                  & Vanilla GAT &           0.582 $\pm$ 0.006 &           0.544 $\pm$ 0.004 &           \textit{0.571 $\pm$ 0.011} &           \textit{0.520 $\pm$ 0.011} &                    0.577 $\pm$ 0.008 &           0.527 $\pm$ 0.004 &                    0.574 $\pm$ 0.004 &                    0.522 $\pm$ 0.005 &                    0.584 $\pm$ 0.012 &                    0.524 $\pm$ 0.006 &           0.647 $\pm$ 0.012 \\
                                  & Soft Medoid GDC &           0.704 $\pm$ 0.004 &           0.701 $\pm$ 0.002 &                    0.694 $\pm$ 0.004 &                    0.682 $\pm$ 0.003 &                    0.691 $\pm$ 0.003 &           0.683 $\pm$ 0.002 &           \textit{0.688 $\pm$ 0.002} &                    0.677 $\pm$ 0.003 &                    0.688 $\pm$ 0.004 &           \textit{0.675 $\pm$ 0.003} &           0.708 $\pm$ 0.003 \\
                                  & SVD GCN &           0.635 $\pm$ 0.011 &           0.623 $\pm$ 0.012 &                    0.632 $\pm$ 0.012 &                    0.617 $\pm$ 0.012 &                    0.633 $\pm$ 0.012 &           0.615 $\pm$ 0.011 &                    0.630 $\pm$ 0.010 &           \textit{0.599 $\pm$ 0.013} &           \textit{0.626 $\pm$ 0.013} &                    0.604 $\pm$ 0.009 &           0.646 $\pm$ 0.012 \\
                                  & Jaccard GCN &           0.716 $\pm$ 0.005 &           0.708 $\pm$ 0.004 &                    0.663 $\pm$ 0.004 &                    0.622 $\pm$ 0.006 &                    0.654 $\pm$ 0.004 &           0.616 $\pm$ 0.003 &                    0.666 $\pm$ 0.004 &                    0.630 $\pm$ 0.003 &           \textit{0.650 $\pm$ 0.005} &           \textit{0.609 $\pm$ 0.005} &           0.721 $\pm$ 0.005 \\
                                  & RGCN &           0.676 $\pm$ 0.006 &           0.663 $\pm$ 0.006 &           \textit{0.622 $\pm$ 0.003} &           \textit{0.568 $\pm$ 0.005} &                    0.624 $\pm$ 0.005 &           0.584 $\pm$ 0.004 &                    0.629 $\pm$ 0.003 &                    0.589 $\pm$ 0.004 &                    0.628 $\pm$ 0.004 &                    0.583 $\pm$ 0.006 &           0.686 $\pm$ 0.005 \\
\cline{1-13}
\multirow{6}{*}{\rotatebox{90}{\textbf{PubMed}}} & \underline{Soft Median GDC} &           0.761 $\pm$ 0.002 &           0.752 $\pm$ 0.003 &                                    - &                                    - &           \textit{0.721 $\pm$ 0.004} &  \textit{0.693 $\pm$ 0.005} &                                    - &                                    - &                    0.730 $\pm$ 0.005 &                    0.708 $\pm$ 0.005 &           0.769 $\pm$ 0.002 \\
                                  & \underline{Soft Median PPRGo} &           0.764 $\pm$ 0.001 &           0.752 $\pm$ 0.002 &                                    - &                                    - &  \textbf{\textit{0.723 $\pm$ 0.000}} &  \textbf{0.694 $\pm$ 0.001} &                                    - &                                    - &                    0.727 $\pm$ 0.000 &           \textit{0.692 $\pm$ 0.000} &           0.776 $\pm$ 0.002 \\
                                  & Vanilla GCN &           0.766 $\pm$ 0.003 &           0.751 $\pm$ 0.002 &                                    - &                                    - &           \textit{0.661 $\pm$ 0.003} &  \textit{0.592 $\pm$ 0.004} &                                    - &                                    - &                    0.686 $\pm$ 0.004 &                    0.620 $\pm$ 0.003 &  \textbf{0.781 $\pm$ 0.003} \\
                                  & Vanilla GDC &           0.766 $\pm$ 0.003 &           0.748 $\pm$ 0.002 &                                    - &                                    - &           \textit{0.680 $\pm$ 0.004} &  \textit{0.620 $\pm$ 0.005} &                                    - &                                    - &                    0.696 $\pm$ 0.004 &                    0.645 $\pm$ 0.005 &           0.781 $\pm$ 0.002 \\
                                  & Vanilla PPRGo &           0.717 $\pm$ 0.001 &           0.721 $\pm$ 0.007 &                                    - &                                    - &                    0.714 $\pm$ 0.001 &           0.673 $\pm$ 0.002 &                                    - &                                    - &           \textit{0.704 $\pm$ 0.007} &           \textit{0.658 $\pm$ 0.004} &           0.765 $\pm$ 0.008 \\
                                  & Soft Medoid GDC &  \textbf{0.766 $\pm$ 0.003} &  \textbf{0.756 $\pm$ 0.003} &                                    - &                                    - &           \textit{0.722 $\pm$ 0.004} &  \textit{0.693 $\pm$ 0.005} &                                    - &                                    - &           \textbf{0.732 $\pm$ 0.004} &           \textbf{0.711 $\pm$ 0.005} &           0.774 $\pm$ 0.003 \\
\cline{1-13}
\multirow{6}{*}{\rotatebox{90}{\textbf{arXiv}}} & \underline{Soft Median GDC} &           0.645 $\pm$ 0.002 &           0.629 $\pm$ 0.002 &                                    - &                                    - &                    0.504 $\pm$ 0.003 &           0.462 $\pm$ 0.001 &                                    - &                                    - &           \textit{0.479 $\pm$ 0.002} &           \textit{0.420 $\pm$ 0.005} &           0.666 $\pm$ 0.002 \\
                                  & \underline{Soft Median PPRGo} &           0.669 $\pm$ 0.001 &           0.654 $\pm$ 0.001 &                                    - &                                    - &           \textbf{0.606 $\pm$ 0.001} &  \textbf{0.589 $\pm$ 0.002} &                                    - &                                    - &  \textbf{\textit{0.598 $\pm$ 0.001}} &  \textbf{\textit{0.567 $\pm$ 0.002}} &           0.684 $\pm$ 0.001 \\
                                  & Vanilla GCN &  \textbf{0.690 $\pm$ 0.004} &  \textbf{0.671 $\pm$ 0.004} &                                    - &                                    - &                    0.361 $\pm$ 0.003 &           0.292 $\pm$ 0.005 &                                    - &                                    - &           \textit{0.351 $\pm$ 0.003} &           \textit{0.235 $\pm$ 0.006} &  \textbf{0.706 $\pm$ 0.004} \\
                                  & Vanilla GDC &           0.672 $\pm$ 0.001 &           0.648 $\pm$ 0.001 &                                    - &                                    - &                    0.446 $\pm$ 0.001 &           0.390 $\pm$ 0.001 &                                    - &                                    - &           \textit{0.399 $\pm$ 0.001} &           \textit{0.297 $\pm$ 0.003} &           0.701 $\pm$ 0.001 \\
                                  & Vanilla PPRGo &           0.680 $\pm$ 0.002 &           0.662 $\pm$ 0.002 &                                    - &                                    - &                    0.571 $\pm$ 0.001 &           0.543 $\pm$ 0.002 &                                    - &                                    - &           \textit{0.558 $\pm$ 0.003} &           \textit{0.507 $\pm$ 0.002} &           0.699 $\pm$ 0.002 \\
                                  & Soft Medoid GDC &           0.554 $\pm$ 0.004 &           0.543 $\pm$ 0.003 &                                    - &                                    - &           \textit{0.482 $\pm$ 0.001} &  \textit{0.460 $\pm$ 0.001} &                                    - &                                    - &                    0.490 $\pm$ 0.002 &                    0.460 $\pm$ 0.002 &           0.567 $\pm$ 0.004 \\
\cline{1-13}
\multirow{5}{*}{\rotatebox{90}{\textbf{Products}}} & \underline{Soft Median GDC} &           0.637 $\pm$ 0.000 &           0.624 $\pm$ 0.000 &                                    - &                                    - &                    0.605 $\pm$ 0.000 &           0.597 $\pm$ 0.000 &                                    - &                                    - &           \textit{0.599 $\pm$ 0.001} &           \textit{0.580 $\pm$ 0.001} &           0.656 $\pm$ 0.000 \\
                                  & \underline{Soft Median PPRGo} &           0.725 $\pm$ 0.001 &  \textbf{0.712 $\pm$ 0.001} &                                    - &                                    - &           \textbf{0.721 $\pm$ 0.001} &  \textbf{0.708 $\pm$ 0.001} &                                    - &                                    - &  \textbf{\textit{0.716 $\pm$ 0.001}} &  \textbf{\textit{0.695 $\pm$ 0.001}} &           0.749 $\pm$ 0.001 \\
                                  & Vanilla GCN &           0.717 $\pm$ 0.002 &           0.688 $\pm$ 0.002 &                                    - &                                    - &                    0.586 $\pm$ 0.002 &           0.538 $\pm$ 0.002 &                                    - &                                    - &           \textit{0.495 $\pm$ 0.002} &           \textit{0.465 $\pm$ 0.001} &           0.751 $\pm$ 0.002 \\
                                  & Vanilla GDC &           0.693 $\pm$ 0.000 &           0.661 $\pm$ 0.000 &                                    - &                                    - &                    0.594 $\pm$ 0.001 &           0.568 $\pm$ 0.001 &                                    - &                                    - &           \textit{0.554 $\pm$ 0.002} &           \textit{0.521 $\pm$ 0.001} &           0.736 $\pm$ 0.001 \\
                                  & Vanilla PPRGo &  \textbf{0.727 $\pm$ 0.002} &           0.711 $\pm$ 0.002 &                                    - &                                    - &                    0.713 $\pm$ 0.003 &           0.699 $\pm$ 0.004 &                                    - &                                    - &           \textit{0.696 $\pm$ 0.003} &           \textit{0.672 $\pm$ 0.004} &  \textbf{0.757 $\pm$ 0.002} \\
\bottomrule
\end{tabular}
}
\end{sidewaystable}

\clearpage
\subsection{Local Attacks}\label{sec:appendix_local}

\begin{figure}[H]
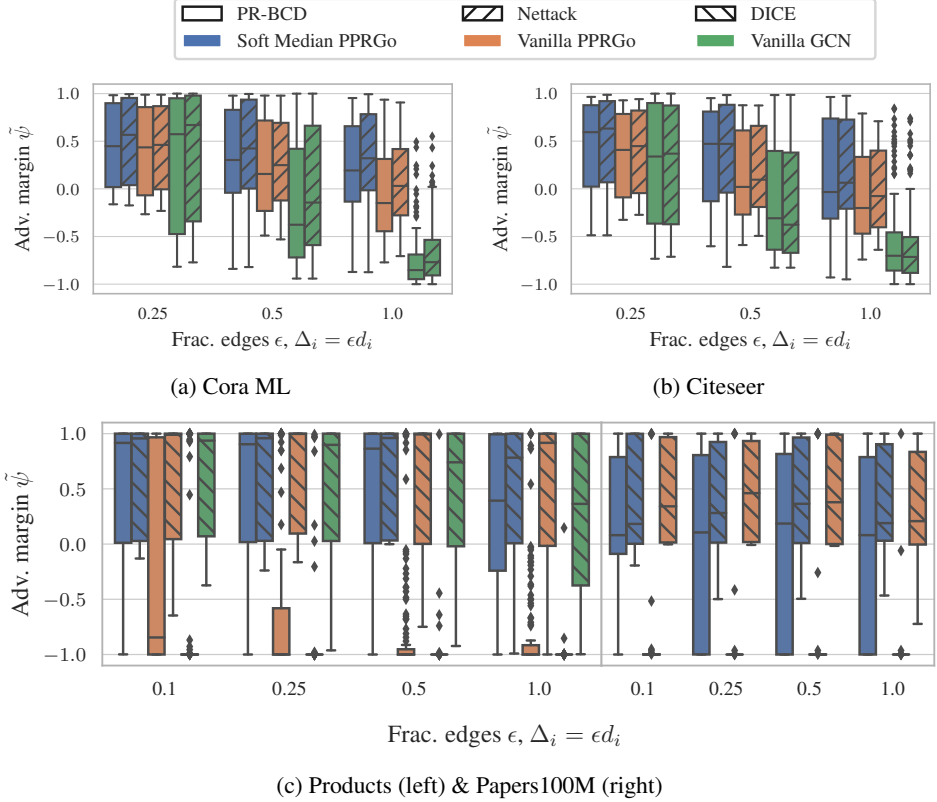

  \vspace{-7pt}
  \centering
  \hbox{\hspace{78pt} \resizebox{0.7\linewidth}{!}{\input{assets/local_prbcd_vs_nettack_cora_ml_boxplmargin_detailed_legend.pgf}}}
  \vspace{-14pt}
  \makebox[\linewidth][c]{
    \(\arraycolsep=1pt\def\arraystretch{2}\begin{array}{cc}
      \subfloat[Cora ML]{\resizebox{0.45\linewidth}{!}{\input{assets/local_prbcd_vs_nettack_cora_ml_boxplmargin_nl.pgf}}} &
      \subfloat[Citeseer]{\resizebox{0.45\linewidth}{!}{\input{assets/local_prbcd_vs_nettack_citeseer_boxplmargin_nl.pgf}}} \\
    \end{array}\)
  }
  \subfloat[Products (left) \& Papers100M (right)]{\resizebox{0.9175\linewidth}{!}{\input{assets/local_papers_and_products_boxplmargin_detailed.pgf}}}
  \caption{Adversarial classification margins \(\tilde{\psi}_i\) of the attacked nodes. This figure extends \autoref{fig:emplocal} with an additional dataset and more budgets. In (a) and (b), we compare our local PR-BCD attack with Nettack~\cite{Zugner2018} on (undirected) Cora ML and Citeseer. In (c), we show the results on the (directed) large-scale datasets Products (2.5 million nodes) and Papers 100M (111 million nodes), respectively. Our Soft Medoid PPRGo resists the attacks much better than the baselines.} \label{fig:appendix_emplocal}
\end{figure}

We present additional results for the local attacks. In \autoref{fig:appendix_emplocal} we complement \autoref{fig:emplocal} with the additional dataset Citeseer and more budgets on Products as well as Papers100M. For the experiments on the undirected Cora ML and Citeseer, we treat the graph as if it was directed during the attack and symmetrize it afterward. Despite this approximation, our attack remains strong.

\begin{table}[ht]
\centering
\caption{Attack success rates for the local attacks. \underline{Our approaches} are underlined. For the attack a higher value is better and for the defence a lower value is better. We highlight the \textbf{strongest defense} in bold.}
\label{tab:appendix_localprbcdflirate}
\vskip 0.11in
\resizebox{\linewidth}{!}{
\begin{tabular}{llcccc}
\toprule
\textbf{Dataset}     & \textbf{Attack} & \multicolumn{4}{c}{\underline{\textbf{PR-BCD}}} \\
                     & Frac. edges $\epsilon$, $\Delta_i = \epsilon d_i$ &                        0.10 &                        0.25 &                        0.50 &                        1.00 \\
\midrule
\multirow{3}{*}{\textbf{Cora ML}} & \underline{Soft Median PPRGo} &  \textbf{0.125 $\pm$ 0.003} &  \textbf{0.208 $\pm$ 0.003} &  \textbf{0.333 $\pm$ 0.004} &  \textbf{0.417 $\pm$ 0.004} \\
                     & Vanilla PPRGo &           0.133 $\pm$ 0.003 &           0.317 $\pm$ 0.004 &           0.425 $\pm$ 0.004 &           0.583 $\pm$ 0.004 \\
                     & Vanilla GCN &           0.292 $\pm$ 0.004 &           0.375 $\pm$ 0.004 &           0.642 $\pm$ 0.004 &           0.958 $\pm$ 0.002 \\
\cline{1-6}
\multirow{3}{*}{\textbf{Citeseer}} & \underline{Soft Median PPRGo} &  \textbf{0.058 $\pm$ 0.002} &  \textbf{0.242 $\pm$ 0.004} &  \textbf{0.358 $\pm$ 0.004} &  \textbf{0.517 $\pm$ 0.004} \\
                     & Vanilla PPRGo &  \textbf{0.058 $\pm$ 0.002} &           0.333 $\pm$ 0.004 &           0.492 $\pm$ 0.004 &           0.600 $\pm$ 0.004 \\
                     & Vanilla GCN &           0.125 $\pm$ 0.003 &           0.417 $\pm$ 0.004 &           0.658 $\pm$ 0.004 &           0.850 $\pm$ 0.003 \\
\cline{1-6}
\multirow{3}{*}{\textbf{Products}} & \underline{Soft Median PPRGo} &  \textbf{0.108 $\pm$ 0.003} &  \textbf{0.083 $\pm$ 0.002} &  \textbf{0.142 $\pm$ 0.003} &  \textbf{0.250 $\pm$ 0.004} \\
                     & Vanilla PPRGo &           0.683 $\pm$ 0.004 &           0.858 $\pm$ 0.003 &           0.925 $\pm$ 0.002 &           0.950 $\pm$ 0.002 \\
                     & Vanilla GCN &           0.883 $\pm$ 0.003 &           0.950 $\pm$ 0.002 &           0.992 $\pm$ 0.001 &           0.992 $\pm$ 0.001 \\
\cline{1-6}
\multirow{2}{*}{\textbf{Papers 100M}} & \underline{Soft Median PPRGo} &  \textbf{0.250 $\pm$ 0.011} &  \textbf{0.325 $\pm$ 0.012} &  \textbf{0.275 $\pm$ 0.011} &  \textbf{0.300 $\pm$ 0.011} \\
                     & Vanilla PPRGo &           0.900 $\pm$ 0.007 &           0.925 $\pm$ 0.007 &           0.875 $\pm$ 0.008 &           0.975 $\pm$ 0.004 \\
\bottomrule
\toprule
\textbf{Dataset}     & \textbf{Attack} & \multicolumn{4}{c}{\textbf{Nettack}} \\
                     & Frac. edges $\epsilon$, $\Delta_i = \epsilon d_i$ &                        0.10 &                        0.25 &                        0.50 &                        1.00 \\
\midrule
\multirow{3}{*}{\textbf{Cora ML}} & \underline{Soft Median PPRGo} &  \textbf{0.058 $\pm$ 0.002} &  \textbf{0.125 $\pm$ 0.003} &  \textbf{0.217 $\pm$ 0.003} &  \textbf{0.317 $\pm$ 0.004} \\
                     & Vanilla PPRGo &           0.092 $\pm$ 0.002 &           0.258 $\pm$ 0.004 &           0.358 $\pm$ 0.004 &           0.475 $\pm$ 0.004 \\
                     & Vanilla GCN &           0.208 $\pm$ 0.003 &           0.367 $\pm$ 0.004 &           0.533 $\pm$ 0.004 &           0.917 $\pm$ 0.002 \\
\cline{1-6}
\multirow{3}{*}{\textbf{Citeseer}} & \underline{Soft Median PPRGo} &  \textbf{0.042 $\pm$ 0.002} &  \textbf{0.150 $\pm$ 0.003} &  \textbf{0.308 $\pm$ 0.004} &  \textbf{0.458 $\pm$ 0.004} \\
                     & Vanilla PPRGo &           0.050 $\pm$ 0.002 &           0.300 $\pm$ 0.004 &           0.433 $\pm$ 0.004 &           0.608 $\pm$ 0.004 \\
                     & Vanilla GCN &           0.133 $\pm$ 0.003 &           0.425 $\pm$ 0.004 &           0.667 $\pm$ 0.004 &           0.858 $\pm$ 0.003 \\
\bottomrule
\toprule
\textbf{Dataset}     & \textbf{Attack} & \multicolumn{4}{c}{\textbf{DICE}} \\
                     & Frac. edges $\epsilon$, $\Delta_i = \epsilon d_i$ &                        0.10 &                        0.25 &                        0.50 &                        1.00 \\
\midrule
\multirow{3}{*}{\textbf{Products}} & \underline{Soft Median PPRGo} &  \textbf{0.017 $\pm$ 0.001} &  \textbf{0.017 $\pm$ 0.001} &  \textbf{0.008 $\pm$ 0.001} &  \textbf{0.100 $\pm$ 0.003} \\
                     & Vanilla PPRGo &           0.108 $\pm$ 0.003 &           0.167 $\pm$ 0.003 &           0.250 $\pm$ 0.004 &           0.308 $\pm$ 0.004 \\
                     & Vanilla GCN &           0.183 $\pm$ 0.003 &           0.233 $\pm$ 0.004 &           0.267 $\pm$ 0.004 &           0.467 $\pm$ 0.004 \\
\cline{1-6}
\multirow{2}{*}{\textbf{Papers 100M}} & \underline{Soft Median PPRGo} &           0.150 $\pm$ 0.009 &           0.125 $\pm$ 0.008 &  \textbf{0.125 $\pm$ 0.008} &  \textbf{0.100 $\pm$ 0.008} \\
                     & Vanilla PPRGo &  \textbf{0.075 $\pm$ 0.007} &  \textbf{0.100 $\pm$ 0.008} &           0.200 $\pm$ 0.010 &           0.350 $\pm$ 0.012 \\
\bottomrule
\end{tabular}
}
\end{table}

In \autoref{tab:appendix_localprbcdflirate}, we give an alternative metric to assess our attack PR-BCD and defense Soft Median PPRGo. Here we compare the \emph{attack success rate}. A ``success'' stands for a change of prediction in the course of the attack. On the upside, the attack success rate is less susceptible to the distribution of the clean margin (especially if comparing different models). On the downside, the attack success rate is not as fine-grained as comparing the margins. We see that our PR-BCD is stronger in most cases (i.e.\ the attack success rate is higher). Moreover, our Soft Median PPRGo beats all baselines for every budget except two cases for small budgets with the random, weak DICE attack. 

\begin{figure}[H]
  \vspace{-7pt}
  \centering
  \hbox{\hspace{28pt} \resizebox{0.825\linewidth}{!}{
\begingroup%
\makeatletter%
\begin{pgfpicture}%
\pgfpathrectangle{\pgfpointorigin}{\pgfqpoint{4.750697in}{0.388266in}}%
\pgfusepath{use as bounding box, clip}%
\begin{pgfscope}%
\pgfsetbuttcap%
\pgfsetmiterjoin%
\definecolor{currentfill}{rgb}{1.000000,1.000000,1.000000}%
\pgfsetfillcolor{currentfill}%
\pgfsetlinewidth{0.000000pt}%
\definecolor{currentstroke}{rgb}{1.000000,1.000000,1.000000}%
\pgfsetstrokecolor{currentstroke}%
\pgfsetstrokeopacity{0.000000}%
\pgfsetdash{}{0pt}%
\pgfpathmoveto{\pgfqpoint{0.000000in}{0.000000in}}%
\pgfpathlineto{\pgfqpoint{4.750697in}{0.000000in}}%
\pgfpathlineto{\pgfqpoint{4.750697in}{0.388266in}}%
\pgfpathlineto{\pgfqpoint{0.000000in}{0.388266in}}%
\pgfpathclose%
\pgfusepath{fill}%
\end{pgfscope}%
\begin{pgfscope}%
\pgfsetbuttcap%
\pgfsetmiterjoin%
\definecolor{currentfill}{rgb}{1.000000,1.000000,1.000000}%
\pgfsetfillcolor{currentfill}%
\pgfsetfillopacity{0.800000}%
\pgfsetlinewidth{1.003750pt}%
\definecolor{currentstroke}{rgb}{0.800000,0.800000,0.800000}%
\pgfsetstrokecolor{currentstroke}%
\pgfsetstrokeopacity{0.800000}%
\pgfsetdash{}{0pt}%
\pgfpathmoveto{\pgfqpoint{0.122222in}{0.100000in}}%
\pgfpathlineto{\pgfqpoint{4.628474in}{0.100000in}}%
\pgfpathquadraticcurveto{\pgfqpoint{4.650697in}{0.100000in}}{\pgfqpoint{4.650697in}{0.122222in}}%
\pgfpathlineto{\pgfqpoint{4.650697in}{0.266044in}}%
\pgfpathquadraticcurveto{\pgfqpoint{4.650697in}{0.288266in}}{\pgfqpoint{4.628474in}{0.288266in}}%
\pgfpathlineto{\pgfqpoint{0.122222in}{0.288266in}}%
\pgfpathquadraticcurveto{\pgfqpoint{0.100000in}{0.288266in}}{\pgfqpoint{0.100000in}{0.266044in}}%
\pgfpathlineto{\pgfqpoint{0.100000in}{0.122222in}}%
\pgfpathquadraticcurveto{\pgfqpoint{0.100000in}{0.100000in}}{\pgfqpoint{0.122222in}{0.100000in}}%
\pgfpathclose%
\pgfusepath{stroke,fill}%
\end{pgfscope}%
\begin{pgfscope}%
\pgfsetbuttcap%
\pgfsetmiterjoin%
\definecolor{currentfill}{rgb}{0.347059,0.458824,0.641176}%
\pgfsetfillcolor{currentfill}%
\pgfsetlinewidth{0.501875pt}%
\definecolor{currentstroke}{rgb}{0.298039,0.298039,0.298039}%
\pgfsetstrokecolor{currentstroke}%
\pgfsetdash{}{0pt}%
\pgfpathmoveto{\pgfqpoint{0.144444in}{0.166044in}}%
\pgfpathlineto{\pgfqpoint{0.366667in}{0.166044in}}%
\pgfpathlineto{\pgfqpoint{0.366667in}{0.243822in}}%
\pgfpathlineto{\pgfqpoint{0.144444in}{0.243822in}}%
\pgfpathclose%
\pgfusepath{stroke,fill}%
\end{pgfscope}%
\begin{pgfscope}%
\definecolor{textcolor}{rgb}{0.150000,0.150000,0.150000}%
\pgfsetstrokecolor{textcolor}%
\pgfsetfillcolor{textcolor}%
\pgftext[x=0.455556in,y=0.166044in,left,base]{\color{textcolor}\rmfamily\fontsize{8.000000}{9.600000}\selectfont \(\displaystyle \epsilon=0.05\)}%
\end{pgfscope}%
\begin{pgfscope}%
\pgfsetbuttcap%
\pgfsetmiterjoin%
\definecolor{currentfill}{rgb}{0.798529,0.536765,0.389706}%
\pgfsetfillcolor{currentfill}%
\pgfsetlinewidth{0.501875pt}%
\definecolor{currentstroke}{rgb}{0.298039,0.298039,0.298039}%
\pgfsetstrokecolor{currentstroke}%
\pgfsetdash{}{0pt}%
\pgfpathmoveto{\pgfqpoint{1.092791in}{0.166044in}}%
\pgfpathlineto{\pgfqpoint{1.315014in}{0.166044in}}%
\pgfpathlineto{\pgfqpoint{1.315014in}{0.243822in}}%
\pgfpathlineto{\pgfqpoint{1.092791in}{0.243822in}}%
\pgfpathclose%
\pgfusepath{stroke,fill}%
\end{pgfscope}%
\begin{pgfscope}%
\definecolor{textcolor}{rgb}{0.150000,0.150000,0.150000}%
\pgfsetstrokecolor{textcolor}%
\pgfsetfillcolor{textcolor}%
\pgftext[x=1.403903in,y=0.166044in,left,base]{\color{textcolor}\rmfamily\fontsize{8.000000}{9.600000}\selectfont \(\displaystyle \epsilon=0.1\)}%
\end{pgfscope}%
\begin{pgfscope}%
\pgfsetbuttcap%
\pgfsetmiterjoin%
\definecolor{currentfill}{rgb}{0.374020,0.618137,0.429902}%
\pgfsetfillcolor{currentfill}%
\pgfsetlinewidth{0.501875pt}%
\definecolor{currentstroke}{rgb}{0.298039,0.298039,0.298039}%
\pgfsetstrokecolor{currentstroke}%
\pgfsetdash{}{0pt}%
\pgfpathmoveto{\pgfqpoint{1.982110in}{0.166044in}}%
\pgfpathlineto{\pgfqpoint{2.204332in}{0.166044in}}%
\pgfpathlineto{\pgfqpoint{2.204332in}{0.243822in}}%
\pgfpathlineto{\pgfqpoint{1.982110in}{0.243822in}}%
\pgfpathclose%
\pgfusepath{stroke,fill}%
\end{pgfscope}%
\begin{pgfscope}%
\definecolor{textcolor}{rgb}{0.150000,0.150000,0.150000}%
\pgfsetstrokecolor{textcolor}%
\pgfsetfillcolor{textcolor}%
\pgftext[x=2.293221in,y=0.166044in,left,base]{\color{textcolor}\rmfamily\fontsize{8.000000}{9.600000}\selectfont \(\displaystyle \epsilon=0.25\)}%
\end{pgfscope}%
\begin{pgfscope}%
\pgfsetbuttcap%
\pgfsetmiterjoin%
\definecolor{currentfill}{rgb}{0.710784,0.363725,0.375490}%
\pgfsetfillcolor{currentfill}%
\pgfsetlinewidth{0.501875pt}%
\definecolor{currentstroke}{rgb}{0.298039,0.298039,0.298039}%
\pgfsetstrokecolor{currentstroke}%
\pgfsetdash{}{0pt}%
\pgfpathmoveto{\pgfqpoint{2.930457in}{0.166044in}}%
\pgfpathlineto{\pgfqpoint{3.152679in}{0.166044in}}%
\pgfpathlineto{\pgfqpoint{3.152679in}{0.243822in}}%
\pgfpathlineto{\pgfqpoint{2.930457in}{0.243822in}}%
\pgfpathclose%
\pgfusepath{stroke,fill}%
\end{pgfscope}%
\begin{pgfscope}%
\definecolor{textcolor}{rgb}{0.150000,0.150000,0.150000}%
\pgfsetstrokecolor{textcolor}%
\pgfsetfillcolor{textcolor}%
\pgftext[x=3.241568in,y=0.166044in,left,base]{\color{textcolor}\rmfamily\fontsize{8.000000}{9.600000}\selectfont \(\displaystyle \epsilon=0.5\)}%
\end{pgfscope}%
\begin{pgfscope}%
\pgfsetroundcap%
\pgfsetroundjoin%
\pgfsetlinewidth{1.003750pt}%
\definecolor{currentstroke}{rgb}{0.100000,0.100000,0.100000}%
\pgfsetstrokecolor{currentstroke}%
\pgfsetdash{}{0pt}%
\pgfpathmoveto{\pgfqpoint{3.819775in}{0.204933in}}%
\pgfpathlineto{\pgfqpoint{4.041997in}{0.204933in}}%
\pgfusepath{stroke}%
\end{pgfscope}%
\begin{pgfscope}%
\pgfsetbuttcap%
\pgfsetroundjoin%
\definecolor{currentfill}{rgb}{0.100000,0.100000,0.100000}%
\pgfsetfillcolor{currentfill}%
\pgfsetlinewidth{1.003750pt}%
\definecolor{currentstroke}{rgb}{0.100000,0.100000,0.100000}%
\pgfsetstrokecolor{currentstroke}%
\pgfsetdash{}{0pt}%
\pgfsys@defobject{currentmarker}{\pgfqpoint{-0.041667in}{-0.041667in}}{\pgfqpoint{0.041667in}{0.041667in}}{%
\pgfpathmoveto{\pgfqpoint{-0.041667in}{-0.041667in}}%
\pgfpathlineto{\pgfqpoint{0.041667in}{0.041667in}}%
\pgfpathmoveto{\pgfqpoint{-0.041667in}{0.041667in}}%
\pgfpathlineto{\pgfqpoint{0.041667in}{-0.041667in}}%
\pgfusepath{stroke,fill}%
}%
\begin{pgfscope}%
\pgfsys@transformshift{3.930886in}{0.204933in}%
\pgfsys@useobject{currentmarker}{}%
\end{pgfscope}%
\end{pgfscope}%
\begin{pgfscope}%
\definecolor{textcolor}{rgb}{0.150000,0.150000,0.150000}%
\pgfsetstrokecolor{textcolor}%
\pgfsetfillcolor{textcolor}%
\pgftext[x=4.130886in,y=0.166044in,left,base]{\color{textcolor}\rmfamily\fontsize{8.000000}{9.600000}\selectfont Accuracy}%
\end{pgfscope}%
\end{pgfpicture}%
\makeatother%
\endgroup%
}}
  \resizebox{0.8\linewidth}{!}{\input{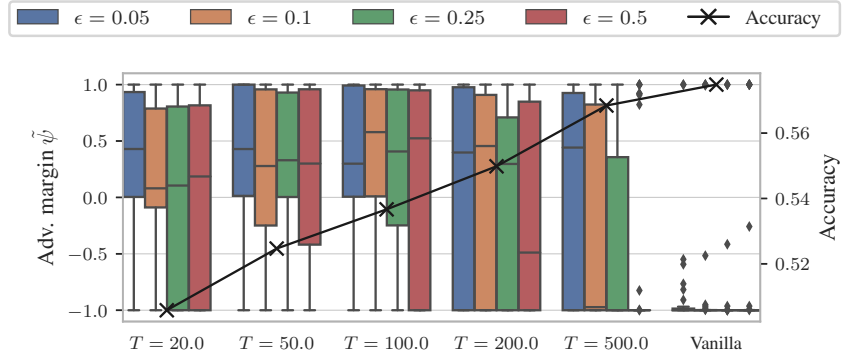}}
  \caption{Adversarial classification margins \(\tilde{\psi}_i\) (left y axis) and clean accuracy (right y axis) over the temperature and for different budgets on Papers100M (111 million nodes). We see that the clean accuracy increases if the temperature increases. The robustness decreases for very low and large temperatures. There is a sweet spot for maximum robustness in between.\label{fig:appendix_paperstemperature}}
\end{figure}

\textbf{The temperature hyperparameter $T$.} Last, we study how the temperature affects the robustness of the Soft Median PPRGo model (see \autoref{fig:appendix_paperstemperature}). Similarly to \citep{Geisler2020}, we observe that for large values it performs comparably to the vanilla model. If we lower the temperature the robustness increases at the cost of a slightly lower clean accuracy. If we set the temperature too low, the accuracy still declines but also does the robustness. This shows the trade-off between clean accuracy and robustness. Hence, one needs to carefully choose a good temperature for the application at hand. A good strategy is to (1) tune a vanilla model to meet the desired predictive performance and (2) successively decay the temperature (starting high) until the robustness is decreasing or until the drop in predictive performance exceeds the application-specific threshold.

\subsection{Relationship of Graph Size and GNNs Robustness}\label{sec:appendix_graphsize}

We now analyze the results of PR-BCD w.r.t.\ the graph size. We start comparing the global attacks with a budget of \(\epsilon=0.1\) for the Vanilla GCN model. We observe a relative drop in the adversarial accuracy by 22\% on Cora, 67 \% on arXiv, and 38\% on Products. On the larger graphs, the degradation of the accuracy is much larger which indicates a relationship between the adversarial robustness and the size of the graph. This relationship seems to persist for architectures other than GCN as well. Of course, this comparison neglects the characteristics of the dataset itself (e.g.\ in contrast to Cora ML, arXiv's test set is smaller and it contains continuous features). However, the trend that large graphs are more fragile becomes more radical if we consider local attacks. For local attacks on large graphs, we observe that even small budgets suffice to fool almost all nodes. On small graphs, PPRGo already seems to be quite an effective defense in comparison to a Vanilla GCN (see \autoref{fig:emplocal}). In contrast, without our Soft Median and particularly on the large graphs, it is easy to flip the prediction of basically every node (also compare with \autoref{tab:appendix_localprbcdflirate}). We leave a detailed and rigorous study of this relationship for future work.

\end{document}